%% file: jmlr.tex
\documentclass[twoside,11pt]{article}

\usepackage[abbrvbib]{jmlr2e}
\usepackage{booktabs}
\usepackage{xcolor}
\usepackage{soul}
\usepackage{tikz}
\usetikzlibrary{shapes,arrows,decorations.pathmorphing,backgrounds,positioning,fit,automata}
\usepackage{float}

\usepackage{lastpage}
\jmlrheading{24}{2023}{1-\pageref{LastPage}}{8/22; Revised
4/23}{6/23}{22-0880}{Connor Lawless, Sanjeeb Dash, Oktay G\"unl\"uk, and Dennis Wei}

\ShortHeadings{Boolean Rule Sets via Column Generation}{Lawless, Dash, G\"unl\"uk, and Wei}
\firstpageno{1}

\include{preamble}

\begin{document}

\title{Interpretable and Fair Boolean Rule Sets via Column Generation}

\author{\name Connor Lawless\textsuperscript{1} \email cal379@cornell.edu \\
       \AND
       \name Sanjeeb Dash\textsuperscript{2} \email sanjeebd@us.ibm.com \\
       \AND
       \name Oktay  G\"unl\"uk\textsuperscript{1} \email ong5@cornell.edu \\
       \AND
       \name Dennis Wei\textsuperscript{2} \email dwei@us.ibm.com \\
       \addr \textsuperscript{1} Operations Research and Information Engineering, Cornell University, Ithaca, NY, 14850   \\
       \addr \textsuperscript{2} IBM Research, Yorktown Heights, NY, 10598
       }
       
\editor{Silvia Chiappa}

\maketitle

\begin{abstract}%
This paper considers the learning of Boolean rules in disjunctive normal form (DNF, OR-of-ANDs, equivalent to decision rule sets) as an interpretable model for classification.  An integer program is formulated to optimally trade classification accuracy for rule simplicity. We also consider the fairness setting and extend the formulation to include explicit constraints on two different measures of classification parity: equality of opportunity and equalized odds. Column generation (CG) is used to efficiently search over an exponential number of candidate rules without the need for heuristic rule mining. To handle large data sets, we propose an approximate CG algorithm using randomization.  Compared to three recently proposed alternatives, the CG algorithm dominates the accuracy-simplicity trade-off in 8 out of 16 data sets. When maximized for accuracy, CG is competitive with rule learners designed for this purpose, sometimes finding significantly simpler solutions that are no less accurate. Compared to other fair and interpretable classifiers, our method is able to find rule sets that meet stricter notions of fairness with a modest trade-off in accuracy.
\end{abstract}


\section{Introduction}\label{sec:intro}
In recent years, key decision making tasks in areas ranging from finance to driving have been automated via machine learning (ML) tools. However,  many ML techniques are ``black boxes" that do not provide the rationale behind their predictions. This aspect renders ML unsuitable for making high stakes decisions in areas such as criminal justice and medicine, where ML is mostly used as a support tool to complement human decision making. In these areas, transparency is necessary for domain experts to understand, critique and consequently trust the ML models. 
In decision making tasks with a large societal impact, a natural question is whether or not the ML model is fair to all those affected. Recent papers \citep[e.g.][]{mehrabi2019survey} have argued that popular ML algorithms can be racially biased in applications as varied as facial identification in picture tagging to predicting criminal recidivism. Designing ML algorithms that are accurate, fair AND interpretable is therefore an important societal goal. 

Many classification tasks 
can be expressed as mathematical optimization problems. Discrete optimization is a natural tool for a variety of interpretable ML models that can be represented by low-complexity discrete objects. Integer programming (IP), a technique within discrete optimization, is widely used in many industrial applications such as production planning, scheduling, and logistics. However, it is less widely used in ML applications partly because of large data sets that lead to large-scale IPs that are computationally intractable. However, recent algorithmic and hardware advances have led to IP being used for certain ML problems such as optimal decision trees \citep{gunluk2019optimal, bertsimas2017, molero2020}, risk scores \citep{ustun2019learning}, and rule sets \citep{wang2015} amongst others.

\subsection{Contributions}\label{sec:contrib}
In this paper, we use discrete optimization to construct interpretable classifiers that can include constraints on fairness. We focus on a well-studied interpretable class of ML models for binary classification, namely rule sets in disjunctive normal form (DNF, `OR-of-ANDs'). For example, a DNF rule set with two rules for predicting criminal recidivism could be 
\begin{center}
\big[\text{(Priors $\ge$ 3) AND (Age $\le$ 45) AND (Score Factor $=$ TRUE)}\big] \\ \text{~OR~}
\\\big[\text{(Priors $\ge$ 20) AND (Age $\ge$ 45)}\big] 
\end{center}

\noindent where Priors, Age, and Score Factor are features related to the defendant. This rule set has two rules, also known as clauses, each of which check certain conditions on the features of the data.
The fewer the rules or conditions in each rule, the more interpretable the rule set.
 Our approach can also be used to learn rule sets in conjunctive normal form (CNF, `AND-of-ORs').
 
In contrast to other interpretable model classes related to rule sets 
such as decision trees \citep{breiman1984, quinlan1993}  and decision lists \citep{rivest1987}, \cl{the rules within a DNF rule set are unordered (i.e., do not need to be evaluated in a hierarchical structure) and have been shown in a user study to require less effort to understand} \citep{lakkaraju2016}. We summarize our main contributions as follows:



\begin{itemize}
    \item We propose an IP formulation for Boolean rule (DNF) learning that aims to minimize Hamming loss, a computationally efficient proxy for 0-1 classification loss. The formulation includes explicit bounds on the complexity of the rule set to enhance interpretability and prevent overfitting.
    \item Rather than mining rules, we use the technique of column generation (CG) to search over the exponential number of all possible rules, without enumerating even a pre-mined subset (which can be large).
    \item We present an IP formulation for the Pricing Problem in CG to generate candidate rules. For large data sets we present a greedy heuristic and sub-sampling schemes to approximately solve the Pricing Problem.
    \item We extend the IP formulation to include controls on two forms of algorithmic fairness: equality of opportunity and equalized odds. 
    \item We present an empirical study on the performance of Hamming loss as a proxy for 0-1 loss.
    \item A numerical evaluation is presented using $16$ data sets, including one from the FICO Explainable Machine Learning Challenge \citep{FICO2018}, in the standard classification setting, as well as $3$ data sets for fair classification.
\end{itemize}

An initial version of this work was published in a conference proceeding \citep{dash2018boolean} that introduced our IP formulation and column generation framework. In this article, we extend the formulation to the fairness setting and integrate two notions of classification parity as constraints. In addition, we provide a deeper analysis of Hamming loss as a proxy for 0-1 classification loss, proving a negative theoretical result and conducting a computational study of its performance in practice. Finally, we present new computational studies that show the performance of our approach in the fairness setting, and the performance of new computational heuristics.

\subsection{Related Work}\label{sec:related}
Our work builds upon a body of work related to learning rule sets, leveraging discrete optimization for machine learning, and fair machine learning.

\textbf{Rule Sets: } The learning of Boolean rules and rule sets has an extensive history spanning multiple fields.  DNF learning theory \citep[e.g.][]{valiant1984,klivans2004,feldman2012} focuses on the ideal noiseless setting (sometimes allowing arbitrary queries) and is less relevant to the practice of learning compact models from noisy data.  Predominant practical approaches include a covering or separate-and-conquer strategy \citep{clark1989,clark1991,cohen1995,frank1998,friedman1999,marchand2003,fuernkranz2012} of learning rules one by one and removing ``covered'' examples, a bottom-up strategy of combining more specific rules into more general ones \citep{salzberg1991,domingos1996,muselli2002}, and associative classification in which association rule mining is followed by rule selection using various criteria \citep{liu1998,li2001,yin2003,wang2005,chen2006,cheng2007}.  Broadly speaking, these approaches employ heuristics and/or multiple criteria not directly related to classification accuracy.  Moreover, they do not explicitly consider model complexity, a problem that has been noted especially with associative classification.  Rule set models have been generalized to rule ensembles \citep{cohen1999,friedman2008,dembczynski2010,wei2019generalized}, using boosting and linear combination rather than logical disjunction; the interpretability of such models is again not comparable to rule sets.  Models produced by logical analysis of data \citep{boros2000,hammer2006} from the operations research community are similarly weighted linear combinations.

Spurred by the recent demand for interpretable models, several papers have revisited Boolean and rule set models and proposed methods that jointly optimize accuracy and simplicity within a single objective function.  These works restricted the problem and approximated its solution.  In \cite{lakkaraju2016,wang2017,wang2015}, frequent rule miners are first used to produce a set of candidate rules.  A greedy forward-backward algorithm \citep{lakkaraju2016}, simulated annealing \citep{wang2017}, or integer programming (IP) \citep[in an unpublished manuscript][]{wang2015} are then used to select rules from the candidates.  The drawback of rule mining is that it limits the search space while often still producing a large number of rules, which then have to be filtered using criteria such as information gain.  \cite{wang2015} also presented an IP formulation (but no computational results) that jointly constructs and selects rules without pre-mining.  \cite{su2016} developed an IP formulation for DNF and CNF learning in which the number of rules (conjunctions or disjunctions) is fixed.  The problem is then solved approximately by decomposing into subproblems and applying a linear programming (LP) method \citep{malioutov2013}, which requires rounding of fractional solutions.

\textbf{Discrete optimization for machine learning: } \cl{Beyond rule sets, the last few years have seen a renewed interest in using discrete optimization to solve machine learning problems \cl{\citep[see][for an overview of recent work]{gambella2021optimization}}. Ustun and Rudin used IP methods to create sparse linear integer models for classification \citep{ustun2014supersparse} and risk scores \citep{ustun2019learning}. IP models to construct decision trees are studied in \cite{bennett1996optimal,aghaei2019learning, bertsimas2017, molero2020, gunluk2019optimal,verwer2019learning,hu2019optimal,lin2020generalized,aglin2020learning}. Most of these approaches aim to establish a certificate of optimality at the expense of computational effort, requiring hours of computation time. In contrast, our approach does not guarantee a certificate of optimality but uses column generation with a time limit as a faster heuristic. We note that CG has been proposed for other machine learning tasks such as boosting \citep{demiriz2002,bi2004}, regression \citep{eckstein2017rule}, prescriptive decision trees \citep{subramanian2022constrained}, matrix factorization \citep{kovacs2021binary}, support vector machines \citep{carrizosa2010binarized}, and hash learning \citep{li2013}. 
}

\textbf{Fair machine learning: } Quantifying fairness is not a straightforward task and a number of metrics have been proposed in the fair machine learning literature. These metrics broadly fall into approaches for individual fairness \citep{dwork2011fairness}, which ensure that `similar' individuals with respect to the classification task have similar outcomes, or group fairness, which ensures similar treatment for members of different protected groups (i.e., race, gender, sexual orientation). Group fairness metrics can be further broken down into three approaches: disparate treatment, classification parity, and calibration \citep{corbettdavies2018measure}. \cl{The first measure, disparate treatment, ensures that predictions are not being made based on sensitive attributes. It has been addressed in the literature by excluding sensitive features (e.g., race, gender), or proxies of these sensitive attributes, from the data. However removing sensitive attributes can lead to sub-optimal predictive performance \citep{corbettdavies2018measure}.} Classification parity, or group fairness, ensures that some statistical measure of the predictions (ex.~Type I/II error, accuracy) is equal across all groups. Recent results have built fair classifiers around various related metrics including demographic parity \citep{agarwal2018reductions, calders2010, dwork2011fairness, edwards2015censoring, kamiran2012,kamishima2012fairness,zafar2015fairness,celis2019classification}, equalized odds \citep{agarwal2018reductions, hardt2016equality, Zafar_2017,celis2019classification}, and equality of opportunity \citep{hardt2016equality, Zafar_2017,donini2018empirical}. The final measure, calibration, requires that conditioned on the prediction, the actual outcomes are independent of the protected characteristics. For example, among those given a 70\% prediction of repeating a criminal offence, calibration would require that Black and white offenders repeat crimes at a similar rate. Importantly, recent impossibility results  \citep{Chouldechova_2017, kleinberg2016inherent} have shown that simultaneously attaining  perfect calibration and certain measures of classification parity is not possible. Thus, the goal of a fair classifier is to maximize predictive accuracy subject to some requirement on fairness. 


Our work focuses on  explicitly integrating fairness considerations directly into the training of a classification model. Previous work in this area, referred to as \emph{in-processing} \citep{agarwal2018reductions,calders2010,kamishima2012fairness,zafar2015fairness,celis2019classification,Zafar_2017,donini2018empirical,berk2017convex,wu2018fairnessaware,pmlr-v119-lohaus20a}, has focused on adding either some form of fairness regularization to the loss function \citep{kamishima2012fairness,berk2017convex, pmlr-v28-zemel13}, or a constraint to the underlying optimization problem \citep{aghaei2019learning, zafar2015fairness, Zafar_2017,donini2018empirical}. However many current approaches require the use of a relaxed version of the fairness constraints (i.e., convex, linear) during optimization \citep{donini2018empirical, wu2018fairnessaware, zafar2015fairness, Zafar_2017}, which have been shown to have sub-par fairness on out-of-sample data \citep{pmlr-v119-lohaus20a}. Similar to our approach, \cite{aghaei2019learning} formulate optimal decision trees subject to explicit constraints on fairness. \cl{However, unlike our approach which addresses Boolean rules and uses heuristics to speed up the solve time, their approach aims to solve the MIP formulation to optimality.}

\subsection{Outline of the Paper}\label{sec:outline}

The remainder of the paper is organized as follows. Section \ref{sec:class} introduces our MIP formulation for constructing optimal Boolean rule sets from training data. Section \ref{sec:fairness} extends this framework to the two different notions of fairness that we consider. Section \ref{sec:cg} describes the column generation procedure to generate candidate rules. It also discusses  computational approaches to improve the speed and scalability of our framework as well as the optimality guarantees in our framework. In Section \ref{sec:exp}  we present empirical results on testing data to measure the performance of our approach using cross validation.

%

\section{Classification Framework: Boolean Rule Sets}\label{sec:class}
We consider the supervised binary classification setting where we are given a training data set of $n$ samples \cl{sampled from an underlying (unknown) distribution} with $d$ features $(\textbf{X}_i, y_i)$, $i \in I = \{1,\dots,n\}$ where $\textbf{X}_i \in \{0,1\}^d$ and labels $y_i \in \{-1, 1\}$. A sample $i \in I$ is said to have a feature $j$ if $X_{ij} = 1$ where $X_{ij}$ is the $j$-th element of $\textbf{X}_i$. Assuming the data to be binary-valued is not a restrictive assumption in practice. \cl{For instance, categorical features can be converted to binary features via a one-hot encoding scheme. Similarly, real valued features can be converted to binary features by considering a sequence of thresholds (i.e., creating new binary features $X_{ij} \leq v$ for real valued feature $j$ and different thresholds $v$). Binarization of categorical and real-valued features does come at a cost to the dimensionality of the data set (i.e., it may take multiple binary columns to represent a single categorical or real-valued feature). A detailed discussion of how we deal with numerical and categorical features is included in Section \ref{sec:exp}.} 

We focus on the problem of learning a Boolean classifier $\hat{y}(\textbf{X})$ in DNF (OR-of-ANDs). When the features are binary each clause in a DNF corresponds to a conjunction of features or their negation (i.e., $A \land \neg B \land C$ for features A, B and C). If for every feature we assume its negation is also included as a feature (i.e., for every feature $A$ there exists a feature $\neg A$), then a clause simply corresponds to a conjunction of features, and a sample satisfies a clause if it has all features contained in the clause (i.e., ~$X_{ij} = 1$ for all such features $j$). Since a DNF classifier is equivalent to a rule set, the terms clause, conjunction, and (single) rule (within a rule set) are used interchangeably. As shown in \citep{su2016} using De Morgan's laws, the same formulation applies equally well to CNF learning by negating both labels $y_i$ and features $\bx_i$.  The method can also be extended to multi-class classification in the usual one-versus-rest manner.


Let $\K$ denote the set of all candidate rules and $\K_i\subset\mathcal{K}$ be the set of rules satisfied by data point $i\in I$. Furthermore, let $K \subseteq \mathcal{K}$  be a DNF rule set {composed of candidate rules}  selected  from $\mathcal{K}$. \cl{For a given DNF rule set $K$, the classifier $\hat{y}(\textbf{X}_i)$   checks whether the data point $\textbf{X}_i$ satisfies at least one rule in $K$.

\begin{definition}[DNF Classifier]
Given a rule set $K$, the DNF classifier is defined as follows:
$$\hat{y}(\textbf{X}_i) ~=~
\begin{cases}
1 & \text{if } \abs{{\cal K}_i \cap K} > 0 \\
-1 &\text{else } 
\end{cases}$$
\end{definition}
}

\cl{Note that not every possible rule may be included in the set of candidate rules. For instance, there may be a limit on the number of features allowed in a candidate rule.} In any case, given $d$ binary features, there can only be  at most a finite number ($2^d-1$) of possible decision rules. 
Therefore, in theory it is possible to enumerate all possible rules\cl{, though in practice this may be computationally intractable,} and then formulate a large scale integer program (IP) to select a small subset of these rules that minimizes error on the training data. 
In this framework, it is also possible to explicitly require the rule set to satisfy certain properties such as fairness or interpretability.
However, for most practical applications, such an IP would be onerously large and computationally intractable. We introduce a column generation framework to tackle this challenge in Section~\ref{sec:cg}.

 \subsection{0-1 loss} \label{sec:01}
  
 When constructing a rule set, our ultimate aim is to minimize 0-1 classification error, which is equivalent to maximizing the classification accuracy. Assume that the data points are partitioned into two sets based on their labels: 
	\begin{equation*}\mathcal{P} = \{i\in I : y_i = 1\},\text{~~and~~}\mathcal{N} = \{i\in I : y_i = -1\}.\end{equation*}
We call ${\cal P}$ and ${\cal N}$ the positive and negative classes respectively.  For data points from the positive class (i.e., $i \in {\cal P}$), the 0-1 loss is simply the indicator that the data point satisfies no rules in the rule set (i.e., \cl{$\abs{{\cal K}_i \cap K} = 0$}). For points in the negative class, the 0-1 loss is the indicator of whether the data point satisfies at least one rule in the rule set (i.e., $\abs{\mathcal{K}_i \cap K} > 0$). Putting both terms together, we get that the 0-1 loss for a data point $(\textbf{X}_i, y_i)$ and rule set $K$ is as follows:

\begin{definition}[0-1 loss]
$$\ell_{01}(\textbf{X}_i, y_i, K) ~=~
\begin{cases}
\mathbb{I}(\abs{{\cal K}_i \cap K }= 0) & \text{if } y_i = 1\\
\mathbb{I}(\abs{\mathcal{K}_i \cap K} > 0)&\text{if } y_i = -1
\end{cases}$$
\end{definition}
\noindent
where $\mathbb{I}(\mathcal{E})$ is the indicator function (i.e., $\mathbb{I}(\mathcal{E}) = 1$ if $\mathcal{E}$ is true and 0 otherwise). Let  $w_{k}\in\{0,1\}$  be a  variable indicating if rule $k\in\K$ is selected in $K$;  $\zeta_i\in\{0,1\}$  be a  variable indicating if data point $i\in \P \cup {\cal N}$ is misclassified. With this notation in mind, the problem of identifying the rule set that minimizes 0-1 loss becomes

\begin{align}
\textbf{min} \quad & \sum_{i\in \cal{P}} \zeta_i +\sum_{i\in \cal{N}} \zeta_i ~~ \label{obj:01}\\
	\textbf{s.t.}  \quad
	& \zeta_i + \sum_{k\in {\cal K}_i} w_k \geq 1 , \quad \forall i \in \cal{P}~~ \label{const:accP01} \\
	& w_k  \leq   \zeta_i , \quad  \forall i \in {\cal N}, k \in {\cal K}_i ~~ \label{const:accZ01} \\[.1cm]
&	w\in \{0,1\}^{\abs{\mathcal{K}}},~ \zeta\in \{0,1\}^{\abs{\mathcal{P} \cup {\cal N}}} ~~ \label{const:accBin01}
\end{align}
Any feasible solution $(\bar w,\bar \zeta)$ to \eqref{const:accP01}-\eqref{const:accBin01} corresponds to a rule set $K=\{k\in\K\::\: \bar w_k=1\}$. 
Constraint \eqref{const:accP01} identifies false negatives by forcing $\zeta_i $ to take value 1 if no rule that is satisfied by the point $i \in \mathcal{P}$ is selected. Similarly, constraint \eqref{const:accZ01} identifies false positives by forcing $\zeta_i$ to take a value of 1 if any rule satisfied by $i \in \mathcal{N}$ is selected. The objective forces $\zeta_i$ to be 0 when possible, so there is no need to add constraints to track whether data point $i$ is classified correctly in this formulation. 
\cl{Note that there may be an exponential number of rules with respect to the number of features $d$,  and thus there may be an exponential number of constraint (\ref{const:accZ01}) as there needs to be one constraint for each rule met by data point $i$. 
We present and discuss an alternative version of this formulation with aggregated false positive constraints in Appendix \ref{app:01aggregate}.}

 \subsection{Hamming loss} \label{sec:ham}

While our aim is to minimize the 0-1 loss, the corresponding IP formulation is prohibitively large and hard to solve in practice. We instead minimize the {\em Hamming loss} of the rule set as is also done in \cite{su2016,lakkaraju2016}.  For each incorrectly classified sample, the Hamming loss counts 
    the number of rules that have to be selected or removed to classify it correctly.
	More precisely, it 
    is equal to the number of samples with label 1 that are classified incorrectly (false negatives) plus  the sum of the number of selected rules that each sample with label -1 satisfies.
    
\begin{definition}[Hamming loss]
$$\ell_{h}(\textbf{X}_i, y_i, K) ~=~
\begin{cases}
\mathbb{I}(\abs{{\cal K}_i \cap K }= 0) & \text{if } y_i = 1\\
~\abs{\mathcal{K}_i \cap K} &\text{if } y_i = -1
\end{cases}$$

\end{definition}
Notice that Hamming loss is asymmetric with respect to errors for the positive and negative classes. Specifically, while the loss for a false negative remains the same as 0-1 loss, a false positive may incur a loss greater than 1 if it is satisfies multiple rules.\footnote{\cl{If using this framework for CNF rules with Hamming loss, the false positive loss remains the same and the false negative loss is relaxed (i.e., may be higher than one).}}
Using the same notation as the previous IP formulation \eqref{obj:01}---\eqref{const:accBin01}, the problem of finding the rule set that minimizes Hamming loss is simply

\begin{align}
\textbf{min} \quad & \sum_{i\in \cal{P}} \zeta_i +\sum_{i\in \cal{N}} \sum_{k \in {\cal K}_i} w_k ~~ \label{obj:ham}\\
	\textbf{s.t.}  \quad
	& \zeta_i + \sum_{k\in {\cal K}_i} w_k \geq 1 , \quad i \in \cal{P}~~ \label{const:hamP} \\
	&	w\in \{0,1\}^{\abs{\mathcal{K}}},~ \zeta\in \{0,1\}^{\abs{\mathcal{P}}}  \label{const:hamBin}
\end{align}

The objective is now Hamming loss where for each $i\in \mathcal{N}$ the second term adds up the total number of selected rules satisfied by $i$. Compared to the 0-1 loss formulation, this formulation does not have the large number of constraints ($\sum_{i \in {\cal N}}\abs{{\cal K}_i}$ constraints) needed to track false positives. Note that the size of the set $\abs{{\cal K}_i}$ can be exponentially large with respect to the dimensionality of the feature space. While Hamming loss leads to a much more compact formulation, we next observe that optimizing it might also lead to an arbitrarily bad overestimation of the 0-1 loss.

\begin{theorem}[Hamming loss vs. 0-1 loss] \label{theorem:hamm}
When evaluating the 0-1 loss on a data set $\mathcal{D}$, the rule set selected to minimize Hamming loss can perform arbitrarily worse than the rule set selected to minimize 0-1 loss. Formally, for a data set $\mathcal{D} = [(\mathbf{X}_i, y_i)]_1^n$ where $\mathbf{X}_i \in \{0,1\}^d$, let 
$$K^*_{\ell} = \argmin_{K \subseteq \mathcal{K}} \sum_{i=1}^{n} \ell(\mathbf{X}_i, y_i, K)$$ be the optimal DNF rule set for loss function $\ell$ using candidate rule set $\mathcal{K}$. 
There does \textbf{not} exist a global constant $\Psi \in [1,\infty)$ such that
$$
\Psi \sum_{i=1}^n \ell_{01}(\mathbf{X}_i, y_i, K^*_{\ell_{01}}) \geq \sum_{i=1}^n \ell_{01}(\mathbf{X}_i, y_i, K^*_{\ell_{h}})
$$
for all candidate rule sets $\mathcal{K}$, data $\mathcal{D}$. 
\end{theorem}


Even though Hamming loss can theoretically lead to arbitrarily worse performance on the 0-1 classification problem, we decide to use it in our formulation out of practicality. Its compact formulation can be solved more efficiently than that of the 0-1 formulation, leading to a more computationally tractable framework. Empirically, models trained with Hamming loss perform comparably to those trained with 0-1 loss (as discussed in Section \ref{sec:HammingEmp}).

\subsection{Master Integer Programming Formulation}

In addition to the formulation presented in Section~\ref{sec:ham}, we add an additional constraint on the {\em complexity} of the rule set both to prevent over-fitting and to control interpretability. For concreteness, we define the complexity of a rule to be one plus the number of conditions in the rule; other affine functions 
can be handled equally well. We denote the complexity of rule $k$ by $c_k$. The total complexity of a rule set is defined as the sum of the complexities of its rules.
It is possible to include an additional term in the objective function to penalize complexity but we instead explicitly impose an upper bound on complexity of the rule set by a given parameter $C$ as it offers better control in applications where interpretable rules are preferred. Clearly one can use both a constraint and a penalty term.

Building upon the previous Hamming loss IP formulation \eqref{obj:ham}---\eqref{const:hamBin}, the full formulation for selecting an optimal rule set becomes
	
	\begin{align}
	z_{MIP} = ~\textbf{min} ~~\sum_{i \in \mathcal{P}} \zeta_i + \sum_{i \in \mathcal{N}} \sum_{k \in \mathcal{K}_i} w_k \label{Mobj}\\
	\textbf{s.t.}~\qquad\qquad  \zeta_i + \sum_{k \in \mathcal{K}_i} w_k &\geq 1 ~~~~~i \in \mathcal{P} \label{MmisP}\\
	\sum_{k \in \mathcal{K}} c_k w_k &\leq C \label{Mcomplex}\\
	w\in \{0,1\}^{\abs{\mathcal{K}}},&~ \zeta\in \{0,1\}^{\abs{\mathcal{P}}}  \label{Mbinary}
	\end{align}
Constraint \eqref{Mcomplex} provides the bound on complexity of the final rule set. We call the integer program (\ref{Mobj})-(\ref{Mbinary}) the Master Integer Program (MIP), and its associated linear relaxation the Master LP (MLP) (obtained by replacing \eqref{Mbinary} with $w\in [0,1]^{\abs{\mathcal{K}}},~ \zeta\in [0,1]^{\abs{\mathcal{P}}}$). We denote the optimal values of the MIP and MLP by $z_{MIP}$ and $z_{MLP}$ respectively. 

\cl{While the MIP \eqref{Mobj}---\eqref{Mbinary} optimizes for Hamming Loss, we also employed a minor enhancement described in detail in Appendix \ref{sec:solPool} that uses the 0-1 Loss. This enhancement, which we call \textit{Pool Select}, uses the fact that the mixed integer programming solver retains a set of feasible integer solutions encountered while solving the restricted MIP. Selecting the best of these feasible solutions with respect to 0-1 Loss, even while the solver is optimizing for Hamming Loss, leads to a small increase in performance.
  }  

\section{Fairness}\label{sec:fairness}
We now consider the case when each data point also has an associated group (or protected feature) $g_i\in\mathcal{G}$ where $\mathcal{G}$ is a given discrete set. Some common examples of protected features are gender, race, and social class. For each group $g\in \G$ we denote the data points that have the protected feature $g$ with
	\begin{equation*}\mathcal{I}_g = \{i\in I : g_i = g \}\end{equation*}
and let $\mathcal{P}_g=\mathcal{P} \cap \mathcal{I}_g$ and $ \mathcal{N}_g=\mathcal{N} \cap \mathcal{I}_g$. For simplicity, we describe the setting where $\G=\{1,2\}$ for the remainder of the paper and note that extending it to multiple groups is straightforward and simply adds constraints that scale linearly with the number of groups. In this setting, the classifier is not only required to predict the labels well, but it is also required to treat each group fairly. In particular, we will focus on two measures of group fairness related to classification parity: equality of opportunity, and equalized odds.

\subsection{Equality of Opportunity} 
Equality of opportunity requires the Type II error rate (i.e., false negative rate) to be equal across groups by enforcing the following condition \citep{hardt2016equality}: 	
	\begin{equation} 
    {\mathbb P} (\hat{y}(\textbf{X}) \neq Y  \mid Y = 1, G = g) = {\mathbb P} (\hat{y}(\textbf{X}) \neq Y \mid Y = 1)  \label{eq:EOO}
    \end{equation}
 for all $ g \in \mathcal{G} $\cl{, where the probability ${\mathbb P}$ is taken with respect to the underlying data distribution. Since the true distribution is unknown we approximate it with the empirical distribution for the training data.} Condition~\eqref{eq:EOO} requires the false negative rate of the classifier to be independent of the group the data point belongs to. This fairness criterion may be desirable 
when there is a much larger societal cost to false negatives than false positives, making it particularly well-suited for applications such as loan approval or hiring decisions. For example, in the context of hiring, it ensures qualified candidates would be offered a job with equal probability, independent of their group membership (ex. male/female). 
    
In a practical setting, it is unrealistic to expect to find classifiers that can satisfy the above criterion exactly. In fact, in most non-trivial applications strong adherence to fairness criteria comes at a large cost to accuracy \citep{kleinberg2016inherent} and therefore one needs to consider how much these conditions are violated as a measure of fairness. In the context of equality of opportunity, the maximum violation can be used to measure the {\em unfairness} of the classifier by the following expression:
\smallskip
\begin{align*}   \Delta(\hat{y})=  \max_{g, g' \in \mathcal{G}} \Big|\Pr(\hat{y}(\textbf{X})& \neq Y \mid Y = 1, G = g) - \Pr(\hat{y}(\textbf{X}) \neq Y \mid Y = 1, G = g') \Big|
   \end{align*}

When training the classifier $\hat{y}$, one can then use \cl{the sample estimate of} $  \Delta(\hat{y})$ in the objective function as a penalty term or can explicitly require a constraint of the form $\Delta(\hat{y})\le\epsilon$ to be satisfied by the classifier $\hat{y}$. We will focus on the latter case as it allows for explicit control over tolerable unfairness.

To incorporate the equality of opportunity criterion into the MIP, we bound the difference in the false negative rates between groups linearly as follows: 
	\begin{align}
	\frac{1}{\abs{\mathcal{P}_1}}\sum_{i \in  \mathcal{P}_1 } \zeta_i 
	        - \frac{1}{\abs{\mathcal{P}_2}}\sum_{i \in  \mathcal{P}_2} \zeta_i &\leq \epsilon_1 \label{const:eo1}\\
	 \frac{1}{\abs{\mathcal{P}_2}}\sum_{i \in  \mathcal{P}_2} \zeta_i 
	        - \frac{1}{\abs{\mathcal{P}_1}}\sum_{i \in  \mathcal{P}_1} \zeta_i&\leq \epsilon_1 \label{const:eo2}
	\end{align}
Constraints \eqref{const:eo1} and \eqref{const:eo2} bound the maximum allowed unfairness, denoted by $\Delta$ above, by a specified constant $\epsilon_1\ge 0$ that corresponds to an acceptable  level of unfairness. If $\epsilon_1$ is chosen to be 0, then the fairness constraint is imposed strictly.

\subsection{Equalized Odds} 
A stricter condition on the classifier is to require that both the Type I and Type II error rates are equal across groups \citep{hardt2016equality}.  
This requirement prevents possible trade-off between false negative and false positive errors across groups and can be seen as a generalization of the equality of opportunity criterion to include false positives. 
To achieve equalized odds, together with equation \eqref{eq:EOO}, the following condition is also enforced:
   	\begin{equation*} 
    \Pr(\hat{y}(\textbf{X}) \neq Y ~ \mid Y = -1, G = g) = \Pr(\hat{y}(\textbf{X}) \neq Y ~ \mid Y = -1)  
    \end{equation*}
for all $ g\in \mathcal{G}$. 

Similar to our use of Hamming loss as a proxy for 0-1 loss for the negative class, we use it as a proxy for equalized odds. Specifically, instead of bounding the difference in false positive rates between groups we bound the difference in the Hamming loss terms for the negative class. We call this Hamming loss proxy for equalized odds \textit{Hamming Equalized Odds}. Thus in conjunction with constraints \eqref{const:eo1} and \eqref{const:eo2}, we also include the following constraints in the formulation:

  	\begin{align}
    \frac{1}{\abs{\mathcal{N}_1}}\sum_{i \in \mathcal{N}_1} \sum_{k \in \mathcal{K}_i} w_k  - \frac{1}{\abs{\mathcal{N}_2}}\sum_{i \in \mathcal{N}_2} \sum_{k \in \mathcal{K}_i} w_k &\leq \epsilon_2\label{const:eo3}\\
     \frac{1}{\abs{\mathcal{N}_2}}\sum_{i \in \mathcal{N}_2} \sum_{k \in \mathcal{K}_i} w_k  - \frac{1}{\abs{\mathcal{N}_1}}\sum_{i \in \mathcal{N}_1} \sum_{k \in \mathcal{K}_i} w_k &\leq \epsilon_2,\label{const:eo4}
	\end{align}
where  $\epsilon_2\ge 0$ is a given constant. The tolerance parameter $\epsilon_2$ in  \eqref{const:eo3} and \eqref{const:eo4}  can be set equal to $\epsilon_1$ in  \eqref{const:eo1} and \eqref{const:eo2}, or, alternatively, they can be chosen separately. Note that we normalize the Hamming loss terms to account for the difference in group sizes and positive response rates between groups. \cl{Unfortunately, similar to using Hamming loss in the objective, Theorem \ref{theorem:heo_perf} shows that the Hamming loss proxy for false positives can lead to arbitrarily unfair classifiers with respect to the true equalized odds criterion (the proof can be found in Appendix \ref{app:pf_heo_perf}). However once again, the Hamming loss proxy performs well empirically and generates classifiers that meet the true fairness constraint.

\begin{theorem}[Hamming Equalized Odds Proxy] \label{theorem:heo_perf}
A rule set satisfying Hamming Equalized Odds can have an arbitrarily large gap in the false positive rate between groups. Formally, for a data set $\mathcal{D} = [(\mathbf{X}_i, y_i)]_1^n$ where $\mathbf{X}_i \in \{0,1\}^d$ with group membership $\mathcal{G}$, let 
$$\Delta(K)= \max_{g,g' \in \mathcal{G}} \Big| \frac{1}{|\mathcal{N}_{g}|}\sum_{i \in \mathcal{N}_{g}} \mathbbm{I}(|K \cap {\mathcal{K}}_i| > 0) -  \frac{1}{|\mathcal{N}_{g'}|}\sum_{i \in \mathcal{N}_{g'}} \mathbbm{I}(|K \cap {\mathcal{K}}_i| > 0) \Big|$$ 
be the maximum gap in false positive rates between groups for a DNF rule set K. 

Let $K^*$ be a DNF rule set that satisfies Hamming Equalized Odds (i.e., constraints \eqref{const:eo1}, \eqref{const:eo2}, \eqref{const:eo3} and \eqref{const:eo4}) with parameter $\epsilon$. There does \textbf{not} exist a global constant $\Psi \in [1,\infty)$ such that
$$
\Delta(K^*) \leq \Psi \epsilon
$$
for all candidate rule sets $\mathcal{K}$, data $\mathcal{D}$, and $\epsilon$.
\end{theorem}
 }

\mysec{Other fairness metrics} 	
While we restrict our focus to equality of opportunity and equalized odds, we note that our framework can be adapted for any notion of classification parity (i.e., balancing false positive rates, overall accuracy, or demographic parity). The only caveat is that for notions of fairness involving false positives our framework would use the Hamming loss term for false positives (similar to equalized odds). As discussed in Section~\ref{sec:cg}, to solve our formulation an initial set of rules ${K}$ are needed to make the problem feasible. With the fairness criteria presented above, the empty set of rules gives a trivial feasible solution (i.e., all points in ${\cal P}$ are misclassified, which is fair with respect to equality of opportunity and equalized odds). However, this may no longer be the case for other notions of fairness, such as ensuring similar Hamming loss or accuracy between the two groups. In such cases, a two stage approach could be used to first generate a set of candidate rules that are feasible for a given fairness criteria prior to optimizing for accuracy.

\subsection{Base Formulation for Fairness Setting} 


In the IP formulation presented in Section~\ref{sec:class}, there is no need to track true positives (i.e., ensure $\zeta_i = 0$ when point $i \in {\cal P}$ is correctly classified) as $\zeta_i = 0$ in any optimal solutions provided that $\sum_{k \in \mathcal{K}_i} w_k\geq 1$. However, this is not the case when fairness constraints are added to the formulation. For example, a data point $i$ could be correctly classified but setting $\zeta_i = 1$ preserves the feasibility of the solution with respect to the fairness constraint. To modify our formulation for the fairness setting, we add additional constraints to correctly track the error for the positive class ${\cal P}$. Thus for the fairness setting we add the following constraint to IP formulation (\ref{Mobj})-(\ref{Mbinary}):
	\begin{align}
	C\zeta_i + \sum_{k \in \mathcal{K}_i} \beta_k w_k &\leq C ~~~~~i \in \mathcal{P} \label{MmisP2}
	\end{align}

Constraint \eqref{MmisP2} ensures that $\zeta_i = 0$ if any rules satisfied by $i \in \mathcal{P}$ are selected. Here $\beta_k > 0$ is any set of coefficients such that $\sum_{k \in \mathcal{K}_i} \beta_k w_k \leq C$ for all feasible solutions of problem \eqref{Mobj}-\eqref{Mbinary}. A natural choice would be to set $\beta_k = c_k$ to recover constraint \eqref{Mcomplex}, however this leads to computational challenges during column generation (see Appendix~\ref{app:cycling} for a detailed discussion). Instead we set $\beta_k = 2$ as  $c_k\ge2$ for all $k\in\K$.

\section{Column Generation}\label{sec:cg}

It is not practical to solve the MIP \eqref{Mobj}-\eqref{Mbinary} using standard branch-and-bound techniques	\citep{LandDoig} as it would require enumerating exponentially many rules. To overcome this problem, we solve the LP relaxation of MIP using the column generation technique \citep{GilmoreGomory,IPref} without explicitly enumerating all possible rules. 
Once we solve the LP to optimality or near optimality, we then restrict our attention to the rules generated during the process and pick the best subset of these rules by  solving a restricted MIP. We note that it is possible to integrate the column generation technique with branch-and-bound to solve the MIP to provable optimality using the branch-and-price approach \citep{Nemhauser:1998}. However, this would be quite time consuming in practice. 

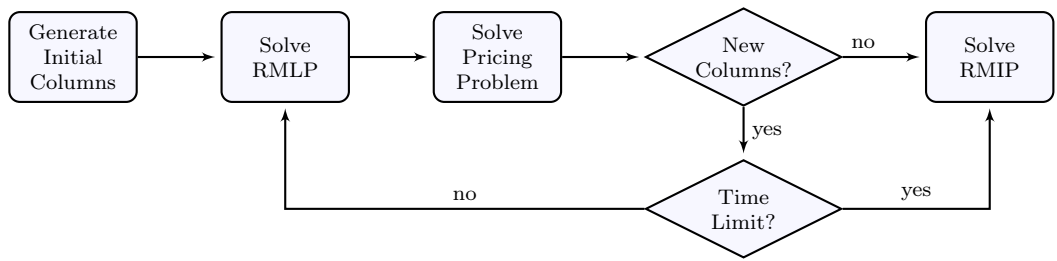
\begin{figure}[t]
\tikzset{every picture/.style={line width=0.75pt}} 

\begin{tikzpicture}\scriptsize
\tikzstyle{decision} = [diamond, aspect=2, draw=black, thick, fill=blue!3,	text width=4.5em, text badly centered, inner sep=1pt]
\tikzstyle{block} = [rectangle, draw=black, thick, fill=blue!3,	text width=5em, text centered, rounded corners, minimum height=4em]
\tikzstyle{line} = [draw, thick, -latex',shorten >=2pt];
\tikzstyle{cloud} = [draw=red, thick, ellipse,fill=red!20, minimum height=2em];

 \matrix [column sep=11mm,row sep=7mm]
	{
		\node [block] (init) {Generate Initial Columns}; &
		\node [block] (solve) {Solve RMLP}; &
		\node [block] (price) {Solve Pricing Problem}; &
		\node [decision] (new) {New Columns?}; &
		\node [block] (rmip) {Solve RMIP}; \\
		& &&		\node [decision] (timlim) {Time Limit?}; \\
	};
 
\tikzstyle{every path}=[line]
	\path (init) -- (solve);
	\path (solve) -- (price);
	\path (price) -- (new);
	\path (timlim) -| node [near start,above] {no} (solve);
	\path (timlim) -| node [near start,above] {yes} (rmip);
	\path (new) -- node [near start,above] {no} (rmip);	
	\path (new) -- node [midway,right] {yes} (timlim);
\end{tikzpicture}
\caption{\label{fig:cg_diagram} Flowchart of column generation procedure.}
\end{figure}

To solve the LP relaxation of the MIP, called the MLP, we start with a possibly empty subset $\hat\K\subset\K$ of all candidate rules and solve an LP restricted to the variables associated with these rules only. Once this small LP is solved, we use its optimal dual solution to identify a missing variable (rule) that has a negative reduced cost. The search for such a rule is called the {\em Pricing Problem} and in our case this can be done by solving a separate integer program.
If a rule with a negative reduced cost is found, then  $\hat \K$ is augmented with the rule and the LP is solved again. This process is repeated until no such rule can be found. This mimics how LPs are solved to optimality in practice using the revised simplex method. For large problems, even solving the MLP to optimality is not always computationally feasible. Out of practicality, we put an overall time limit on the column generation process and terminate without a certificate of optimality if the limit is reached. \cl{In Figure \ref{fig:cg_diagram}, we present a flowchart describing the column generation procedure.}

\subsection{Pricing Problem: Base Formulation without Fairness}
Given a subset of rules $\hat\K\subset\K$, let the RMLP be the restriction of MLP to the rules in $\hat \K$. In other words, RMLP is the restriction of MLP where all variables $w_k$ associated with $k\in\K\setminus\hat\K$ are fixed to 0. Let $(\mu,\lambda)$ be an optimal {\em dual} solution to RMLP, where dual variables $\mu \in \mathbb{R}_+^{\abs{\mathcal{P}}}$, $\lambda \in \mathbb{R}_+$ are associated with constraints \eqref{MmisP} and \eqref{Mcomplex} respectively. Using this dual solution, the reduced cost of a variable $w_k$ associated with a rule $k\notin\hat\K$ can be expressed as
\begin{equation}
\hat \rho_k =  \sum_{i \in \mathcal{N}} \mathbbm{I}(k \in {\cal K}_i)
-\sum_{i \in \mathcal{P}}  \mu_i  \mathbbm{I}(k \in {\cal K}_i) +\lambda  c_k  \label{eq:redcost}
\end{equation}
where the first term simply counts the number of data points $i\in{\cal N}$ that satisfy the rule $k$.
If there exists a $k\in\K\setminus\hat\K$ with $\hat \rho_k<0,$ then including variable $w_k$ in RMLP has the potential of decreasing the objective function of the LP. Also note that $\hat \rho_k\geq0$ for all $k\in\hat\K$ as the dual solution at hand is optimal for RMLP.

We can now formulate an integer program to find a rule $k\in\K$ with the minimum reduced cost $\hat \rho_k$.  Let $\cJ =\{ 1,\dots,d\} $ be the set of binary-valued features and $X_{ij}$ be the $j$-th feature value for data point $\textbf{X}_i$.
Remember that  a decision rule corresponds to a subset of the binary features ${\cal J}$  and classifies a data point with a positive response if the point has all the features selected by the rule. Let $S_i=\{j\in{\cal J}: X_{ij} = 0\}$ correspond to the zero-valued features 
in sample $i\in{\cal P}\cup {\cal N}$.
Let variable $z_j\in\{0,1\}$ for  $j\in {\cal J}$  denote if the rule includes feature $j$ and let variable  $\delta_i\in\{0,1\}$ for $i\in I$ denote if the rule is satisfied by data point $i$.
Using these variables, the complexity of a rule can be computed as $(1 + \sum_{j \in {\cal J}} z_j)$ and the reduced cost of the rule becomes

    \begin{equation} \label{eq:pricingObj}
     \enskip \sum_{i \in \mathcal{N}} \delta_i - \sum_{i \in \mathcal{P}} \mu_i \delta_i + 
     \lambda\Big(1 + \sum_{j \in {\cal J}} z_j\Big)  .
    \end{equation}


The full Pricing Problem thus simply minimizes (\ref{eq:pricingObj}) subject to the constraints:
    \begin{align}
	D\delta_i + \sum_{j \in S_i} z_j &\leq D ~~~~~ i \in I^{-}\quad \label{PmisP}\\
	\delta_i +  \sum_{j \in S_i} z_j &\geq  1~~~~~i \in I^{+}  \quad\label{PmisZ}\\
	\sum_{j \in {\cal J}} z_j &\leq D \quad \label{Pcomplex}\\
	z\in \{0,1\}^{\abs{{\cal J}}},&~ \delta\in \{0,1\}^{\abs{I}}\label{Pint}\qquad
	\end{align}
where the set $I^-\subseteq I$ contains the indices of $\delta_i$ variables that have a negative  coefficient in the objective (i.e., $-\mu_i < 0$, $i \in {\cal P}$), and $I^+=I\setminus I^-$. We denote the optimal value to the Pricing Problem by $z_{CG}$. Constraints \eqref{PmisP} and \eqref{PmisZ} ensure that $\delta_i$ accurately reflects whether the new rule classifies data point $i$ with a positive label. Constraint \eqref{Pcomplex} puts an explicit bound on the complexity of any rule using the parameter $D$. This individual rule complexity constraint can be set independently of $C$ in the master problem or can be relaxed by setting it to $C - 1$. 

\cl{We also note that a naive implementation of the column generation procedure as presented above can be susceptible to \textit{cycling}, the phenomenon where the Pricing Problem repeatedly produces the same column. Cycling can stall the column generation procedure, preventing it from finding new columns, and was observed in initial experiments with this framework. An example of the phenomenon and strategies to mitigate the problem are included in Appendix \ref{app:cycling}.} 

We next formally show that the Pricing Problem is a difficult optimization problem. The proof follows from a reduction of the minimum vertex cover (MVC) problem and is included in Appendix \ref{app:pf_np_pricing}.

\begin{theorem} \label{theorem:np_pricing}
The Pricing Problem \eqref{eq:pricingObj}-\eqref{Pint} is NP Hard.		
\end{theorem}

\cl{Given the hardness of the Pricing Problem it can be computationally intractable to solve the MLP to optimality using column generation.} For {\em small} data sets, defined loosely as having less than a couple of thousand samples and less than a few hundred 
binary (binarized) features, it is still computationally feasible to employ column generation with the Pricing Problem IP formulation. However, to handle larger data sets within a time limit of 10 or 20 minutes, one has to sacrifice the optimality guarantees of the framework. We next describe our computational approach to deal with larger data sets, which can be seen as an optimization-based heuristic. 

We call a data set {\em medium} if it has more than a couple of thousand samples but less than a few hundred binary features. We call it {\em large} if it has many thousands of samples and more than several hundred binary features. The separation of data sets into small, medium and large is done based on empirical experiments to improve the likelihood that the Pricing Problem can produce negative reduced cost solutions in practice.

For medium and large data sets, the number of non-zeros in the Pricing Problem (defined as the sum of the numbers of variables appearing in the constraints of the formulation) is at least 100,000 and solving this  integer problem to optimality in a reasonable amount of time is not always feasible. To deal with this practical issue, we terminate the  Pricing Problem if a fixed time limit is exceeded. 
If the solver can find one (or more) negative reduced cost rules within the time limit, we add all the negative reduced cost rules returned by the solver to the RMLP. As long as one such rule (variable) is obtained, the column generation process continues. For large data sets, the solver typically fails to find any rule with negative reduced cost within the time limit. In this case,  we sub-sample both the training data points and potential features to have on average 2000 rows and 100000 non-zeros in the Pricing Problem. We then solve this reduced version of the Pricing Problem.

If the full  Pricing Problem can be solved to optimality within the time limit and there are no negative reduced cost solutions, then the current RMLP solves the MLP to optimality. However, if it terminates without a negative reduced cost solution due to the time limit, or, when we use sub-sampling, then we do not have a certificate of optimality. In this case, we employ a fast heuristic algorithm, detailed in Section~\ref{sec:heuristic}, to continue to search for negative reduced cost solutions and continue the process.

\subsection{Pricing Heuristic}  \label{sec:heuristic}
To obtain good solutions to the Pricing Problem without solving the integer program we use the following heuristic which employs beam search, starting with a list of single-feature rules. For the best single feature rules, it attempts to expand each by adding another feature. This process is then repeated for the best two-feature rules and so on, till we obtain up to five-feature rules.
The beam widths we use for one-feature rules up to five-feature rules are 50, 20, 6, 6, 5 respectively.

Consider the prediction function $\hat y$ defined by a single rule. Then we can rewrite the equation for the reduced cost (\ref{eq:redcost}) associated with the rule as
\begin{equation}\label{rule-red} \hat\rho = \sum_{i \in \cal N} \hat y ({\bf X_i}) - \sum_{i \in \cal P} \mu_i\hat y({\bf X_i}) + \lambda \hat c.
\end{equation}
The goal is to find a rule for which $\hat \rho$ is as negative as possible.
Now assume that a rule consists of a single feature, say $\ell$. If we create a new rule by adding another feature to the rule with associated prediction function $\bar y$, complexity $\bar c$ and reduced cost $\bar \rho$,  then we have $\bar y({\bf X_i}) \leq \hat y({\bf X_i})$ for all $i \in \cal P \cup \cal N$. 
 Therefore, we have
\[ \sum_{i \in \cal P} \mu_i \bar y({\bf X_i}) \leq \sum_{i \in \cal P} \mu_i \hat y({\bf X_i}) \mbox{~ and }  \sum_{i \in \cal N} \bar y({\bf X_i}) \leq \sum_{i \in \cal N} \hat y(\bf X_i).\]


But we also have $\bar c = 1+ \hat c$. This means that if $\lambda \hat c - \sum_{i \in \cal P} \mu_i\hat y({\bf X_i})$ is positive, it is not possible to create a rule containing the feature $\ell$ with negative reduced cost. We use this basic idea repeatedly.
At the beginning, we mark all such features and rule them out for inclusion in any rule. Then we take the remaining features and create single-feature rules, and calculate (\ref{rule-red}) for each. We sort this list, and keep the best.
Now for a given rule $r$, let $\cal P'$ and $\cal N'$ be calculated as before. If for any feature $\ell$ not in the rule, we have $\lambda (\hat c + 1) - \sum_{i \in \cal P'}\mu_i \bar y({\bf X}_i) > 0$, then no rule that contains the features in $r$ and the feature $\ell$ can yield a negative reduced cost. When we consider extended a rule by adding features, we create a list of features that cannot be added to the rule by the above criterion. This list is copied to {\em every new rule} containing $r$ and additional features are added to the list.

Finally, once we have a list of negative reduced cost rules, we return a ``diverse" subset of them by limiting the number of rules that have the same first feature (assuming features have an ordering) and the number of rules that have the same first two features and so on.

\subsection{Pricing Problem: Fairness Setting} 
We now extend the Pricing Problem formulation to the fairness setting under both equality of opportunity and equalized odds. In this setting, the master problem is augmented with additional constraints that factor into the objective of the Pricing Problem.

\mysec{Equality of opportunity} 
Under the equality of opportunity criterion, the master problem is augmented with constraints \eqref{const:eo1}, \eqref{const:eo2}, and \eqref{MmisP2}. Let $(\mu,\alpha,\lambda,\gamma^1,\gamma^2)$ be an optimal {\em dual} solution to RMLP, where variables $\gamma_1, \gamma_2,\alpha $ are associated with constraints \eqref{const:eo1},  \eqref{const:eo2}, and \eqref{MmisP2}. Using this dual solution, the reduced cost of a variable $w_k$ associated with a rule $k\notin\hat\K$ can be expressed as
\begin{equation}
\hat \rho_k =  \sum_{i \in \mathcal{N}} \mathbbm{I}(k \in {\cal K}_i)
-\sum_{i \in \mathcal{P}}  \mu_i \mathbbm{I}(k \in {\cal K}_i)
+ \sum_{i \in \mathcal{P}} 2 \alpha_i \mathbbm{I}(k \in {\cal K}_i) +\lambda  c_k  \label{eq:redcosteqop}
\end{equation}
Note that variable $w_k$ does not appear in constraints \eqref{const:eo1} or  \eqref{const:eo2}  in  RMLP and consequently \eqref{eq:redcosteqop} does not involve variables $\gamma^1$ or $\gamma^2$. The full Pricing Problem thus becomes
    \begin{align*}
	~z_{CG} = ~~&\textbf{min} ~~  
	  \sum_{i \in \mathcal{N}} \delta_i + \sum_{i \in \mathcal{P}} (2 \alpha - \mu_i) \delta_i + 
     \lambda\left(1 + \sum_{j \in J} z_j\right)  
	\\	&\textbf{s.t.} ~~\eqref{PmisP}-\eqref{Pint},
	\end{align*}
 where the set $I^-\subseteq I$ contains data points $i \in {\cal P}$ such that $2\alpha - \mu_i < 0$, and $I^+=I\setminus I^-$. \cl{Note that while the dual values $\gamma_1, \gamma_2$ for the fairness constraints do not \textit{explicitly} appear in the Pricing Problem, this does not mean that the fairness constraints have no impact on the column generation procedure. The inclusion of constraints \eqref{const:eo1} and  \eqref{const:eo2} may lead to different optimal solutions to the RMLP and by extension different values for the optimal dual variables $\mu, \alpha, \lambda$ involved in the pricing problem. In other words, the inclusion of the fairness constraints implicitly changes the objective of the pricing problem and the rules returned by it.}


\mysec{Equalized odds} 	
In this case we further augment the RMLP from the Equality of Opportunity setting with constraints \eqref{const:eo3} and  \eqref{const:eo4}. Note that unlike  \eqref{const:eo1} and  \eqref{const:eo2}, constraints \eqref{const:eo3} and  \eqref{const:eo4} do involve variables $w_k$.
Let $(\mu,\alpha,\lambda,\gamma^1,\gamma^2,\gamma^3,\gamma^4)$ be an optimal {\ dual} solution to RMLP, where variables  $\gamma^3$ and $\gamma^4$ are associated with fairness constraints \eqref{const:eo3} and  \eqref{const:eo4}, respectively.
Using this dual solution, the reduced cost of a variable is similar to the expression in \eqref{eq:redcosteqop}, except it has the following 4 additional terms:

\begin{align*}
  \sum_{i \in \mathcal{N}_1}  \frac{\gamma_3}{\abs{\mathcal{N}_1}} \mathbbm{I}(k \in {\cal K}_i)
- \sum_{i \in \mathcal{N}_1} \ \frac{\gamma_4}{\abs{\mathcal{N}_1}}\mathbbm{I}(k \in {\cal K}_i)
- \sum_{i \in \mathcal{N}_2}  \frac{\gamma_3}{\abs{\mathcal{N}_2}}\mathbbm{I}(k \in {\cal K}_i)
+ \sum_{i \in \mathcal{N}_2} \frac{\gamma_4}{\abs{\mathcal{N}_2}}\mathbbm{I}(k \in {\cal K}_i)
\label{eq:redcostH}
\end{align*}

Consequently, the 
Pricing Problem becomes
    \begin{align*}
	~z_{CG} = &\textbf{min} ~~  
	 \left(1+\frac{\gamma_3 - \gamma_4}{\abs{\mathcal{N}_1}}\right)\sum_{i \in \mathcal{N}_1} \delta_i 
	+ \left(1+\frac{\gamma_4 - \gamma_3}{\abs{\mathcal{N}_2}}\right)\sum_{i \in \mathcal{N}_2} \delta_i   \\
	&+ \sum_{i \in \mathcal{P}} (2\alpha - \mu_i) \delta_i  
	+ \lambda\left(1 + \sum_{j \in J} z_j\right) 
	\\	&\textbf{s.t.} ~~\eqref{PmisP}-\eqref{Pint},
	\end{align*}
\noindent
where the set $I^-$ contains data points $i \in {\cal P}$ such that $2\alpha - \mu_i < 0$, $i \in {\cal N}_1$ such that $(1+\frac{\gamma_3 - \gamma_4}{\abs{\mathcal{N}_1}}) < 0$, and $i \in {\cal N}_2$ such that $(1+\frac{\gamma_4 - \gamma_3}{\abs{\mathcal{N}_2}}) < 0$.

\cl{Note that the Pricing Problem heuristic introduced in Section \ref{sec:heuristic} can be adapted to the fairness setting by simply changing the condition for which a feature $\ell$ is no longer considered to extend a rule $r$. Instead of the original condition $\lambda (\hat{c}+1) - \sum_{i \in {\cal P}} \mu_i \bar{y}(\textbf{X}_i) > 0 $, we use  $\lambda (\hat{c}+1) + \sum_{i \in {\cal P}}(2\alpha_i - \mu_i) \bar{y}(\textbf{X}_i) > 0 $ to incorporate dual values coming from the additional fairness constraint \eqref{MmisP2}. The adjustment is the same for both the equality of opportunity and equalized odds formulations as the additional constraints for the latter model only impact the reduced cost formula with respect to data points in ${\cal N}$ which does not appear in the condition.}

\subsection{Optimality Guarantees and Bounds} \label{sec:opt_guar}

\cl{
When the column generation framework described above is repeated until $z_{CG}\ge 0$, none of the variables missing from the RMLP have a negative reduced cost and the optimal solution of the MLP and the RMLP coincide.
In addition,  if the optimal solution of the RMLP turns out to be integral, then it is also an optimal solution to the MIP and therefore the MIP is solved to optimality.
If the optimal solution of the RMLP is fractional, then one may have to use column generation within an enumeration framework to solve MIP to optimality. This approach is called {\em branch-and-price} \citep{Nemhauser:1998} and is quite computationally intensive. However, even when the optimal solution to the MLP is fractional or the MLP is not solved to optimality, the following proposition gives a valid lower bound for the value of the MIP (the proof can be found in Appendix \ref{app:lb_proof}).

\begin{proposition}[Lower Bound for $z_{MIP}$] \label{prop:lb}
At the conclusion of column generation, the following is a valid lower bound for the optimal value of the MIP:
$$
\Big \lceil z_{RMLP}+ \min((C/2)z_{CG}, 0) \Big \rceil
$$
where $z_{RMLP}$ is the objective value of the last RMLP solved to optimality, and $z_{CG}$ is the (lower bound for the) optimal value of the final Pricing Problem solved.
\end{proposition}

 This lower bound can be compared to the cost of any feasible solution to MIP.  If the latter equals the lower bound in Proposition \ref{prop:lb}, then, once again, MIP is solved to optimality.  As one example, 
a feasible solution to the MIP could be obtained by solving the \emph{Restricted MIP} obtained by imposing \eqref{Mbinary} on the  variables present in the RMLP.  More generally, any heuristic method can generate feasible solutions to the MIP.

}

\section{Numerical Evaluation}\label{sec:exp}

\cl{To demonstrate the performance of our approach we present a suite of numerical results. We start with describing the data sets, experimental setup, and computational environment that we used. In Section~\ref{sec:HammingEmp}, we present an empirical study aimed at understanding the effect of using Hamming loss as a proxy for 0-1 loss. We benchmark our approach against state-of-the-art algorithms in both the traditional classification setting (Section~\ref{sec:expClass}), and the fairness setting (Section~\ref{sec:expFair}). We also showcase sample rule sets in both sections to demonstrate their simplicity and intuitive appeal.}

\subsection{Experiment Details} \label{sec:exp_details} 

Evaluations were conducted on 15 classification data sets from the UCI repository \citep{dua2017} that have been used in recent works on rule set/Boolean classifiers \citep{malioutov2013,dash2014,su2016,wang2017}.  In addition, we used data from the FICO Explainable Machine Learning Challenge \citep{FICO2018}.  It contains $23$ numerical features of the credit history of $10,459$ individuals ($9871$ after removing records with all entries missing) for predicting repayment risk (good/bad).  The domain of financial services and the clear meanings of the features make this data set a good candidate for a rule set model. \cl{We benchmark the performance of our column generation approach with fairness constraints on three standard fair machine learning data sets: adult, compas, and default. For all three data sets we include the sensitive attribute as a feature for prediction as their exclusion is both not enough to ensure fairness and may lead to worse fairness outcomes  \citep[see][for a discussion]{corbettdavies2018measure}.}

For all the data sets presented in this Section we used standard “dummy”/“one-hot” coding to binarize categorical variables into multiple $X_j = x$ indicators, one for each category $x$, as well as their negations $X_j \neq x$. For numerical features, we compare the original feature values with a sequence of thresholds, again including negations (e.g., $X_j \leq 1$, $X_j \leq 2$ and $X_j > 1$, $X_j > 2$).  For these experiments, as also recommended in \cite{wang2017,su2016}, we use sample deciles as thresholds. The binarized data was used in all experiments and classifiers. More details about each data set and how missing and special values were treated can be found in Appendix \ref{app:data sets}  

Test performance on all data sets is estimated using $10$-fold stratified cross-validation (CV). Unless otherwise stated, all hyperparameters were tuned using nested 10-fold cross-validation. All tables in this section are grouped by problem size in increasing order (i.e., first section corresponds to small data sets, the second to medium and large data sets).

\subsection{Hamming loss}\label{sec:HammingEmp}
    To analyze the empirical performance of using Hamming loss instead of 0-1 loss we ran a sequence of experiments where we evaluated the 0-1 loss of rule sets trained under both objectives\cl{, each with a 300s time limit.} For each experiment we used the same pool of candidate rules generated by running Random Forests with different hyperparameters and extracting rules by looking at the leaf nodes of each tree following the procedure of \cite{birbil2020rule}. We emphasize that no column generation was performed in these experiments. We then solved the Restricted MIP under both objectives and evaluated accuracy of the resulting rule sets on both training and testing data. 
    
    \cl{For these experiments we ran the 0-1 formulation with the aggregated false positive constraints outlined in Appendix \ref{app:01aggregate} as it performed better empirically than the dis-aggregated version.} \cl{All results in this section were obtained on a personal computer with a 3 GHz processor and 16 GB of RAM. All the linear and integer programs were solved using Gurobi 9.0 \citep{gurobi}.}

\begin{figure}[tb]
    \centering
    \begin{subfigure}
      \centering
      \includegraphics[width=0.95\textwidth]{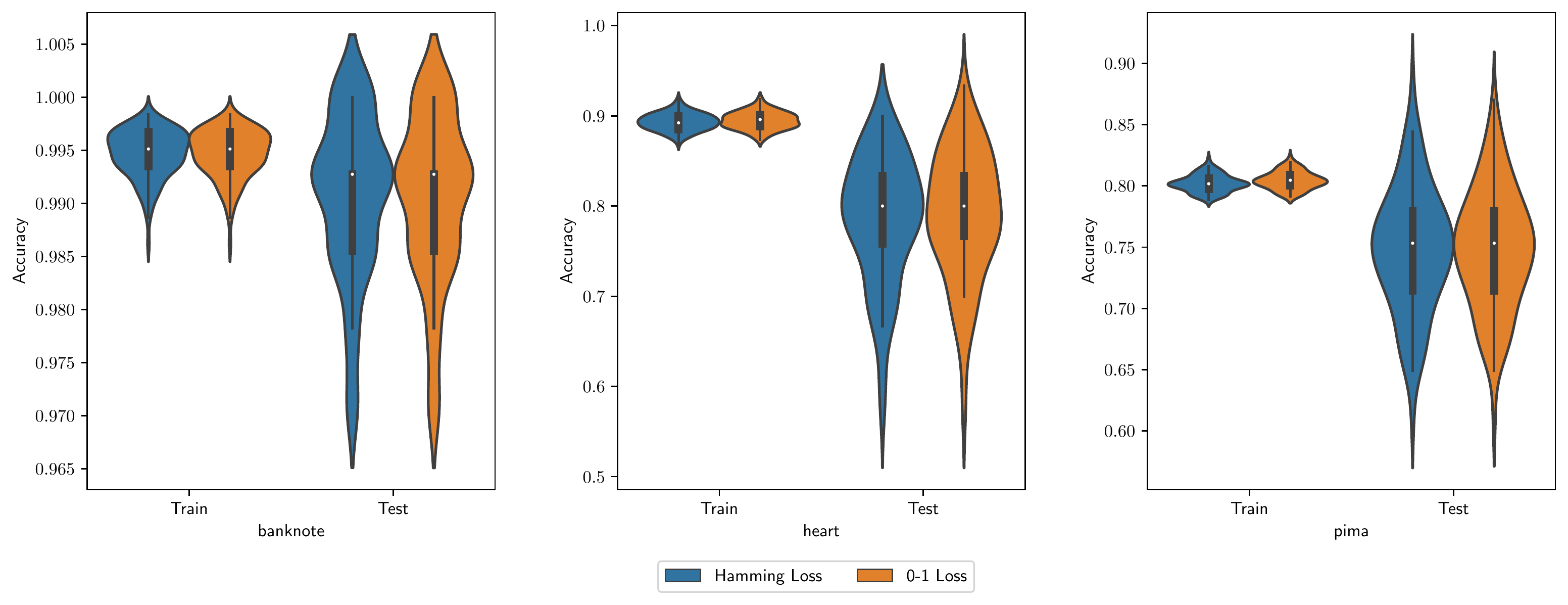}
    \end{subfigure}
  \caption {\label{Hamming_v_01} Distribution of train and test set accuracy of rule sets obtained under different objectives over 100 random splits of each data set.}
\end{figure}

   In Table \ref{HammingComputationTable}  we present the average computation time for both formulations together with accuracy of the resulting rule sets on training and testing data \cl{over 100 random train/test (90\%-10\%) splits of each data set} (recall that maximizing accuracy is equivalent to minimizing 0-1 loss). \cl{Each model used 5-fold cross-validation on the training data to tune $C$.} These experiments establish that Hamming loss is an effective proxy for 0-1 loss while being computationally efficient. Results for other data sets can be found in Appendix \ref{app:hamming_empirical} and show a similar trend. Figure \ref{Hamming_v_01} shows the distribution of accuracy for three different UCI machine learning data sets without fairness constraints. The left (right) side of each figure is a violin plot that shows the distribution of the accuracy values from minimizing Hamming loss (0-1 loss) during training over 100 random splits of the data set (each black dot represents an accuracy for a single split). We can see that Hamming loss and 0-1 loss have practically indistinguishable performance in terms of both train and test set accuracy.

\begin{table*}[tbh]
\centering\footnotesize
\caption{\label{HammingComputationTable} Performance of Hamming loss vs. 0-1 loss with respect to computation time, training, and test set accuracy (standard deviation \cl{over 100 random train/test splits} in parenthesis). Rows below the divider are for the fair setting under an equality of opportunity constraint of $\epsilon=0.025$.}
\setlength{\tabcolsep}{5pt} 
\begin{tabular}{l  c c c c c c }		\toprule
& \multicolumn{2}{c}{IP Solve Time (s)} & \multicolumn{2}{c}{Train Accuracy} & \multicolumn{2}{c}{Test Accuracy} \\
 &Hamming & 0-1 &Hamming & 0-1 & Hamming & 0-1\\\midrule
adult & 1.9 (0.8) & 278.5 (67.8) & 83.1 (0.0) & 83.0 (0.0) & 82.8 (0.0) & 82.8 (0.0) \\
bank-mkt & 0.7 (0.2) & 165.6 (126.5) & 90.2 (0.0) & 90.2 (0.0) & 90.0 (0.0) & 90.0 (0.0) \\
gas & 24.4 (16.4) & 190.7 (94.4) & 97.3 (0.0) & 97.3 (0.0) & 96.9 (0.0) & 97.0 (0.0) \\
FICO &  6.2 (5.8) &  214.2 (123.1) &  72.08 (0.9) &  72.08 (0.6) &  71.0 (1.8) &  71.2 (1.3)\\ 
magic & 7.0 (3.7) & 270.4 (54.7) & 84.1 (0.0) & 84.2 (0.0) & 83.4 (0.0) & 83.6 (0.0) \\
mushroom & 0.1 (0.0) & 0.1 (0.0) & 100.0 (0.0) & 100.0 (0.0) & 100.0 (0.0) & 100.0 (0.0) \\
musk & 0.6 (0.4) & 1.2 (0.6) & 96.8 (0.0) & 96.8 (0.0) & 95.9 (0.0) & 95.9 (0.0) \\ \midrule
adult  &  35.5 (48.6) & 546.0 (101) & 81.9 (0.3) & 82.1 (0.3) & 81.7 (0.3)& 81.7 (0.3)\\
compas & 4.0 (3.5) & 11.4 (6.5)  & 64.8 (0.3) & 65.1 (0.2) &  64.5 (0.4) & 64.4 (0.5) \\
default &  3.5 (0.8) &  12.3 (6.1) &  78.0 (0.0) & 78.0 (0.0) & 77.7 (0.0) & 77.7 (0.0) \\
\bottomrule
\end{tabular}%
\end{table*}%

\begin{figure}[!htb]
        \centering
        \begin{subfigure}
          \centering
          \includegraphics[width=0.32\textwidth]{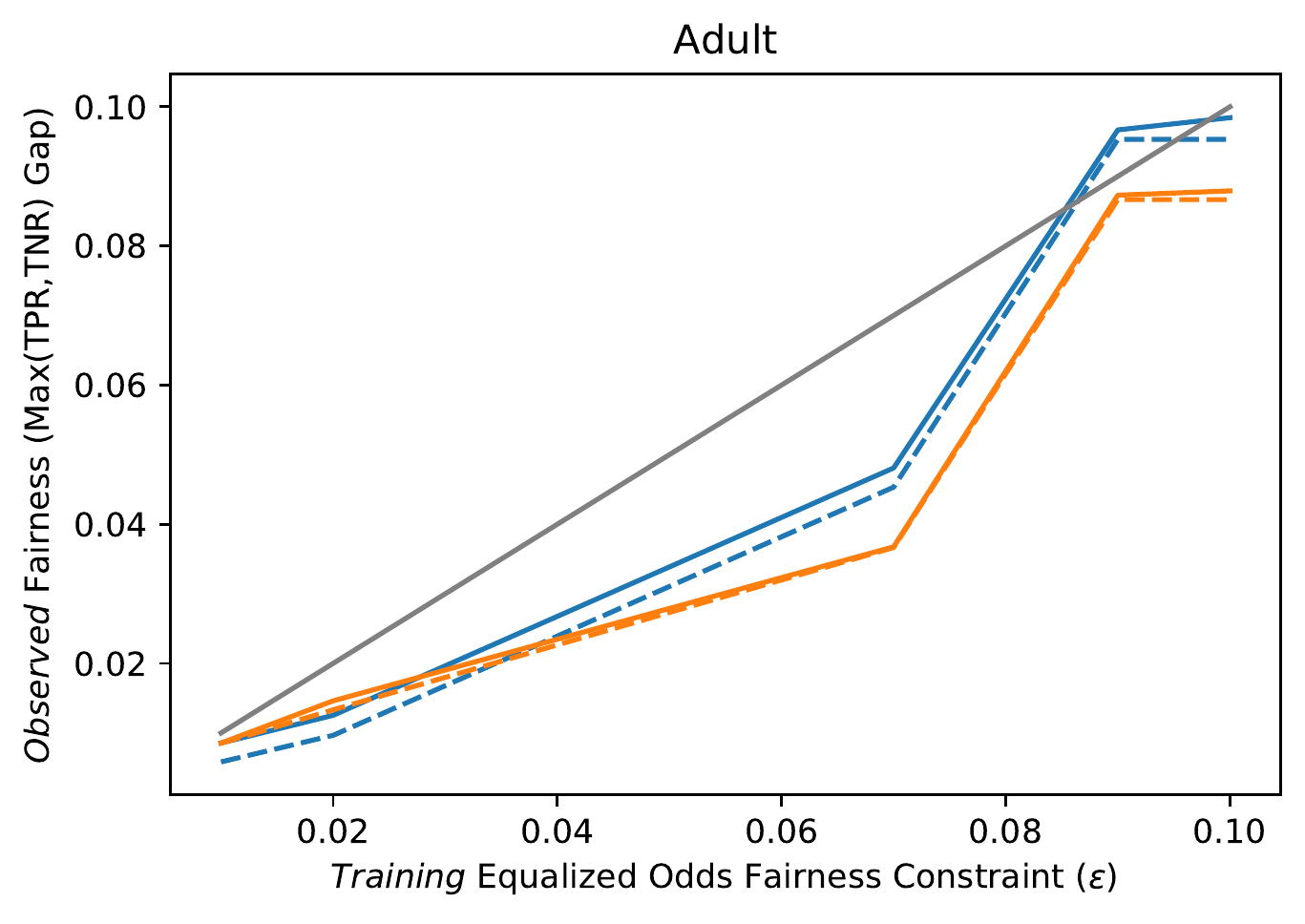}
        \end{subfigure}
        \begin{subfigure}
          \centering
          \includegraphics[width=0.32\textwidth]{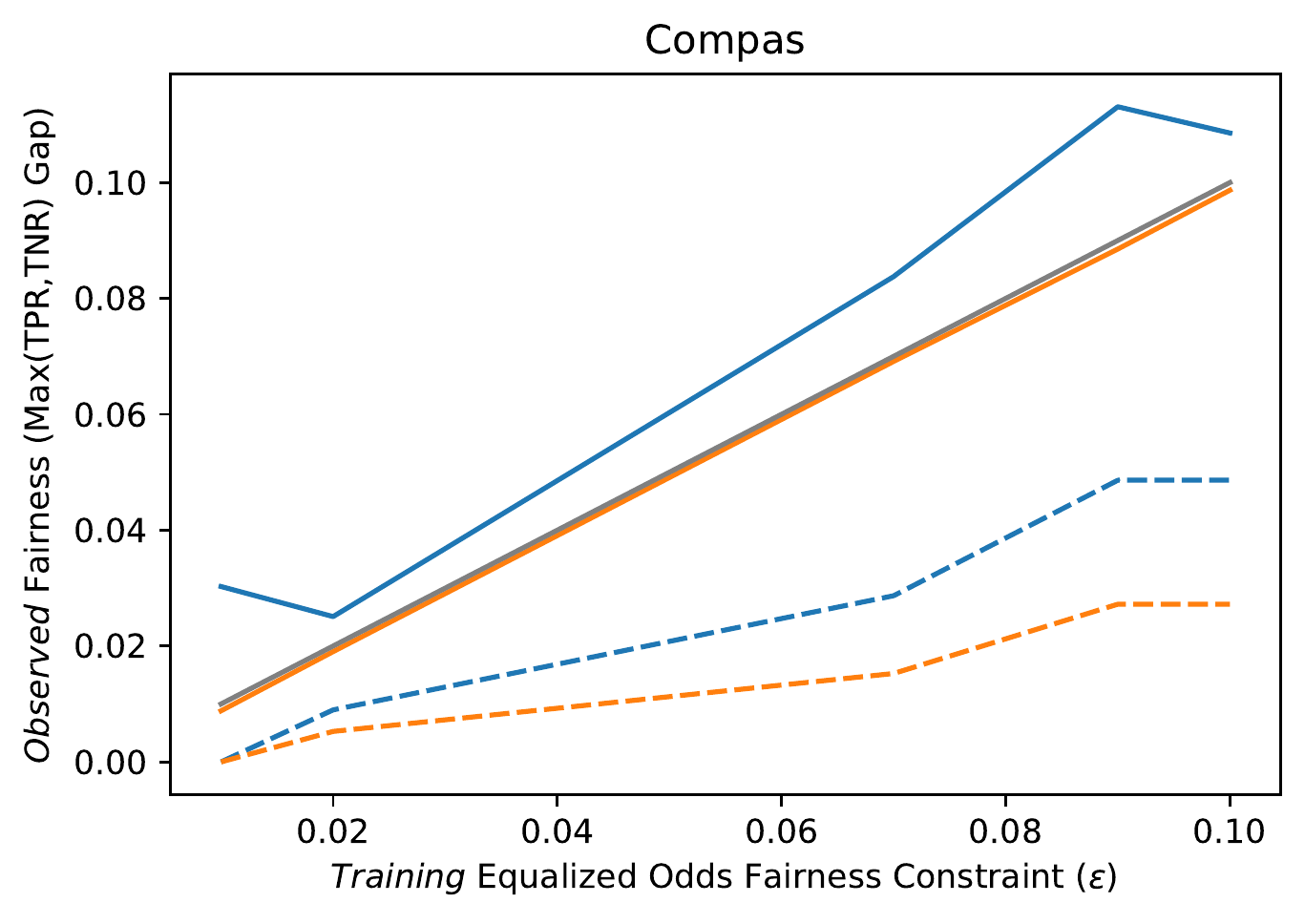}
        \end{subfigure}
        \begin{subfigure}
          \centering
          \includegraphics[width=0.32\textwidth]{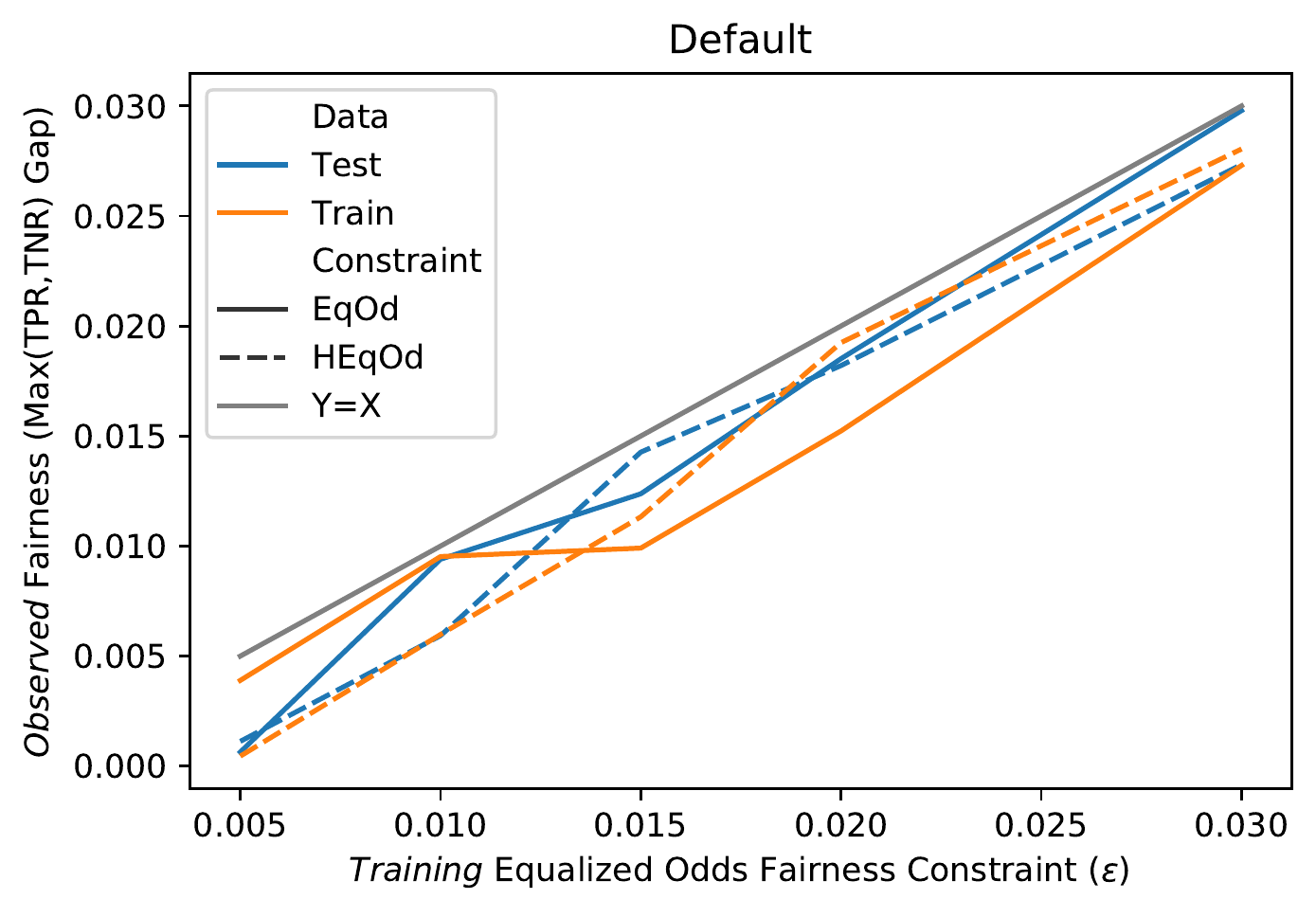}
        \end{subfigure}
         \caption {\label{Hamming_fair} Generalization of equalized odds constraint. X-axis plots the $\epsilon$ used to constrain both the Hamming loss proxy and true equalized odds constraints. Y-axis is the realized unfairness (i.e., maximum of gaps in false positive and false negative rates between groups).  
         Blue line shows fairness on training data, orange line shows fairness on testing data, and grey is the diagonal. Solid lines indicate model training using the 0-1 objective and true equalized odds constraint. Dotted lines indicate Hamming loss objective and Hamming equalized odds constraint.}
    \end{figure}

We also use Hamming loss as a proxy for 0-1 error to bound the false positive error rate in the equalized odds constraints (\ref{const:eo3}) and (\ref{const:eo4}). Figure \ref{Hamming_fair} plots the epsilon used to constrain the Hamming loss proxy for equalized odds on the x-axis, versus the true equalized odds of the resulting rule set on both the training and testing data for three UCI data sets. \cl{We compare the Hamming Loss proxy against the 0-1 formulation with exact constraints on equalized odds (for full details see Appendix \ref{app:01eqod}). In these experiments we see that rule sets trained with the Hamming loss constraint do not have true equalized odds larger than the prescribed $\epsilon$ for the training data. Compared to the 0-1 model, the Hamming loss proxy appears to be overly conservative on the compas data set but has nearly indistinguishable performance to the exact 0-1 formulation on the remaining two data sets. Moreover, the Hamming loss proxy generalizes well to unseen data. This  establishes that the Hamming loss version of the fairness constraint is an effective proxy for the true  constraint.}

\subsection{Classification} \label{sec:expClass}
For comparison with our column generation (CG) algorithm, we considered three alternative methods that also aim to control rule complexity: Bayesian Rule Sets (BRS) \citep{wang2017} and the alternating minimization (AM) and block coordinate descent (BCD) algorithms from \cite{su2016}.  Additional comparisons include the WEKA \citep{weka} JRip implementation of RIPPER \citep{cohen1995}, a rule set learner that is still state-of-the-art in accuracy, and the scikit-learn \citep{scikit-learn} implementations of the decision tree learner CART \citep{breiman1984} and Random Forests (RF) \citep{breiman2001}.  The last is an uninterpretable model intended as a benchmark for accuracy.  Appendix \ref{app:additional_classifiers} includes further comparisons to logistic regression (LR) and support vector machines (SVM).  The parameters of BRS and FPGrowth \citep{borgelt2005}, the frequent rule miner that BRS relies on, were set as recommended in \cite{wang2017} and the associated code (see Appendix \ref{app:brs_params} details).  For AM and BCD, the number of rules was fixed at $10$ with the option to disable unused rules; initialization and BCD updating are done as in \cite{su2016}.  While both \cite{su2016} and our method are equally capable of learning CNF rules, for these experiments we restricted both to learning DNF rules only. We also experimented with code made available by the authors of \cite{lakkaraju2016}.  Unfortunately, we were unable to execute this code with practical running time when the number of mined candidate rules exceeded $1000$. Appendix \ref{app:additional_classifiers} includes partial results from \citep{lakkaraju2016} that are inferior to those from the other methods. \cl{All results in this section were obtained using a single 2.0 GHz core of a server with 64 GB of memory (only a small fraction of which was used). All linear and integer programs were solved using CPLEX 12.7.1 \citep{cplex2009v12}.}

We first evaluated the accuracy-simplicity trade-offs achieved by our CG algorithm as well as BRS, AM, and BCD, methods that explicitly perform this trade-off. For CG, we used an overall time limit of 300 seconds for training and a time limit of 45 seconds for solving the Pricing Problem in each iteration. Low time limits were chosen partly due to practical considerations of running the algorithm multiple times (e.g., ~for CV) on many data sets, and partly to demonstrate the viability of IP with limited computation.  For each algorithm, the parameter controlling model complexity\footnote{Bound $C$ in \eqref{Mcomplex}, the regularization parameter $\theta$, and multiplier $\kappa$ in prior hyperparameter $\beta_l = \kappa \abs{\cA_l}$ control complexity for our approach, \cite{su2016}, and \cite{wang2017} respectively.} is varied, resulting in a set of complexity-test accuracy pairs.  A sample of these plots is shown in Figure~\ref{fig:pareto} with the full set in Appendix \ref{app:acc_simplicity_tradeoffs}.  Line segments connect points that are Pareto efficient, i.e., not dominated by solutions that are more accurate and at least as simple or vice versa.  CG \cl{outperforms} the other algorithms in $8$ out of $16$ data sets in the sense that its Pareto front is consistently higher; it nearly does so on a $9$th data set (tic-tac-toe) and on a $10$th (banknote), all algorithms are very similar.  BRS, AM, and BCD each achieve (co-)dominance only one or two times, e.g., ~in Figure~\ref{fig:pareto:musk} for AM.  Among cases where CG does not dominate are the highest-dimensional data sets (musk and gas, although for the latter CG does attain the highest accuracy given sufficient complexity) and ones where AM and/or BCD are more accurate at the lowest complexities.  BRS solutions tend to cluster in a narrow range despite varying $\kappa$ from $10^{-3}$ to $10^3$.

    \begin{figure}[t]
    \centering
        \subfigure[Heart Disease]{
          \includegraphics[width=0.45\textwidth]{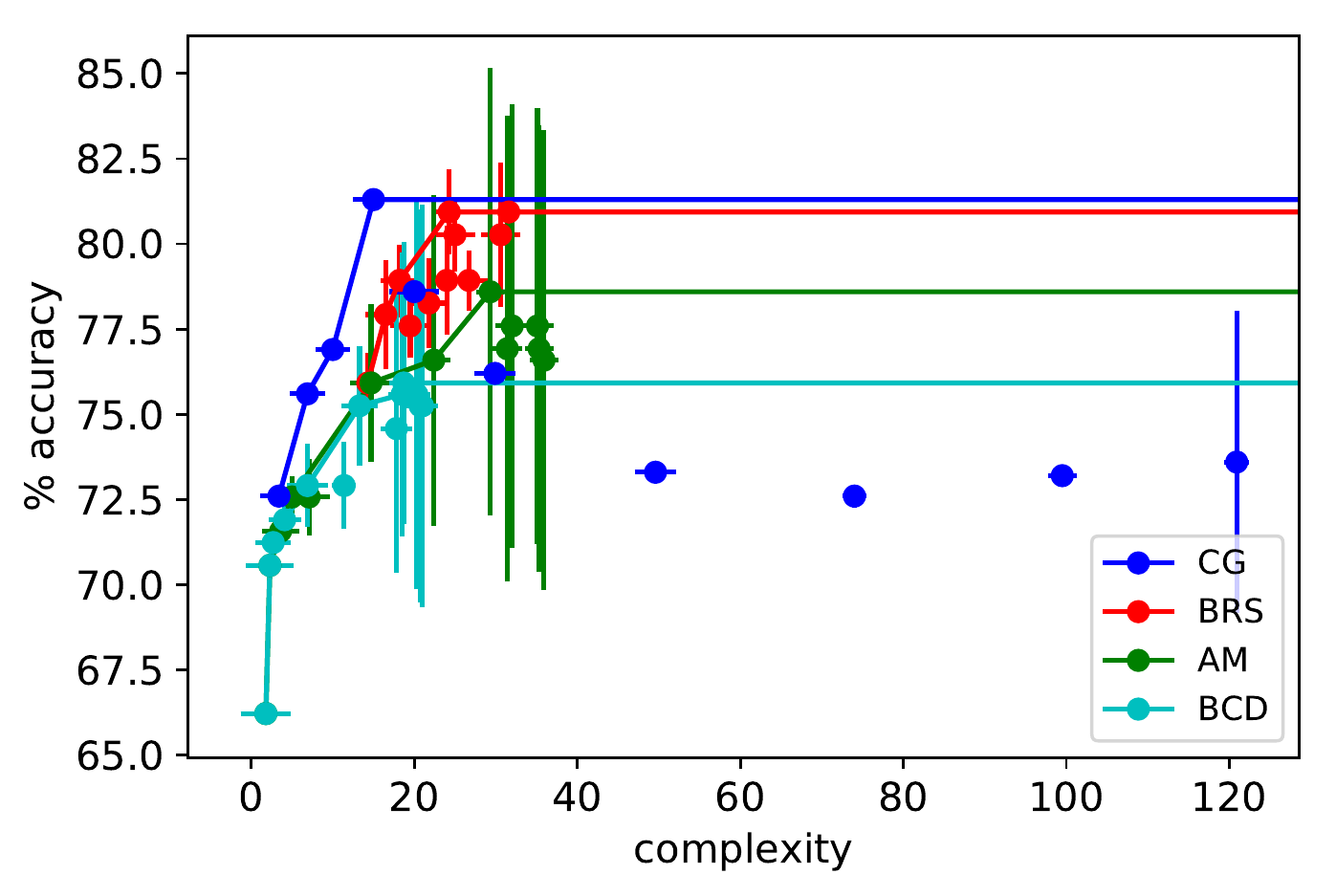}
        }
        \subfigure[FICO]{
          \includegraphics[width=0.45\textwidth]{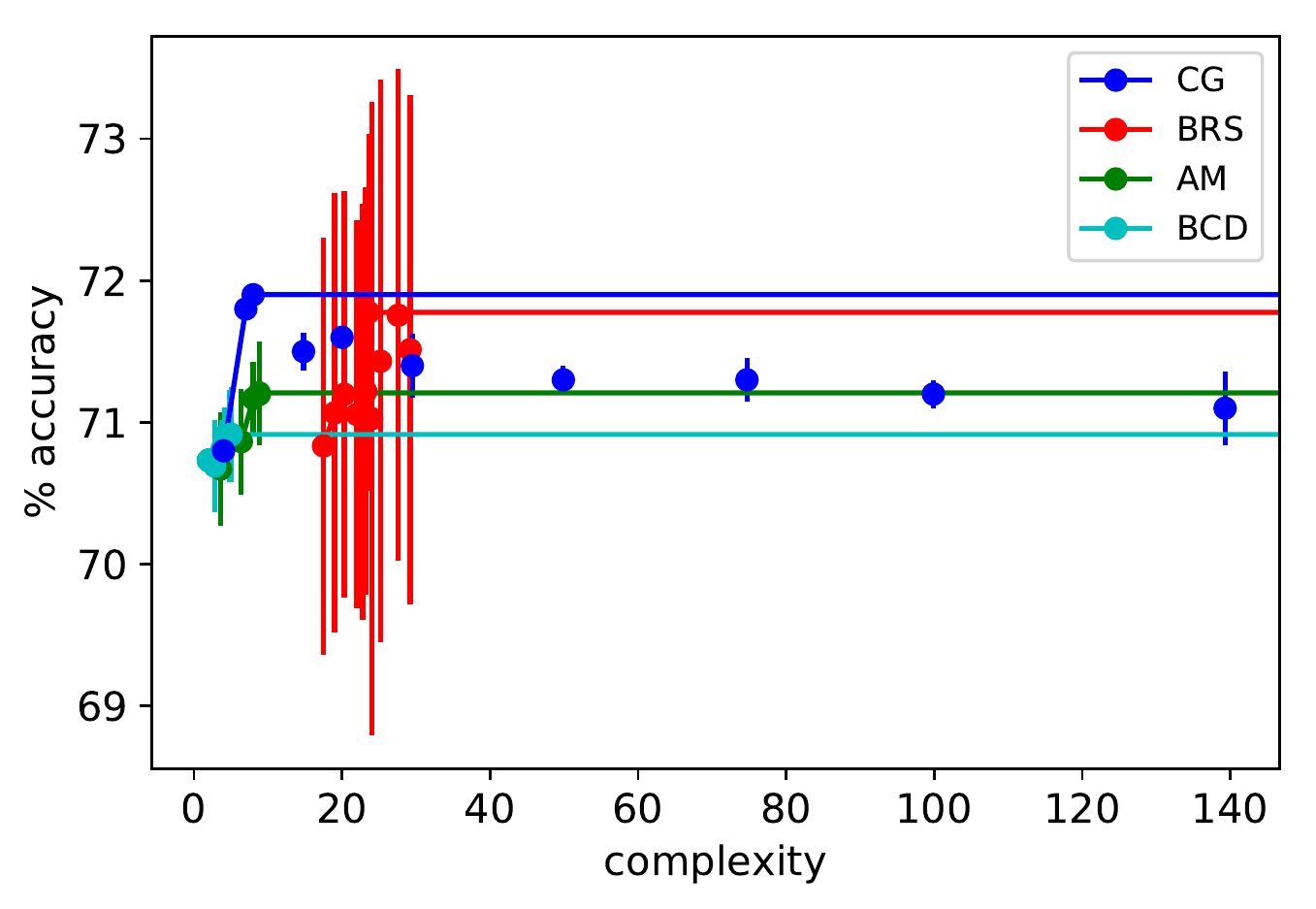}
        }
        \subfigure[MAGIC gamma telescope]{
          \centering
          \includegraphics[width=0.45\textwidth]{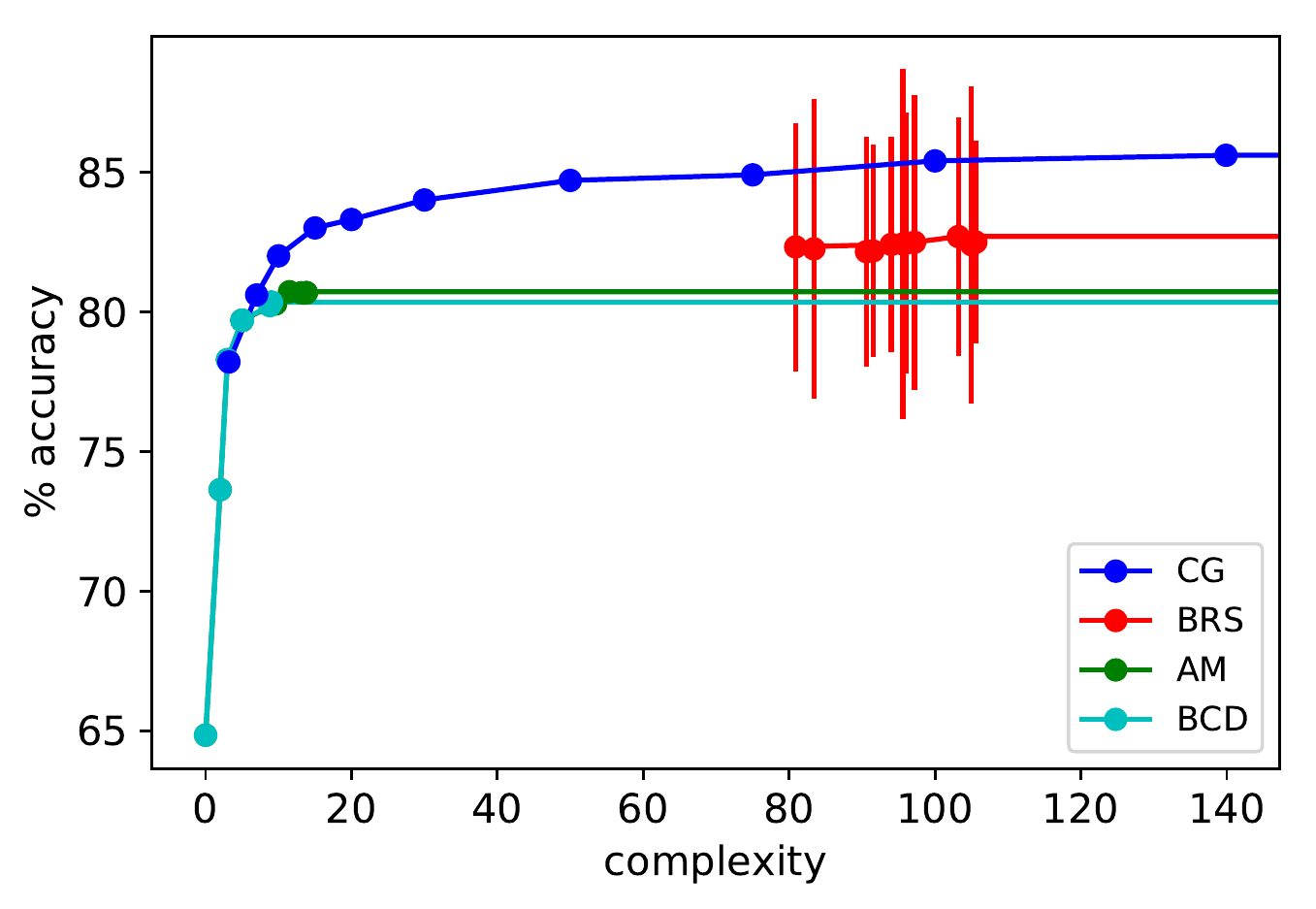}
        }
        \subfigure[Musk molecules]{
          \centering
          \includegraphics[width=0.45\textwidth]{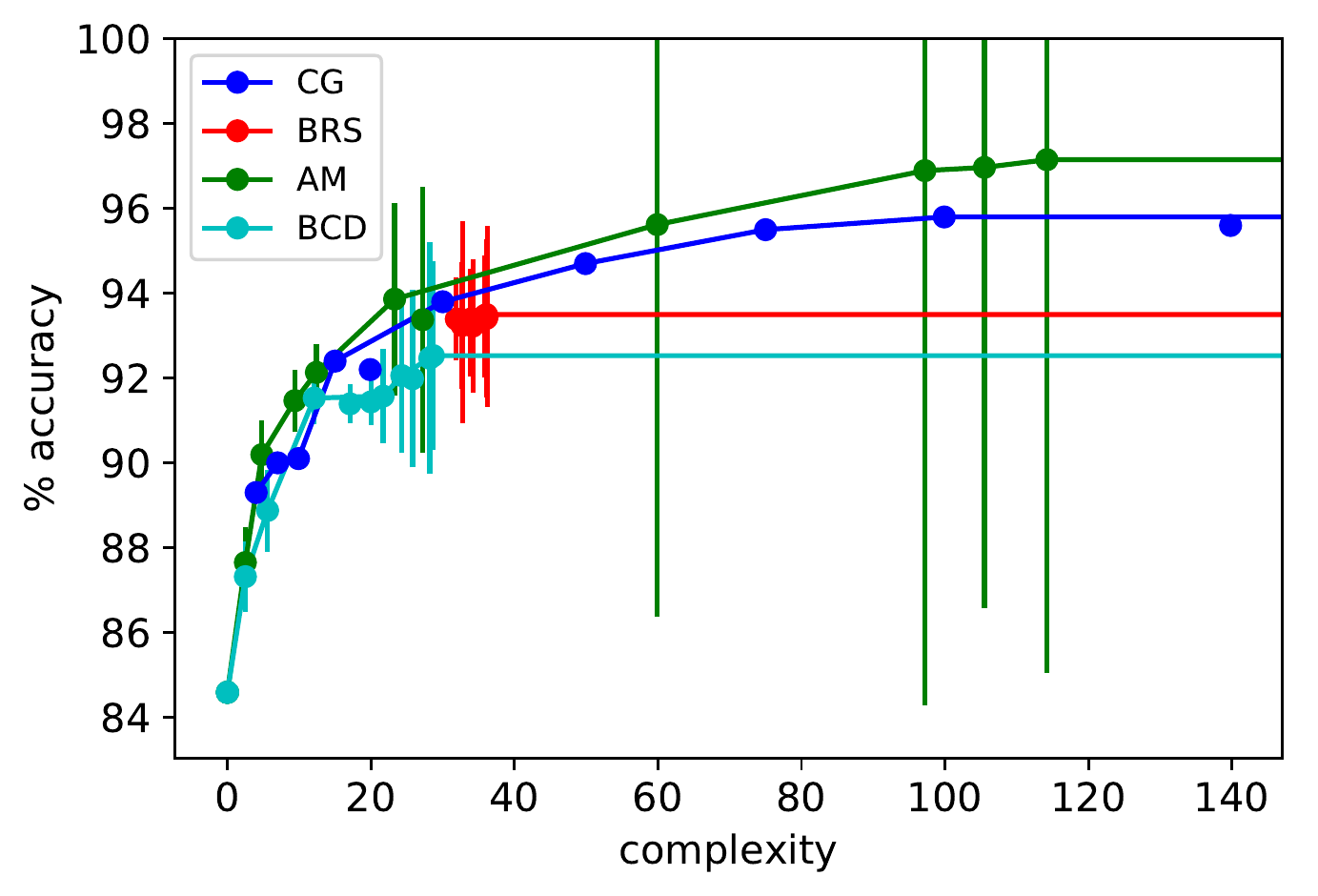}
          \label{fig:pareto:musk}
       }
         \caption {\label{fig:pareto} Rule complexity-test accuracy trade-offs on $4$ data sets. Pareto efficient points are connected by line segments. Horizontal and vertical bars represent standard errors in the means. Overall, the proposed CG algorithm dominates the others on $8$ of $16$ data sets (see Appendix \ref{app:acc_simplicity_tradeoffs} for the full set)}
    \end{figure}

In a second experiment, nested CV was used to select values of $C$ for CG and $\theta$ for AM, BCD to maximize accuracy on each training set. The selected model was then applied to the test set.  
In these experiments, CG was given an overall time limit of 120 seconds for each candidate value of $C$ and the time limit for  the Pricing Problem was set to 30 seconds.
To offset the decrease in the time limit, we performed a second pass for each data set, solving the restricted MIP with all the rules generated for all possible choices of $C$.
Mean test accuracy (over $10$ partitions) and rule set complexity are reported in Tables~\ref{tbl:acc} and \ref{tbl:complex}.  For BRS, we fixed $\kappa = 1$ as optimizing $\kappa$ did not improve accuracy on the whole (as can be expected from Figure~\ref{fig:pareto}).  Tables~\ref{tbl:acc} and \ref{tbl:complex} also include results from RIPPER, CART, and RF.  We tuned the minimum number of samples per leaf for CART and RF, used $100$ trees for RF, and otherwise kept the default settings. The complexity values for CART result from a straightforward conversion of leaves to rules (for the simpler of the two classes) and are meant only for rough comparison.

\begin{table}
\caption{Mean test accuracy (\%, standard error in parentheses). \textbf{Bold}: Best among interpretable models; \textit{Italics}: Best overall.}
\label{tbl:acc}
\centering
\footnotesize
\begin{tabular}{lrrrrrrr}
\toprule
data set&\multicolumn{1}{c}{CG}&\multicolumn{1}{c}{BRS}&\multicolumn{1}{c}{AM}&\multicolumn{1}{c}{BCD}&\multicolumn{1}{c}{RIPPER}&\multicolumn{1}{c}{CART}&\multicolumn{1}{c}{RF}\\
\midrule
banknote&$99.1$ ($0.3$)&$99.1$ ($0.2$)&$98.5$ ($0.4$)&$98.7$ ($0.2$)&$\mathbf{99.2}$ ($0.2$)&$96.8$ ($0.4$)&$\mathit{99.5}$ ($0.1$)\\
heart&$78.9$ ($2.4$)&$78.9$ ($2.4$)&$72.9$ ($1.8$)&$74.2$ ($1.9$)&$79.3$ ($2.2$)&$\mathbf{81.6}$ ($2.4$)&$\mathit{82.5}$ ($0.7$)\\
ILPD&$69.6$ ($1.2$)&$69.8$ ($0.8$)&$\bm{\mathit{71.5}}$ ($0.1$)&$\bm{\mathit{71.5}}$ ($0.1$)&$69.8$ ($1.4$)&$67.4$ ($1.6$)&$69.8$ ($0.5$)\\
ionosphere&$90.0$ ($1.8$)&$86.9$ ($1.7$)&$90.9$ ($1.7$)&$\mathbf{91.5}$ ($1.7$)&$88.0$ ($1.9$)&$87.2$ ($1.8$)&$\mathit{93.6}$ ($0.7$)\\
liver&$\mathbf{59.7}$ ($2.4$)&$53.6$ ($2.1$)&$55.7$ ($1.3$)&$51.9$ ($1.9$)&$57.1$ ($2.8$)&$55.9$ ($1.4$)&$\mathit{60.0}$ ($0.8$)\\
pima&$74.1$ ($1.9$)&$\mathbf{74.3}$ ($1.2$)&$73.2$ ($1.7$)&$73.4$ ($1.7$)&$73.4$ ($2.0$)&$72.1$ ($1.3$)&$\mathit{76.1}$ ($0.8$)\\
tic-tac-toe&$\bm{\mathit{100.0}}$ ($0.0$)&$99.9$ ($0.1$)&$84.3$ ($2.4$)&$81.5$ ($1.8$)&$98.2$ ($0.4$)&$90.1$ ($0.9$)&$98.8$ ($0.1$)\\
transfusion&$77.9$ ($1.4$)&$76.6$ ($0.2$)&$76.2$ ($0.1$)&$76.2$ ($0.1$)&$\bm{\mathit{78.9}}$ ($1.1$)&$78.7$ ($1.1$)&$77.3$ ($0.3$)\\
WDBC&$94.0$ ($1.2$)&$94.7$ ($0.6$)&$\bm{95.8}$ ($0.5$)&$\mathbf{95.8}$ ($0.5$)&$93.0$ ($0.9$)&$93.3$ ($0.9$)&$\mathit{97.2}$ ($0.2$)\\
\midrule
adult&$83.5$ ($0.3$)&$81.7$ ($0.5$)&$83.0$ ($0.2$)&$82.4$ ($0.2$)&$\mathbf{83.6}$ ($0.3$)&$83.1$ ($0.3$)&$\mathit{84.7}$ ($0.1$)\\
bank-mkt&$\bm{\mathit{90.0}}$ ($0.1$)&$87.4$ ($0.2$)&$\bm{\mathit{90.0}}$ ($0.1$)&$89.7$ ($0.1$)&$89.9$ ($0.1$)&$89.1$ ($0.2$)&$88.7$ ($0.0$)\\
gas&$98.0$ ($0.1$)&$92.2$ ($0.3$)&$97.6$ ($0.2$)&$97.0$ ($0.3$)&$\mathbf{99.0}$ ($0.1$)&$95.4$ ($0.1$)&$\mathit{99.7}$ ($0.0$)\\
FICO&$71.7$ ($0.5$)&$71.2$ ($0.3$)&$71.2$ ($0.4$)&$70.9$ ($0.4$)&$\mathbf{71.8}$ ($0.2$)&$70.9$ ($0.3$)&$\mathit{73.1}$  ($0.1$)\\ 
magic&$\mathbf{85.3}$ ($0.3$)&$82.5$ ($0.4$)&$80.7$ ($0.2$)&$80.3$ ($0.3$)&$84.5$ ($0.3$)&$82.8$ ($0.2$)&$\mathit{86.6}$ ($0.1$)\\
mushroom&$\bm{\mathit{100.0}}$ ($0.0$)&$99.7$ ($0.1$)&$99.9$ ($0.0$)&$99.9$ ($0.0$)&$\bm{\mathit{100.0}}$ ($0.0$)&$96.2$ ($0.3$)&$99.9$ ($0.0$)\\
musk&$95.6$ ($0.2$)&$93.3$ ($0.2$)&$\bm{\mathit{96.9}}$ ($0.7$)&$92.1$ ($0.2$)&$95.9$ ($0.2$)&$90.1$ ($0.3$)&$86.2$ ($0.4$)\\
\bottomrule
\end{tabular}
\end{table}

\begin{table}[t]
\caption{Mean complexity (\# rules $+$ total \# conditions, standard error in parentheses)}
\label{tbl:complex}
\centering
\footnotesize
\begin{tabular}{lrrrrrr}
\toprule
data set&\multicolumn{1}{c}{CG}&\multicolumn{1}{c}{BRS}&\multicolumn{1}{c}{AM}&\multicolumn{1}{c}{BCD}&\multicolumn{1}{c}{RIPPER}&\multicolumn{1}{c}{CART}\\
\midrule
banknote&$25.0$ ($1.9$)&$30.4$ ($1.1$)&$24.2$ ($1.5$)&$\mathbf{21.3}$ ($1.9$)&$28.6$ ($1.1$)&$51.8$ ($1.4$)\\
heart&$\mathbf{11.3}$ ($1.8$)&$24.0$ ($1.6$)&$11.5$ ($3.0$)&$15.4$ ($2.9$)&$16.0$ ($1.5$)&$32.0$ ($8.1$)\\
ILPD&$10.9$ ($2.7$)&$4.4$ ($0.4$)&$\mathbf{0.0}$ ($0.0$)&$\mathbf{0.0}$ ($0.0$)&$9.5$ ($2.5$)&$56.5$ ($10.9$)\\
ionosphere&$12.3$ ($3.0$)&$\mathbf{12.0}$ ($1.6$)&$16.0$ ($1.5$)&$14.6$ ($1.4$)&$14.6$ ($1.2$)&$46.1$ ($4.2$)\\
liver&$5.2$ ($1.2$)&$15.1$ ($1.3$)&$8.7$ ($1.8$)&$\mathbf{4.0}$ ($1.1$)&$5.4$ ($1.3$)&$60.2$ ($15.6$)\\
pima&$4.5$ ($1.3$)&$17.4$ ($0.8$)&$2.7$ ($0.6$)&$\mathbf{2.1}$ ($0.1$)&$17.0$ ($2.9$)&$34.7$ ($5.8$)\\
tic-tac-toe&$32.0$ ($0.0$)&$32.0$ ($0.0$)&$24.9$ ($3.1$)&$\mathbf{12.6}$ ($1.1$)&$32.9$ ($0.7$)&$67.2$ ($5.0$)\\
transfusion&$5.6$ ($1.2$)&$6.0$ ($0.7$)&$\mathbf{0.0}$ ($0.0$)&$\mathbf{0.0}$ ($0.0$)&$6.8$ ($0.6$)&$14.3$ ($2.3$)\\
WDBC&$13.9$ ($2.4$)&$16.0$ ($0.7$)&$\mathbf{11.6}$ ($2.2$)&$17.3$ ($2.5$)&$16.8$ ($1.5$)&$15.6$ ($2.2$)\\
\midrule
adult&$88.0$ ($11.4$)&$39.1$ ($1.3$)&$15.0$ ($0.0$)&$\mathbf{13.2}$ ($0.2$)&$133.3$ ($6.3$)&$95.9$ ($4.3$)\\
bank-mkt&$9.9$ ($0.1$)&$13.2$ ($0.6$)&$6.8$ ($0.7$)&$\mathbf{2.1}$ ($0.1$)&$56.4$ ($12.8$)&$3.0$ ($0.0$)\\
gas&$123.9$ ($6.5$)&$\mathbf{22.4}$ ($2.0$)&$62.4$ ($1.9$)&$27.8$ ($2.5$)&$145.3$ ($4.2$)&$104.7$ ($1.0$)\\
FICO&$13.3$ ($4.1$)&$23.2$ ($1.4$)&$8.7$ ($0.4$)&$\mathbf{4.8}$ ($0.3$)&$88.1$ ($7.0$)&$155.0$ ($27.5$)\\ 
magic&$93.0$ ($10.7$)&$97.2$ ($5.3$)&$11.5$ ($0.2$)&$\mathbf{9.0}$ ($0.0$)&$177.3$ ($8.9$)&$125.5$ ($3.2$)\\
mushroom&$17.8$ ($0.3$)&$17.5$ ($0.4$)&$15.4$ ($0.6$)&$14.6$ ($0.6$)&$17.0$ ($0.4$)&$\mathbf{9.3}$ ($0.2$)\\
musk&$123.9$ ($6.5$)&$33.9$ ($1.3$)&$101.3$ ($11.6$)&$24.4$ ($1.9$)&$143.4$ ($5.5$)&$\mathbf{17.0}$ ($0.7$)\\
\bottomrule
\end{tabular}
\end{table}
The favorable performance of CG compared to BRS, AM, and BCD is carried over into Table~\ref{tbl:acc}, especially for larger data sets (bottom partition in the table).  Compared to RIPPER, which is designed to maximize accuracy, CG is very competitive.  The head-to-head ``win-loss'' record is nearly even and on no data set is CG less accurate by more than $1\%$, whereas RIPPER is worse by $\sim 2\%$ on ionosphere, liver, and tic-tac-toe.  Moreover on larger data sets, CG tends to learn significantly simpler rule sets that are nearly as or even more accurate than RIPPER, e.g., on bank-marketing, magic, and FICO.  CART on the other hand is less competitive in this experiment.  Tic-tac-toe is notable in admitting an exact rule set solution, corresponding to all positions with three x's or or's in a row. CG succeeds in finding this rule set whereas the other algorithms including RF cannot quite do so.

Given the performance of our approach, a relevant question is whether certifiably optimal or near optimal solutions to the Master IP are obtained in practice. As previously discussed in Section \ref{sec:opt_guar}, for medium and large data sets where the pricing problem cannot be solved to optimality or sub-sampling is employed we cannot compute a strong lower bound. However, for small data sets our approach is able to obtain optimal or near optimal solutions to the training problem. For example, for transfusion, we can certify that the optimality gap is at most 0.7\% when the bound on the complexity of the rule set $C$ is set to 15.

To give a sense for the interpretability of DNF rules, we present some sample rule sets learned by CG. The following rule set, trained in the standard classification setting, maximizes accuracy on the FICO data with two simple rules:

{
\centering
\[\big(\mathrm{NumSatTrades} \ge 23\big) \mathrm{ AND } \big( \mathrm{ExtRiskEstimate} \geq 70\big) \mathrm{ AND } \big( \mathrm{NetFracRevolvBurden} \le63\big) \]
OR
\[\big(\mathrm{NumSatTrades} \le 22\big) \mathrm{ AND } \big( \mathrm{ExtRiskEstimate} \geq 76\big) \mathrm{ AND } \big( \mathrm{NetFracRevolvBurden} \le78\big). \]}

According to the data dictionary provided with the FICO challenge \citep{FICO2018}, ``NumSatTrades'' is the number of satisfactory accounts, ``ExtRiskEstimate'' is a consolidated version of some risk markers, and ``NetFracRevolvBurden'' is the ratio of revolving balance to credit limit.  The rules thus identify two groups, one with more accounts and less revolving debt, the other with fewer accounts and somewhat more revolving debt.  A slightly higher (better) ``ExtRiskEstimate'' is required for the second, riskier group. This rule set won the FICO interpretable machine learning challenge in 2018 \citep{FICO2018}. 

\cl{ One limitation that has been documented in other work on interpretable models \citep{guidotti2018assessing, semenova2022existence} is that the rule sets output by our method are sensitive to small changes in the training data (see Appendix \ref{app:instability} for an example), as with other rule-based models. It remains an open question on how to adapt the training of an interpretable rule set to be robust to changes in the training data.}

\subsection{Fairness}\label{sec:expFair}
    We compared the performance of our formulation under fairness constraints against four other methods for fair binary classification: those of \cite{Zafar_2017}, denoted Zafar, and \cite{hardt2016equality}, denoted Hardt, and the exponential gradient method included in the Fairlearn package \citep{agarwal2018reductions}, denoted Fair Learn. The first method builds a fair logistic regression classifier which we regard as interpretable, formulating 
    the problem as a constrained optimization model and using a convex relaxation of the fairness constraint. The method of \cite{hardt2016equality} is a post-processing approach that 
    achieves fairness by selecting different discrimination thresholds for the sensitive groups; we choose the classifier to be logistic regression again. The exponential gradient method from Fairlearn works by solving a sequence of cost-sensitive classification problems to construct a randomized classifier with low error and the desired fairness. The framework works with any classifier, however for our experiments we chose to use a decision tree as the base learner to provide the best comparison to our rule set approach. \cl{We also experimented with the publicly available code for Fair CORELS \citep{aivodji2021faircorels}, another fair rule-based method.  Unfortunately, we were unable to replicate the reported performance of Fair CORELS with the publicly available implementation. Appendix \ref{app:faircorels} includes results from our experiments with the publicly available code for Fair CORELS that compare unfavorably to the other methods.} \cl{All results in this section were run on a personal computer with a 3 GHz processor and 16 GB of RAM. All the linear and integer programs were solved using Gurobi 9.0 \citep{gurobi}. We also employed the Pool Select heuristic outlined in Appendix \ref{sec:solPool}.}

\begin{table*}[t]
\footnotesize
\centering
\caption{\label{accTableeNoConst} Mean test accuracy and fairness results with no fairness constraints (standard deviation in parenthesis). Equality of opportunity and equalized odds refer to the amount of unfairness between the two groups under each fairness metric.}
\setlength{\tabcolsep}{5pt} 
\begin{tabular}{l l  c c c c }		\toprule
& & Fair CG & Zafar & Hardt & Fair Learn \\\midrule
\multirow{3}{*}{Adult} & Accuracy & 82.5 (0.5) & 85.2 (0.5) & 83.0 (0.4) & 82.4 (0.4) \\
& Equality of Opportunity & 7.6 (0.5) & 11.9 (3.7) & 18.2 (4.8) & 11.5 (4.6) \\
& Equalized Odds & 7.6 (0.5) & 11.9 (3.7) &  18.2 (4.8) & 11.5 (4.6) \\ \midrule
\multirow{3}{*}{Compas} & Accuracy & 67.6 (1.1) & 64.6 (1.9) & 65.9 (2.7) & 65.8 (2.9) \\
& Equality of Opportunity & 23.8 (5.3)& 42.8 (5.4) & 23.7 (6.4) & 21.7 (7.1)  \\
& Equalized Odds &24.1 (5.1) & 47.6 (5.8)& 27.0 (5.2) & 24.9 (4.5) \\ \midrule
\multirow{3}{*}{Default} & Accuracy & 82.0 (0.7) & 81.2 (0.8) & 77.9 (1.7) & 77.9 (1.7) \\
& Equality of Opportunity & 1.3 (0.6) & 2.7 (1.9) & 0 (0) & 0 (0)  \\
& Equalized Odds & 1.9 (0.5) & 4.2 (2.5) & 0 (0) & 0 (0) \\ 
\bottomrule
\end{tabular}%
\end{table*}%
\begin{table}[tb]
\centering
\caption{\label{complexity} Average (standard deviation) complexity of rule sets under Equality of Opportunity fairness constraints over 10 folds. }
\setlength{\tabcolsep}{5pt} 
\begin{tabular}{c c c c}		\toprule
$\epsilon$ & Adult & Compas & Default \\ \midrule 
0.01 & 59.9 (4.8) & 11.2 (1.8) & 11.4 (2.1) \\ \midrule 
0.1 & 60.1 (5.1)  & 10.9 (0.5) & 9.3 (0.9) \\ \midrule 
0.5 & 50.5 (3.1) & 9.2 (0.8) & 9.3 (1.0) \\ 
\bottomrule
\end{tabular}%
\end{table}%

 To establish a baseline, we first compared the predictive accuracy of these algorithms without fairness considerations. Table \ref{accTableeNoConst} shows the 10-fold mean and standard deviation accuracy for each algorithm without fairness criteria (i.e., $\epsilon = 1$). While the algorithms are not trained to consider fairness, we report the average test set `unfairness' of each algorithm as a baseline for the amount of discrimination that happened in the absence of controls on fairness. On two of the three data sets (compas and default), our rule sets (FairCG) have the strongest predictive performance. However, on the adult data set the rule sets were outperformed by the two logistic regression based classifiers (Zafar and Hardt). We note that for the Default data set, Hardt and FairLearn are unable to find non-trivial classifiers (i.e., do more than predict the majority class) resulting in 0 unfairness. Note that despite having the same base classifier, logistic regression, Zafar and Hardt use different optimization procedures (Zafar uses convex-concave programming instead of standard logistic regression), resulting in different base performance. 
    
    \begin{figure}[bt]
    \centering
        \begin{subfigure}
          \centering
          \includegraphics[width=0.4\linewidth]{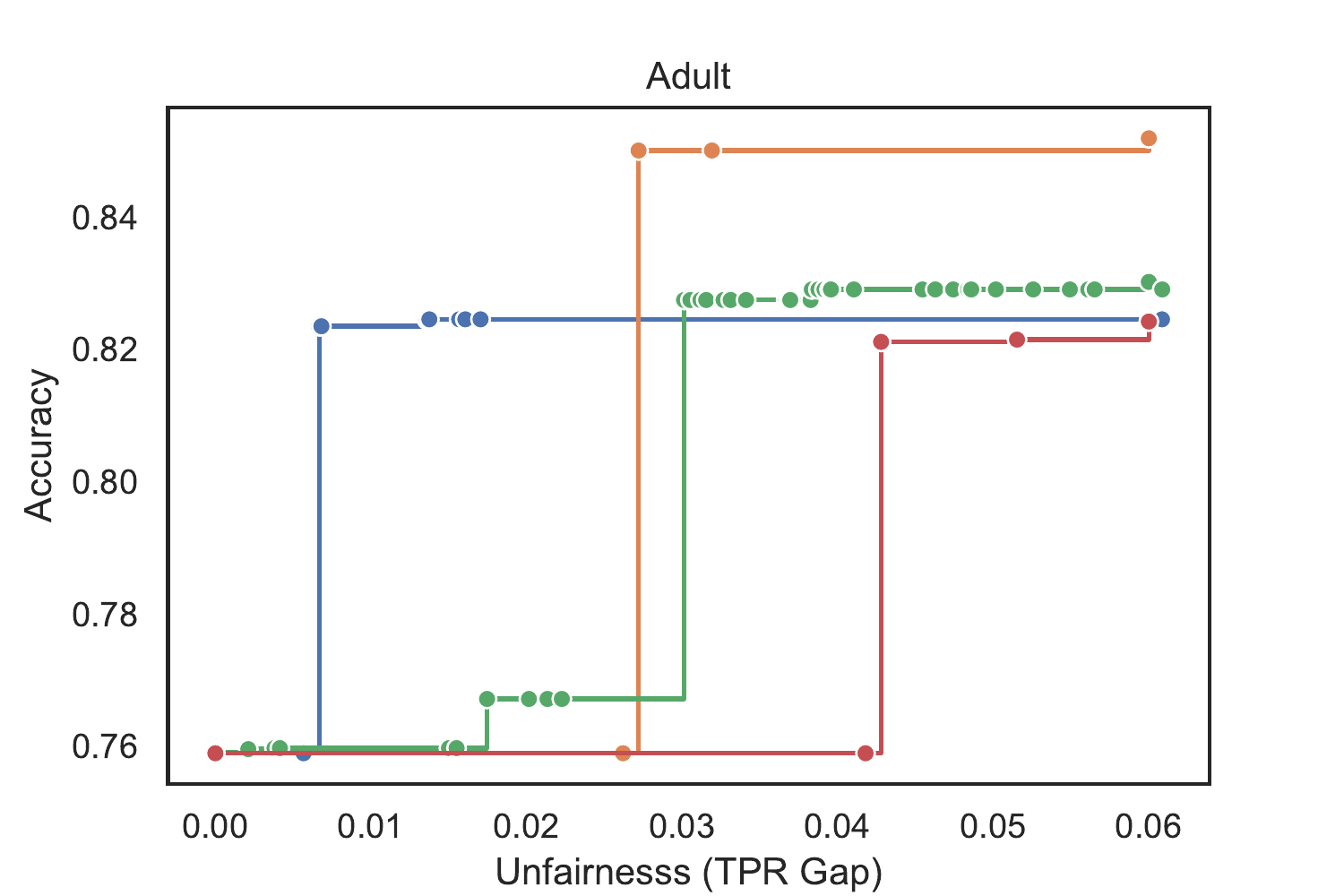}
        \end{subfigure}
        \begin{subfigure}
          \centering
          \includegraphics[width=0.4\linewidth]{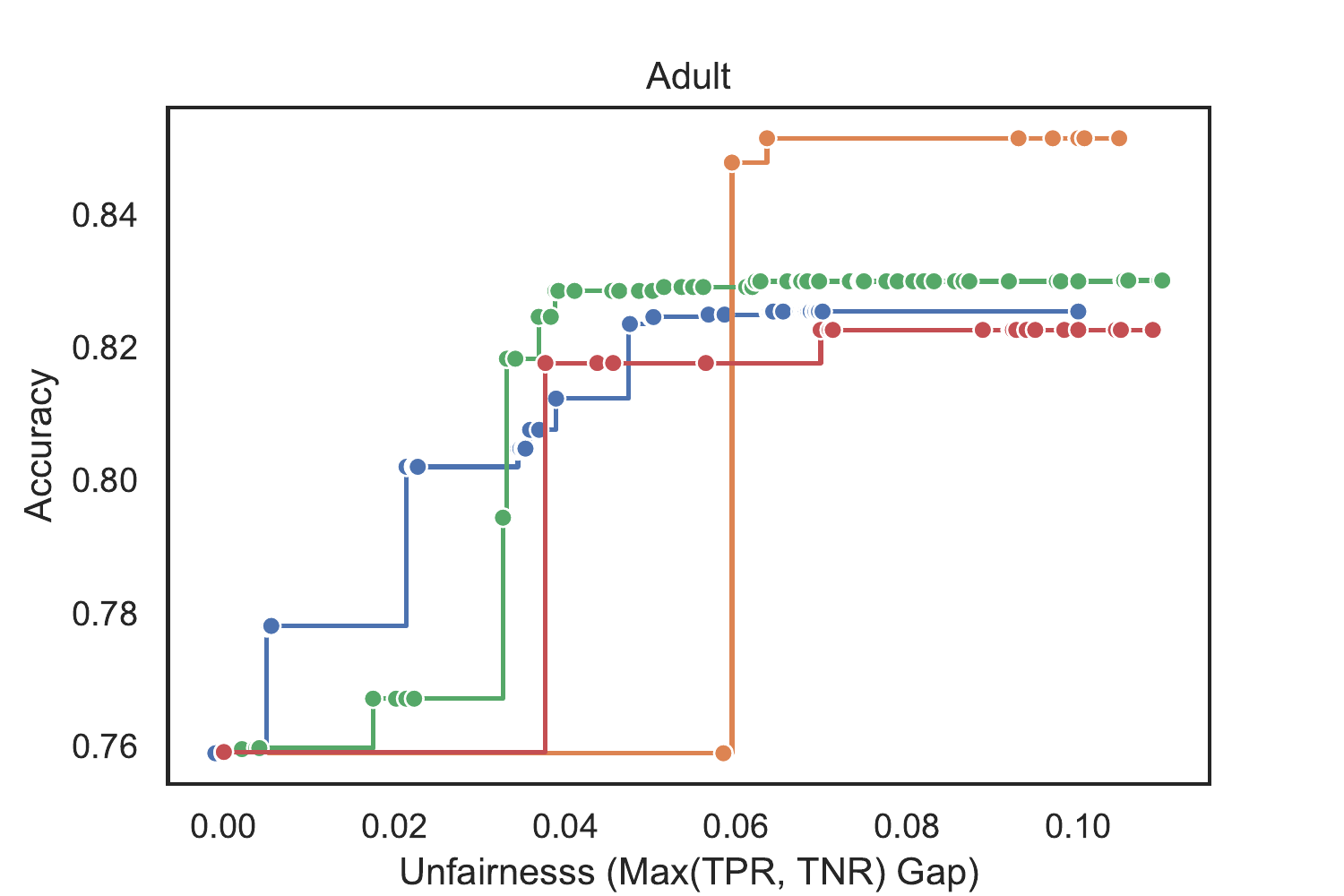}
        \end{subfigure}
        \begin{subfigure}
          \centering
          \includegraphics[width=0.4\linewidth]{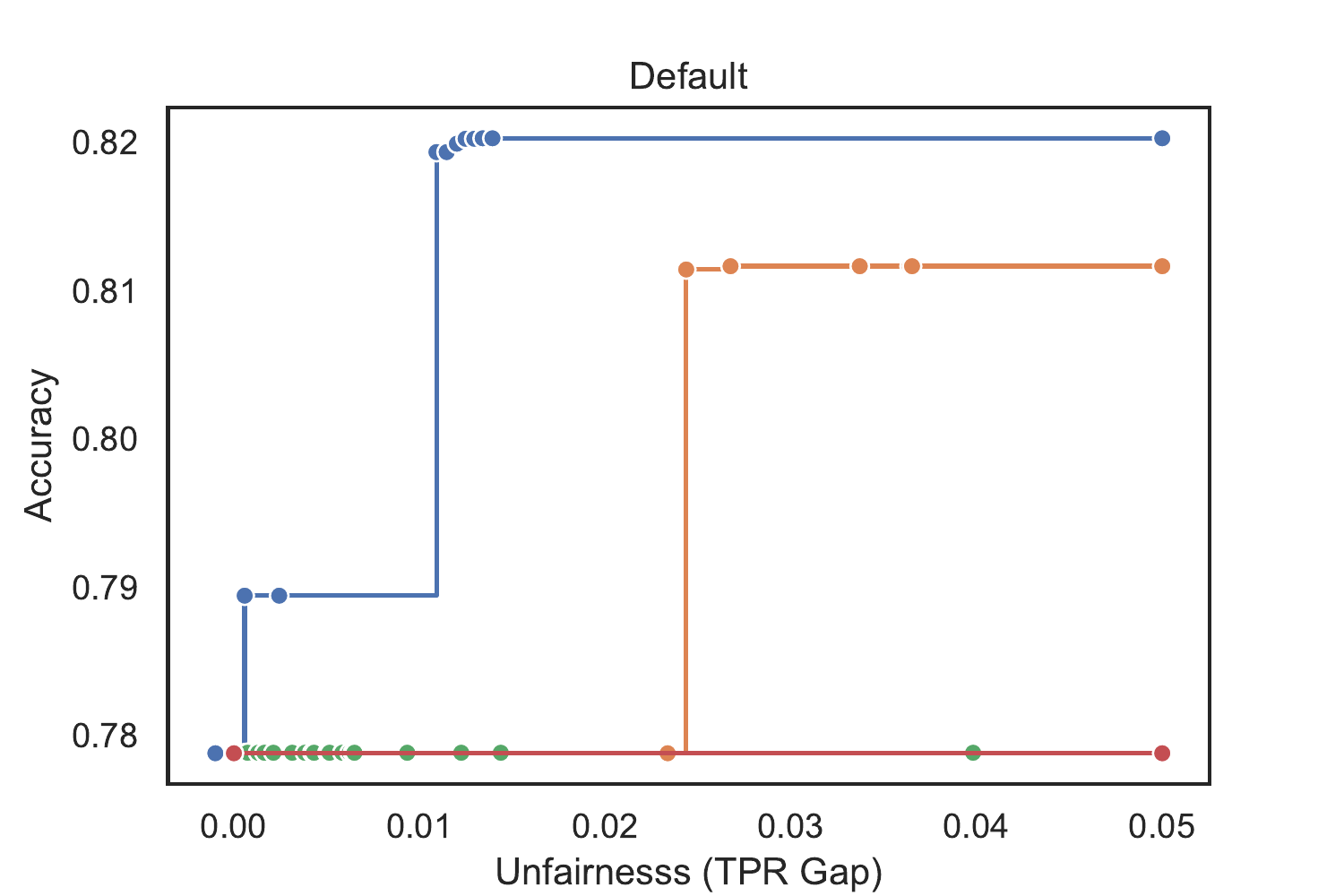}
        \end{subfigure}
        \begin{subfigure}
          \centering
          \includegraphics[width=0.4\linewidth]{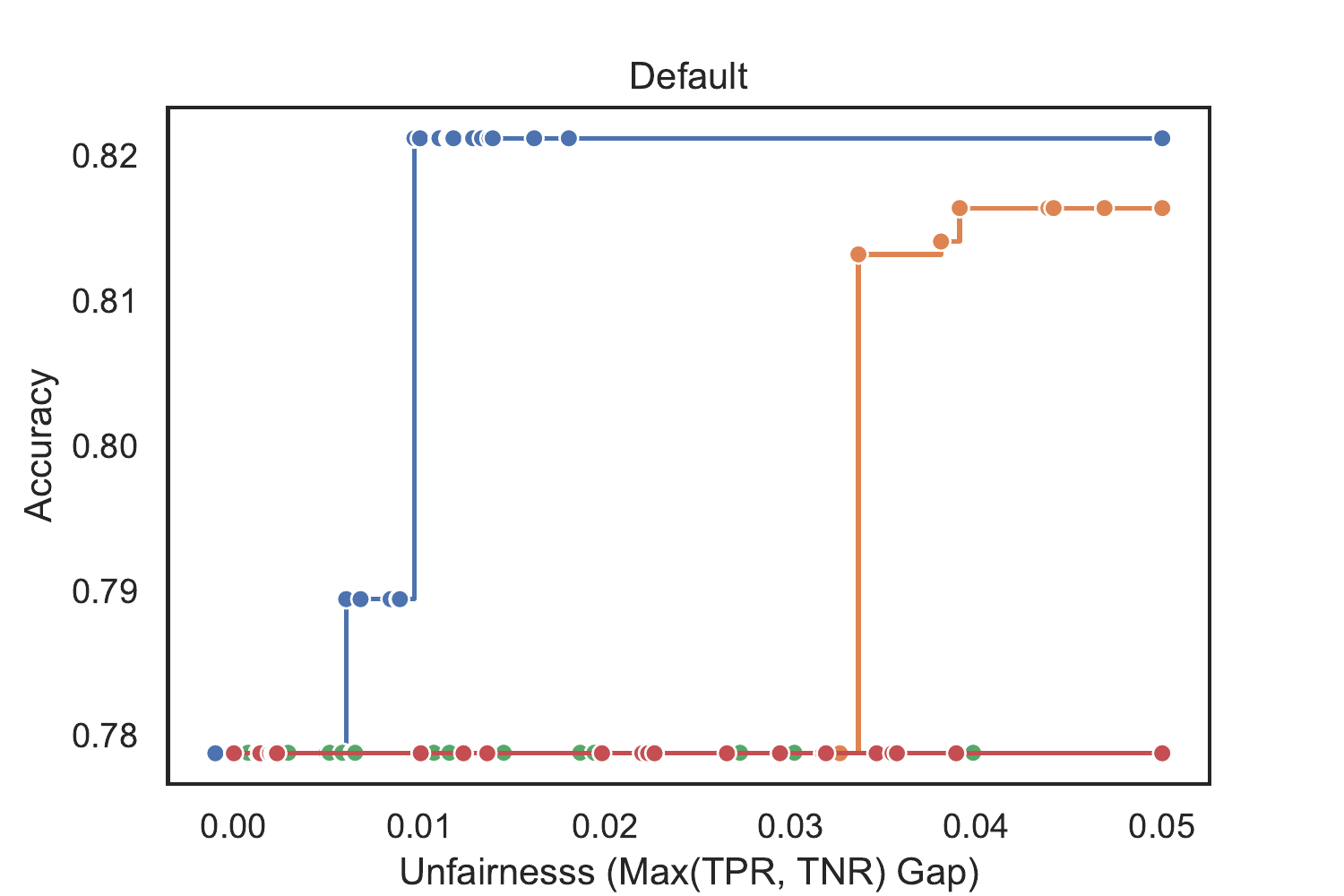}
        \end{subfigure}
        \begin{subfigure}
          \centering
          \includegraphics[width=0.4\linewidth]{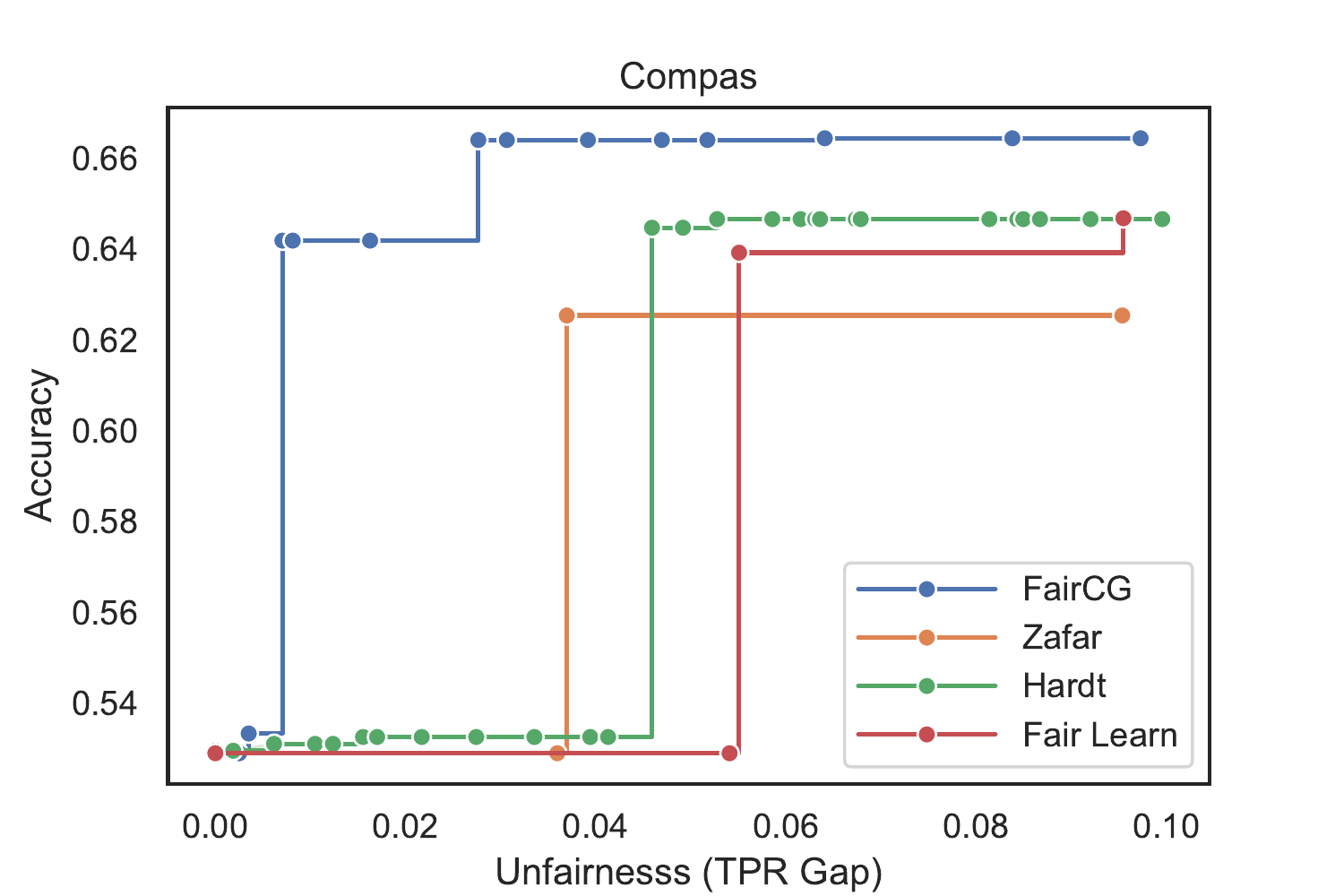}
        \end{subfigure}
    \begin{subfigure}
          \centering
          \includegraphics[width=0.4\linewidth]{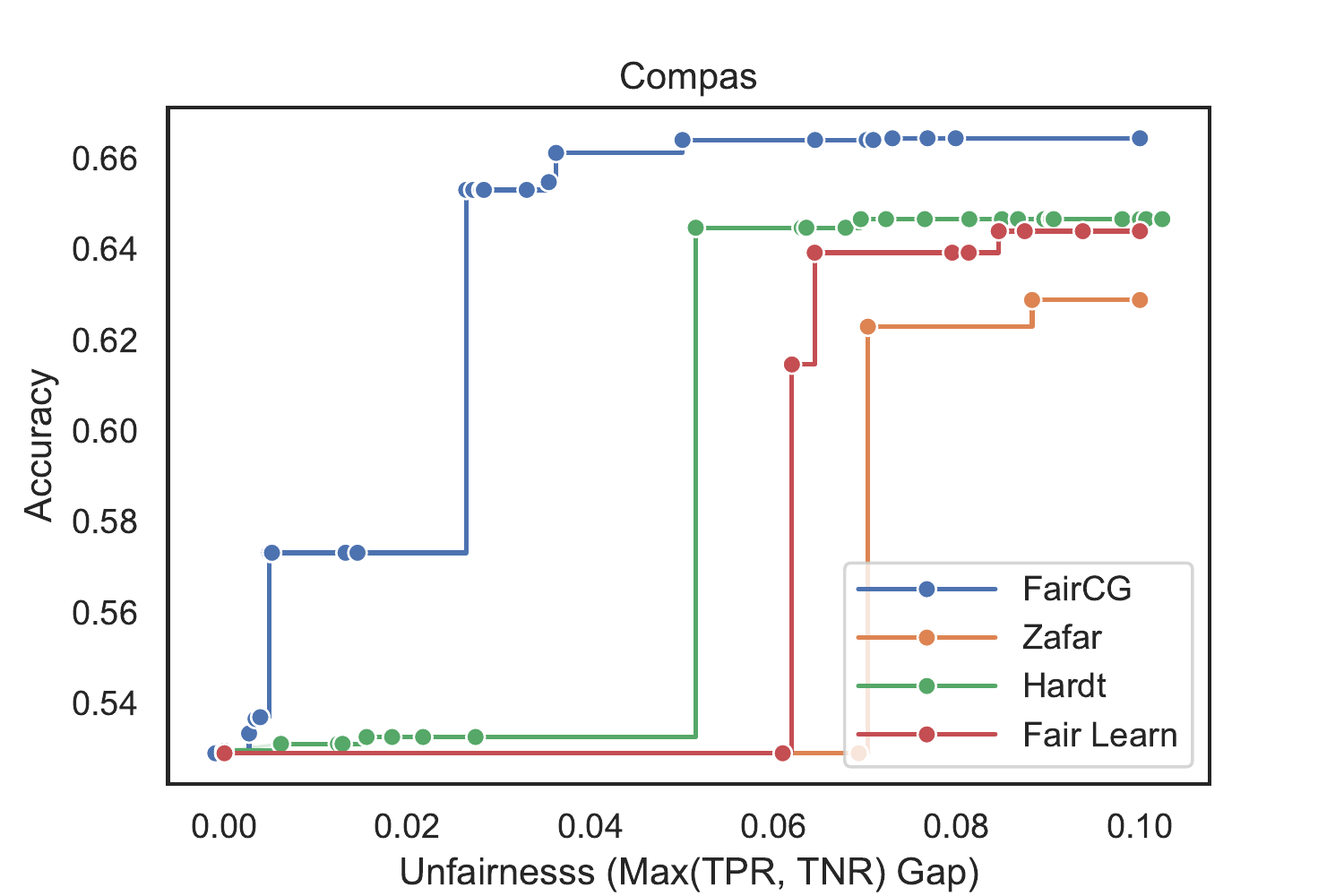}
        \end{subfigure}
         \caption {\label{benchmark_combined} Test Accuracy Fairness Frontier for Fair CG and other interpretable fair classifiers with respect to \emph{equality of opportunity} (left column) and \emph{equalized odds} (right column).}
    \end{figure}
\begin{figure}[t]
    \centering\footnotesize
        \begin{subfigure}
          \centering
          \includegraphics[width=0.6\textwidth]{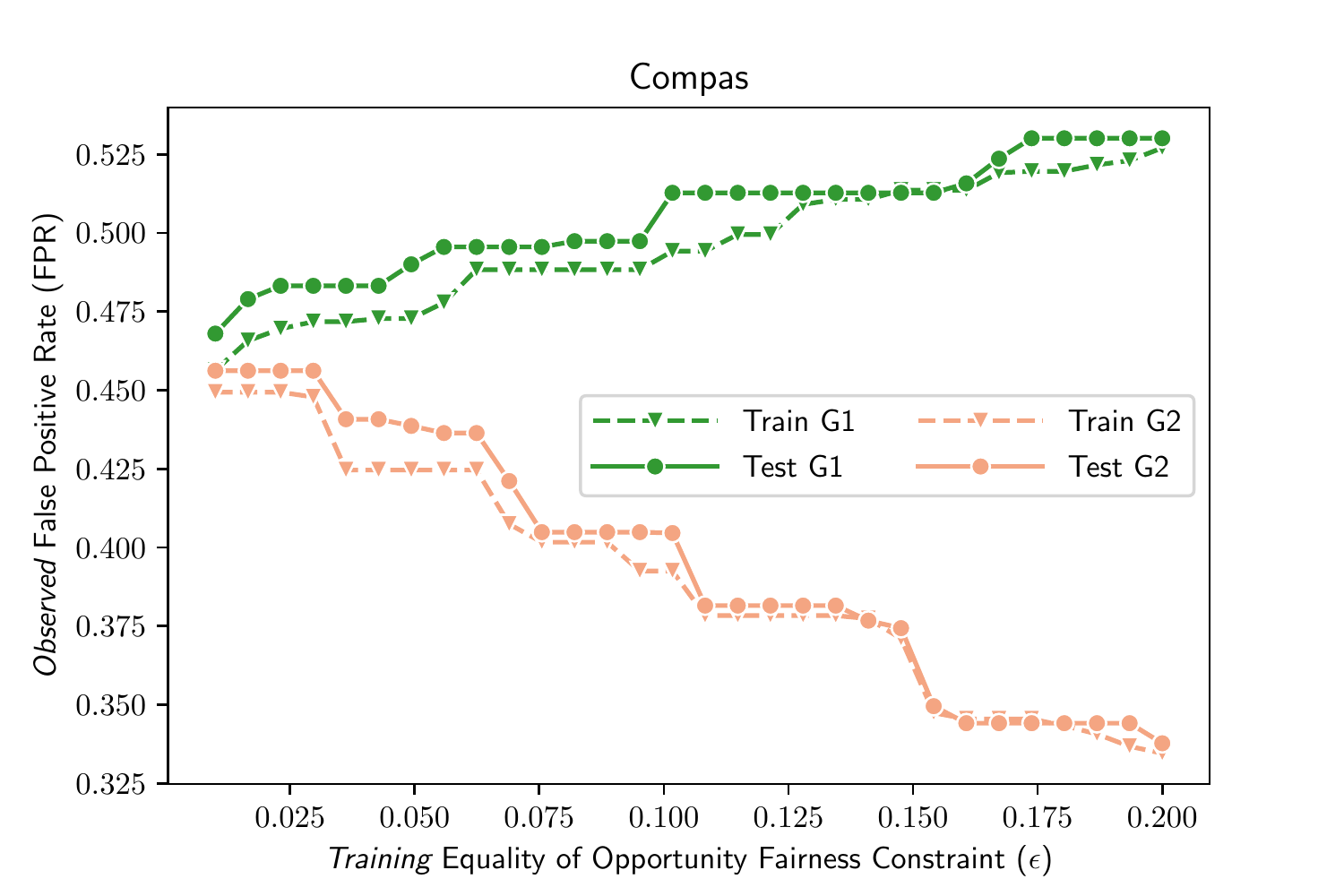}
        \end{subfigure}
         \caption {\label{diverging_tpr} Effect of relaxing the equality of opportunity fairness constraint on both train and test set true positive rate for each group (G1, G2).}\vskip-5mm
    \end{figure}

   We next consider the effect of varying the parameter $\epsilon$ that controls the allowable unfairness of the rule set. 
   We varied the hyperparameters in all the algorithms, performing 10-fold cross validation for each hyperparameter, to generate accuracy fairness trade-offs.  Figure \ref{benchmark_combined} plots the accuracy fairness trade-offs under both notions of fairness. Fair CG generated classifiers that performed well, dominating all other fair classifiers on two of the three data sets. Our algorithm performs especially well under strict fairness requirements. Specifically, we dominate all other algorithms in regimes where unfairness is restricted to less than 2.5\% with either fairness criterion. Furthermore, our fair rule sets remain simple with low complexity as shown in Table~\ref{complexity}.
   Overall these results show that our framework is able to build interpretable models that have competitive accuracy and substantially improved fairness.   Moreover, our algorithm allows for especially fine control over unfairness. Figure \ref{diverging_tpr} shows the effect of relaxing the fairness constraint on the false positive rate of both groups for the compas data set.
   We emphasise that the allowed unfairness level during training practically translates to the same observed unfairness level in testing, thus establishing the robustness of our approach. For the remainder of our results, and more specifics on our experimental framework we refer you to Appendix \ref{app:params_fairness}.

The following are sample rule sets to predict criminal recidivism on the compas data set with and without fairness constraints. We trained each rule set on one train/test split of the data set, and report the rule set as well as its out of sample accuracy and fairness. A sample rule set without any fairness constraints is:

\medskip\noindent\text{Predict repeat offence if:}
\begin{center}
\big[\text{(Score Factor$=$True) AND (Misdemeanor=False)  }\big] \\ \text{~OR~} 
\\\big[\text{(Race$\neq$Black) AND (Score Factor$=$True) AND (Misdemeanor$=$True)} \\ \text{ AND (Age$<$45) AND (Gender$=$'Male')}\big] \\
 \text{~OR~}
\\\big[\text{(Race$=$Black) AND (Score Factor$=$True) AND (Priors$\geq$ 10)} \\ \text{ AND (Age$<$45) AND (Gender$=$'Male')}\big] 
\end{center}

This rule set has a test set accuracy of $67.2\%$, but a $20\%$ gap in the false negative rate, and $22\%$ gap in the false positive rate between the two groups respectively (i.e., $22\%$ unfair with respect to equalized odds). Adding an equalized odds constraint with  $\epsilon_1=0.05$ yields:

\medskip\noindent\text{Predict repeat offence if:}
\begin{center}
\big[\text{(Race$\neq$Black) AND (Score Factor$=$True) AND (Age $<$ 45)}\big] \\ \text{~OR~}
\\\big[\text{(Race$=$Black) AND (Score Factor$=$True) AND (Misdemeanor$=$True)} \\ \text{ AND (Age$<$45) AND (Gender$=$'Male')}\big] 
\end{center}

The fair rule set is less accurate with a test set accuracy of $65.1\%$, but it now has a $1\%$ gap in the false negative rate and $3.4\%$ gap in the false positive rate between the two groups respectively. Both rule sets are arguably quite interpretable as they have a small number of rules, each with a small number of conditions. 


\cl{One omission of both rule sets is that they never predict a repeat offence if the offender is a Black woman. This is a consequence of the relatively low rate of Black women re-offenders in the data set (3 percent of the overall data). In fact, the error rate for Black women (30 percent) is below the average error rate in the test data (37 percent). However, the false negative error rate is 100 percent and thus could present fairness violations under equalized odds. This is a well documented problem in the fair machine learning literature that arises from considering gender and race separately rather than simultaneously, an approach known as intersectional fairness \citep{foulds2020intersectional}. To account for this, our formulation can easily be extended for intersectional fairness by including both race and gender as sensitive attributes (i.e., one group for Black men, Black women, white men, and white women respectively) and adding additional fairness constraints to bound the discrepancy between each pair of groups.}

\section{Conclusion}
In this paper we introduced  a column generation algorithm for learning interpretable DNF or CNF classification rules that efficiently searches the space of rules. Experiments have borne out the superiority of the accuracy-rule simplicity trade-offs achieved. In the fairness setting, experimental results on classic fair machine learning data sets validated that our algorithm is competitive with the state of the art; dominating popular fair classifiers on 2 of 3 data sets, and remaining unbeaten in regimes of strict fairness. Overall, our algorithm CG provides a powerful tool for practitioners that need simple, interpretable, and fair models for machine learning in socially sensitive settings. 

\acks{This work was generously supported by Office of Naval Research (ONR) Grant N00014-21-1-2575.}

\appendix

\section{0-1 Aggregated Formulation} \label{app:01aggregate}
\cl{In this section we present a formulation for the 0-1 loss model where the false positive constraints are aggregated, and compare it to the Hamming Loss formulation.

In the presence of a cardinality constraint on the $w_k$ (i.e., a limit on the total number of rules that can be used), the set of constraints (\ref{const:accZ01}) in model \eqref{obj:01}-\eqref{const:accBin01} for each data point can be aggregated together as follows:
\begin{align}
    \sum_{k \in {\cal K}_i} w_k  \leq   M\zeta_i  \label{const:aggAccZ01}
\end{align}
where $M$ is a suitably large constant (i.e., the maximum number of rules in a rule set). Despite the original dis-aggregated formulation being stronger (i.e., it will yield tighter bounds in a branch and bound algorithm), many modern solvers have heuristics that leverage the aggregate constraint to generate dis-aggregated constraints on the fly whenever constraints are violated by current feasible solutions, leading to much better practical performance \citep{IPref}. Given that this formulation now only has $|{\cal N}|$ constraints, as opposed to $\sum_{i \in {\cal N}} |{\cal K}_i|$ constraints in the dis-aggregated formulation, a natural question is whether this formulation is competitive with the Hamming Loss formulation introduced in Section \ref{sec:ham}. However, this aggregated formulation uses big-M constraints which are known to lead to weak linear relaxations. In practice, we noted that although the aggregated formulation led to better practical performance than the dis-aggregated model, it still under-performed the Hamming Loss formulation (as shown in Section \ref{sec:HammingEmp}).}

\section{Proof of Theorem \ref{theorem:hamm}} \label{app:hamming_theorem}

\begin{proof}
For a given even integer $d \ge 6$, let $t = d/2$.
We construct the set $\cal P$ by taking all distinct 0-1 vectors in $\{0,1\}^d$ that have zeros in the last two components, and exactly $t$ ones in the first $d-2$ components.
We let $\cal N$ consist of the following two data points in $\{0,1\}^d$: the first point has a zero in the last component and ones in the remaining components, whereas the second one has a zero in the $(d-1)$th component and ones in the remaining components.
Then $n = \binom{d-2}{t} + 2$.

The set of candidate  rules $\cal K$ consists of  conjunctions that have exactly $d/2$ out of the first $d-2$ features.
Then each rule correctly classifies exactly one point from $\cal P$ and misclassifies the two points in $\cal N$.
 When optimizing for 0-1 loss, it is clear that the optimal rule set includes all the rules in $\cal K$ which results in an expected loss of $\frac{2}{n-2}$. However, the asymmetry in Hamming loss means that including any one rule, and by extension classifying any one point in $\mathcal{P}$ correctly incurs a cost of 2. Thus the optimal solution when optimizing for Hamming loss is the empty rule set giving an expected 0-1 loss of $\frac{n-2}{n}$. Clearly no constant $\Psi$ exists such that $\Psi\frac{2}{n} \geq \frac{n-2}{n}$ for all $n$, proving the desired claim. 
\end{proof}

\cl{
\section{Proof for Theorem \ref{theorem:heo_perf}} \label{app:pf_heo_perf}

\begin{proof}
For a given even integer $d \ge 6$, let $t = d/2$. Consider a data set $\mathcal{D}$ where all points belong in $\mathcal{N}$ and there exists two groups $G_1$ and $G_2$. We construct $G_1$ by taking all distinct 0-1 vectors in $\{0,1\}^d$ that have zeros in the last component, and exactly $t$ ones in the first $d-1$ components.
We let $G_2$ consist of the following $\binom{d-2}{t}$ data points in $\{0,1\}^d$: the first point has ones in all components, whereas the remaining $\binom{d-2}{t} - 1$ points have zeros in all components. Then $n = 2\binom{d-2}{t}$, and $|G_1| = |G_2| = \binom{d-2}{t}$.

The set of candidate rules $\cal K$ consists of  conjunctions that have exactly $d/2$ out of the first $d-1$ features.
Then each rule misclassifies exactly one point from $G_1$ and the first point in $G_2$. Consider a solution that includes all the candidate rules (i.e., $K = \mathcal{K}$). We first show that such a solution satisfies Hamming Equalized odds with $\epsilon = 0$. Since $\mathcal{P} = \emptyset$ constraints (14) and (15) are met trivially, it just remains to show that constraints (17) and (18) are met. Consider $G_1$, since each data point meets exactly one rule in $\mathcal{K}$ the corresponding Hamming version of the false positive rate is:

$$\frac{1}{\abs{\mathcal{N}_1}}\sum_{i \in \mathcal{N}_1} \sum_{k \in \mathcal{K}_i} w_k = \frac{1}{\abs{G_1}}\sum_{i \in G_1} 1 = 1
$$

For $G_2$ only one point is misclassified, however it is misclassified by all rules in $\mathcal{K}$. By construction $|\mathcal{K}| = |G_1|$ and thus the Hamming version of the false positive rate for $G_2$ is:
$$\frac{1}{\abs{\mathcal{N}_2}}\sum_{i \in \mathcal{N}_2} \sum_{k \in \mathcal{K}_i} w_k = \frac{1}{\abs{G_2}} |\mathcal{K}| = \frac{\abs{G_1}}{\abs{G_2}} = 1
$$
The hamming version of the false positive rate for both groups are equal, and thus constraint (17) and (18) are met for $\epsilon = 0$. However, the actual false positive rates between the two groups can be arbitrarily large. The false positive rate of $G_1$ is $1$, but the false positive rate for $G_2$ is $\frac{1}{n}$. Clearly no constant $\Psi$ exists such that $(1 - \frac{1}{n}) \leq \Psi\epsilon$ for all $n$, proving the desired claim.
\end{proof}
}

\section{Cycling During Column Generation} \label{app:cycling}
A naive implementation of the column generation procedure as presented above can be susceptible to \emph {cycling}, the phenomenon where the Pricing Problem repeatedly produces the same column. Take a simple example with four data points with two binary features: $\textbf{X}_1 = \textbf{X}_2 = (1,0)$, with label $y_1 = y_2 = 1$, $\textbf{X}_3 = (0,1)$ with $y_3 = 1$, and $\textbf{X}_4 = (0,0)$ with $y_4 = -1$. Solving the initial master LP with the empty rule set ($\hat{{\cal K}} = \emptyset$) returns the following dual values (\eqref{MmisP}-$i$ refers to constraint \eqref{MmisP} for data point $i \in \mathcal{P} = \{1,2,3\}$): 


\begin{table}[ht]
    \centering
    \begin{tabular}{c||c|c|c|c}
         Constraint & \eqref{MmisP}-1 & \eqref{MmisP}-2 & \eqref{MmisP}-3 &  \eqref{Mcomplex}  \\ \midrule
         Dual Value & $\mu_1 =1$ & $\mu_2 =1$ & $\mu_3 =1$ & $ \lambda =0$ 
    \end{tabular}
    \caption{Dual Values for Cycling Example. }
    \label{tab:toy_ex}
\end{table}

Solving the Pricing Problem with these dual values returns a rule that checks if the first feature is 1 and has a reduced cost of $-2$. However re-solving the master LP with this rule gives identical dual values which prompts the column generation process to cycle. Meanwhile, the rule that checks if the second feature is 1 has a reduced cost of $-1$, meaning that cycling prevents the column generation procedure from finding this rule and therefore learning the optimal rule set. The reason for cycling in this example is due to the fact that modern optimization solvers implicitly add the upper bound on binary decision variables (i.e., $w_k \leq 1$), causing an additional term in the formula for the reduced cost that is not accounted for in the objective of the Pricing Problem \eqref{eq:redcost}. In this simple example, if we explicitly add the upper bound $w_k \leq 1$ to the RMLP, we see that it has an associated dual value of $-2$, giving the rule for the first feature a reduced cost of 0 in the Pricing Problem. To avoid this issue one can add explicit constraints to the Pricing Problem to prevent it from generating existing columns. This however is computationally challenging and makes the Pricing Problem harder to solve. We instead simply remove the upper bound $w_k \leq 1$ when solving the RMLP (i.e., we allow a rule to be included multiple times in a rule set). Despite removing the constraint, during our experiments we did not observe any instances where $w_k > 1$ for an optimal solution.

The risk of cycling also informed the construction of constraint \eqref{MmisP2} in the master problem. In principle, a tight formulation for the constraint would replace the co-efficient of 2 for every $w_k$ with $c_k$. However, this complicates our column generation procedure, turning the Pricing Problem into a quadratic integer program. One possible approach to circumventing this problem would be to solve the Pricing Problem with a co-efficient of 2 (i.e., the formulation presented), and then substitute the true complexity into the master problem (i.e., use $c_k$ instead of 2 when solving the RMLP). However this approach runs the risk of cycling as the objective of the Pricing Problem does not capture the true reduced cost correctly.

\section{Proof of Theorem \ref{theorem:np_pricing}} \label{app:pf_np_pricing} 

\begin{proof}
The proof is by reduction from the minimum vertex cover (MVC) problem, which is  one of Karp's 21 NP-complete problems. Remember that a vertex cover $V'$  of an undirected graph $G = ( V , E )$  is a subset of $V$  such that $\{u,v\}\cap V'\not=\emptyset$ for all $\{u,v\} \in E$. The MVC problem seeks a minimum cardinality vertex cover. Given an instance of the MVC problem, we construct an instance of the Pricing Problem by taking ${\cal P}=\emptyset$ and $D=\abs{{\cal J}}$ which makes constraints \eqref{PmisP} and \eqref{Pcomplex} redundant.
	Then, for each edge  $\{u,v\} \in E$, we create an $i\in{\cal N}$ with $S_i= \{u,v\}$.
	In addition, letting $\lambda=1/(\abs{{\cal J}}+2)$ leads to the following Pricing Problem 
\begin{alignat*}{3}
\textbf{min} &\quad&\sum_{\{u,v\}\in E}  \delta_{\{u,v\}}  &+ \lambda\Biggl(1 + \sum_{v\in V} z_v\Biggr)&\\
\textbf{s.t.}  
&& \delta_{\{u,v\}} &\geq 1- z_u-z_v, \quad \delta_i \geq 0,\quad \; &\{u,v\}\in E\\
&&z_v  &\in\{0,1\},  &{v \in V} .\end{alignat*}
	which has an optimal solution value strictly less than 1 as $\delta = \bf0$ and $z=\bf1$ is a feasible solution with value $(\abs{{\cal J}}+1)/(\abs{{\cal J}}+2)$.
	Consequently the optimal solution must have  $\delta = \bf0$ and therefore the problem becomes 
$$\textbf{min} ~~ \sum_{v\in V} z_v \quad \textbf{s.t.}~~  z_u+z_v \geq 1~~ \; \{u,v\}\in E,\quad z_v  \in\{0,1\}~~ {v \in V} $$
which is precisely the formulation of the MVC problem.
		\end{proof}

\cl{
\section{Proof of Proposition \ref{prop:lb}} \label{app:lb_proof}
\begin{proof}
At the termination of column generation either the MLP is solved to optimality (i.e., $z_{CG} \geq 0$ and $z^*_{RMLP} = z^*_{MLP}$) or it is not. In the former case, $\lceil z_{MLP} \rceil$ provides a lower bound on $z_{MIP}$ as the objective function \eqref{Mobj} has integer coefficients. In the latter case, there may still be variables excluded from the RMLP that are included in the optimal solution of the MLP. However, $z_{CG}$ provides an upper bound of the reduced cost for the missing variables. Even if the Pricing Problem is not solved to optimality, a valid lower bound on $z_{CG}$ can be used. However, it may be quite weak in practice. Since $c_k\geq2 $ for any rule there are at most $C/2$ missing variables with reduced cost  $z_{CG}$ or more that can be added to the RMLP \citep{VanderbeckWolsey96}. Thus $z_{MLP} \geq z_{RMLP} + (C/2)z_{CG}$. Combining both cases completes the proof of the proposition.
\end{proof}
}

\section{Data sets and Data Processing} \label{app:data sets}
The UCI repository data sets were used largely as-is.  We note the following deviations and label binarizations:
\begin{itemize}
\item Liver disorders: We used the number of drinks as the output variable as recommended by the data donors rather than the selector variable. The number of drinks was binarized as either $\leq 2$ or $> 2$. 
\item Gas sensor array drift:  The label was binarized as either $\leq 3$ or $> 3$ as in \citep{dash2014}. 
\item Heart disease:  We used only the Cleveland data and removed $4$ samples with `ca' = ?, yielding $299$ samples. The label was binarized as either $0$ or $> 0$ as in other works.
\end{itemize}

For the FICO data set, missing and special values were processed as follows. First, $588$ records with all entries missing (values of $-9$) were removed.  Values of $-7$ (no inquiries or delinquencies observed) were replaced by the maximum number of elapsed months in the data plus $1$.  Values of $-8$ (not applicable) and remaining values of $-9$ (missing) were combined into a single null category.  During binarization, a special indicator was created for these null values and all other comparisons with the null value return False.  Values greater than $7$ (other) in `MaxDelq2PublicRecLast12M' were imputed as $7$ (current and never delinquent) based on the corresponding values in `MaxDelqEver'.

For the compas data from ProPublica \citep{propub} we use the fair machine learning cleaned data set from Kaggle. Following the methodology of \citep{zafar2015fairness} we also restrict the data to only look at African American and Caucasian respondents - filtering all data points that belong to other races and creating a new binary column  which indicates whether or not the respondent was African American. We use this new column as our sensitive attribute for the compas data set. For adult and default we use gender as the sensitive attribute. Tables \ref{tab:data sets} and \ref{tab:fairdata sets} summarize the problem instance sizes for the data sets used in the standard classification and fair classification experiments respectively.

\begin{table}[h]
\centering
\caption{\label{tab:data sets} Overview of data sets}
\setlength{\tabcolsep}{5pt} 
\begin{tabular}{c c c c}		\toprule
data set & Samples & Binarized Features & Size \\ \midrule 
banknote & 1372 & 81 & small \\
heart & 299 & 137 & small \\
ILPD & 579 & 175 & small \\
ionosphere & 351 & 597 & small \\
bupa & 345 & 103 & small \\
pima & 768 & 145 & small \\
wdbc & 569 & 601 & small \\
transfusion & 748 & 69 & small \\
tic-tac-toe & 958 & 55 & small \\ \midrule
adult & 48842 & 249 & medium \\
bank-mkt & 41188 & 233 & medium \\
gas & 13910 & 2690 & large \\
FICO & 9871 & 347 & medium \\ 
magic & 19020 & 201 & medium \\
mushroom & 8124 & 225 & medium \\
musk & 6598 & 3414 & large \\
\bottomrule
\end{tabular}%
\end{table}%

\begin{table}[h]
\centering
\caption{\label{tab:fairdata sets} Overview of fairness data sets}
\setlength{\tabcolsep}{5pt} 
\begin{tabular}{c c c c c}		\toprule
data set & Samples & Binarized Features & Sensitive Variable & Size \\ \midrule 
Adult & 48842 & 249 & Gender & large\\
Compas & 5278 & 24 & Race & medium \\
Default & 30000 & 316 & Gender (X2 column) & large\\ 
\bottomrule
\end{tabular}%
\end{table}%

For categorical variables $j$ we use one-hot encoding to binarize each variable into multiple indicator variables that check $X_{j} = x$, and the negation $X_{j} \neq x$. For numerical variables we compare values against a sequence of thresholds for that column and include both the comparison and it's negation (i.e., $X_j \leq 1, X_j \leq 2$ and $X_j > 1, X_j > 2$). For our experiments we use the sample deciles as the thresholds for each column. We use the binarized data for all the algorithms we test to control for the binarization method.

\newpage

\section{Empirical Hamming Loss Results for All data sets} \label{app:hamming_empirical}

Table~\ref{HammingComputationTableFull} summarizes the empirical performance of Hamming loss vs. 0-1 loss for all data sets.

\begin{table*}[h]
\centering\footnotesize
\caption{\label{HammingComputationTableFull} Performance of Hamming loss vs. 0-1 loss with respect to computation time, training, and test set accuracy (standard deviation in parenthesis). A 600s time limit was placed on the solve time for all IP problems. Rows below the second divider are for the fair setting under an equality of opportunity constraint of $\epsilon=0.025$.}
\setlength{\tabcolsep}{5pt} 
\begin{tabular}{l  c c c c c c }		\toprule
& \multicolumn{2}{c}{IP Solve Time} & \multicolumn{2}{c}{Train Accuracy} & \multicolumn{2}{c}{Test Accuracy} \\
 &Hamming & 0-1 &Hamming & 0-1 & Hamming & 0-1\\\midrule
  banknote & 0.0 (0.0) & 0.0 (0.0) & 99.8 (0.0) & 99.8 (0.0) & 99.5 (0.0) & 99.5 (0.0) \\
heart & 0.1 (0.0) & 0.3 (0.2) & 88.2 (0.1) & 89.4 (0.1) & 81.8 (0.1) & 82.6 (0.1) \\
ILPD & 0.0 (0.0) & 0.5 (0.3) & 81.0 (0.1) & 81.1 (0.1) & 73.6 (0.1) & 73.6 (0.1) \\
ionosphere & 0.1 (0.1) & 0.2 (0.2) & 95.8 (0.0) & 96.2 (0.0) & 93.0 (0.0) & 92.8 (0.0) \\
pima & 0.1 (0.1) & 2.7 (4.6) & 80.4 (0.0) & 80.8 (0.0) & 77.1 (0.0) & 77.2 (0.0) \\
tic-tac-toe & 0.2 (0.1) & 0.7 (0.6) & 99.1 (0.0) & 99.1 (0.0) & 96.9 (0.0) & 96.9 (0.0) \\
transfusion & 0.0 (0.0) & 0.0 (0.0) & 80.4 (0.0) & 80.3 (0.0) & 78.6 (0.0) & 78.4 (0.0) \\
WDBC & 0.0 (0.0) & 0.1 (0.0) & 98.6 (0.0) & 98.5 (0.0) & 96.0 (0.0) & 96.1 (0.0) \\ \midrule 
adult & 1.9 (0.8) & 278.5 (67.8) & 83.1 (0.0) & 83.0 (0.0) & 82.8 (0.0) & 82.8 (0.0) \\
bank-mkt & 0.7 (0.2) & 165.6 (126.5) & 90.2 (0.0) & 90.2 (0.0) & 90.0 (0.0) & 90.0 (0.0) \\
gas & 24.4 (16.4) & 190.7 (94.4) & 97.3 (0.0) & 97.3 (0.0) & 96.9 (0.0) & 97.0 (0.0) \\
magic & 7.0 (3.7) & 270.4 (54.7) & 84.1 (0.0) & 84.2 (0.0) & 83.4 (0.0) & 83.6 (0.0) \\
mushroom & 0.1 (0.0) & 0.1 (0.0) & 100.0 (0.0) & 100.0 (0.0) & 100.0 (0.0) & 100.0 (0.0) \\
musk & 0.6 (0.4) & 1.2 (0.6) & 96.8 (0.0) & 96.8 (0.0) & 95.9 (0.0) & 95.9 (0.0) \\ \midrule
adult  &  35.5 (48.6) & 546.0 (101) & 81.9 (0.3) & 82.1 (0.3) & 81.7 (0.3)& 81.7 (0.3)\\
compas & 4.0 (3.5) & 11.4 (6.5)  & 64.8 (0.3) & 65.1 (0.2) &  64.5 (0.4) & 64.4 (0.5) \\
default &  3.5 (0.8) &  12.3 (6.1) &  78.0 (0.0) & 78.0 (0.0) & 77.7 (0.0) & 77.7 (0.0) \\
\bottomrule
\end{tabular}%
\end{table*}%

\cl{
\section{0-1 Formulation with Equalized Odds} \label{app:01eqod}

Model \eqref{obj:01eo}-\eqref{const:bin01eo} is the full formulation for the 0-1 model with equalized odds constraints used in an empirical comparison with the hamming loss proxy. \eqref{obj:01eo}-\eqref{const:accZ01eo} and \eqref{const:bin01eo} are from the original 0-1 formulation introduced in Section \ref{sec:01}. Constraint \eqref{const:complex01eo} is the complexity bound for the rule set. Constraints \eqref{const:fn101e1} and \eqref{const:fn201eo} bound the difference in false negative rate between the two groups. Finally, constraints  \eqref{const:fp101eo} and \eqref{const:fp201eo} bound the false positive rate between the two groups (these constraints are replaced with the hamming loss proxies for false positives in our model).

\begin{align}
\textbf{min} \quad \sum_{i\in \cal{P}} \zeta_i + &\sum_{i\in \cal{N}} \zeta_i ~~ \label{obj:01eo}\\
	\textbf{s.t.}  \quad
	 \zeta_i + \sum_{k\in {\cal K}_i} w_k &\geq 1 , \quad \forall i \in \cal{P}~~  \\
	 w_k  &\leq   \zeta_i , \quad  \forall i \in {\cal N}, k \in {\cal K}_i ~~ \label{const:accZ01eo} \\[.1cm]
\sum_{k \in \mathcal{K}} c_k w_k &\leq C \label{const:complex01eo}\\
\frac{1}{\abs{\mathcal{P}_1}}\sum_{i \in  \mathcal{P}_1 } \zeta_i  - \frac{1}{\abs{\mathcal{P}_2}}\sum_{i \in  \mathcal{P}_2} \zeta_i &\leq \epsilon_1 \label{const:fn101e1}\\
 \frac{1}{\abs{\mathcal{P}_2}}\sum_{i \in  \mathcal{P}_2} \zeta_i - \frac{1}{\abs{\mathcal{P}_1}}\sum_{i \in  \mathcal{P}_1} \zeta_i&\leq \epsilon_1 \label{const:fn201eo}\\ 
\frac{1}{\abs{\mathcal{N}_1}}\sum_{i \in  \mathcal{N}_1 } \zeta_i  - \frac{1}{\abs{\mathcal{N}_2}}\sum_{i \in  \mathcal{N}_2} \zeta_i &\leq \epsilon_2 \label{const:fp101eo}\\
 \frac{1}{\abs{\mathcal{N}_2}}\sum_{i \in  \mathcal{N}_2} \zeta_i - \frac{1}{\abs{\mathcal{N}_1}}\sum_{i \in  \mathcal{N}_1} \zeta_i&\leq \epsilon_2 \label{const:fp201eo}\\ 
	w\in \{0,1\}^{\abs{\mathcal{K}}},~ \zeta &\in \{0,1\}^{\abs{\mathcal{P} \cup {\cal N}}} \label{const:bin01eo}
\end{align}
}

\section{Pool Select} \label{sec:solPool}
    
While solving the restricted MIP, the mixed integer programming solver retains a set of feasible integer solutions encountered during the process. We evaluate the 0-1 loss on all these solutions, and pick the one with the lowest 0-1 loss. In other words, while we optimize for Hamming loss, we select the best solution from the candidate pool of feasible solutions using 0-1 loss. Table \ref{tab:0-1select} summarizes the impact of selecting a final rule set from the solution pool using Hamming loss and 0-1 loss respectively. For each solution selection strategy, we perform 10-fold cross validation with hyper-parameter tuning over a small subset of potential complexities and report the best average test set accuracy results. The Hamming columns correspond to using the rule set with the lowest Hamming loss, whereas the 0-1 columns correspond to selecting the solution from the solution pool that has the lowest train set 0-1 loss. By design, selecting the final rule set using 0-1 loss leads to a higher train set accuracy for every data set. While this translates to  an improvement in test set accuracy for most data sets, it is not always guaranteed. Note that while the overall increase in performance is modest, it comes at practically no additional computational cost. 

\begin{table*}[ht]
\centering\footnotesize
\caption{\label{tab:0-1select} Effect of selecting a final rule set from the solution pool using Hamming loss and 0-1 loss (standard deviation in parenthesis) }
\setlength{\tabcolsep}{5pt} 
\begin{tabular}{l c c c c }		\toprule
& \multicolumn{2}{c}{Train Accuracy} & \multicolumn{2}{c}{Test Accuracy} \\
 &Hamming & 0-1 & Hamming & 0-1\\\midrule
banknote & 97.8 (0.0) & 98.1 (0.0) & 97.3 (0.0) & 97.6 (0.0) \\
heart & 83.2 (0.0) & 83.9 (0.0) & 78.9 (0.1) & 77.9 (0.1) \\
ILPD & 72.7 (0.0) & 72.7 (0.0) & 71.5 (0.0) & 71.5 (0.0) \\
ionosphere & 93.9 (0.0) & 94.6 (0.0) & 87.8 (0.1) & 88.4 (0.1) \\
liver &  69.1 (0.01)&  69.9 (0.01) &  59.1 (0.1) &  59.7 (0.1)\\
pima & 78.1 (0.0) & 78.9 (0.0) & 73.8 (0.1) & 75.2 (0.1) \\
tic-tac-toe & 80.2 (0.0) & 80.9 (0.0) & 77.6 (0.0) & 78.3 (0.0) \\
transfusion & 80.2 (0.0) & 80.4 (0.0) & 77.9 (0.0) & 77.9 (0.0) \\
WDBC & 96.3 (0.0) & 96.3 (0.0) & 95.1 (0.0) & 94.9 (0.0) \\ \midrule
adult & 82.4 (0.0) & 82.5 (0.0) & 82.5 (0.0) & 82.6 (0.0) \\
bank-mkt & 90.0 (0.0) & 90.0 (0.0) & 90.0 (0.0) & 90.0 (0.0) \\
gas & 91.7 (0.0) & 91.9 (0.0) & 91.4 (0.0) & 91.6 (0.0) \\
FICO &  71.6 (0.0) &  72.0 (0.0) &  70.8 (0.1)&  71.1 (0.2)\\ 
magic & 80.9 (0.0) & 81.1 (0.0) & 80.7 (0.0) & 81.0 (0.0) \\
mushroom & 100.0 (0.0) & 100.0 (0.0) & 100.0 (0.0) & 100.0 (0.0) \\
musk & 92.4 (0.0) & 93.0 (0.0) & 92.1 (0.0) & 92.9 (0.0) \\
\bottomrule
\end{tabular}%
\end{table*}%

\section{BRS parameters} \label{app:brs_params}

We followed \citep{wang2017} and its associated code in setting the parameters of BRS and FPGrowth, the frequent rule miner that BRS relies on:  minimum support of $5\%$ and maximum length $3$ for FPGrowth; reduction to $5000$ candidate rules using information gain (this reduction was triggered in all cases); $\alpha_+ = \alpha_- = 500$, $\beta_+ = \beta_- = 1$, and $2$ simulated annealing chains of $500$ iterations for BRS itself.

\section{Accuracy-Simplicity Trade-offs for All data sets} \label{app:acc_simplicity_tradeoffs}

Below in Figures~\ref{fig:paretoAll2} and \ref{fig:paretoAll3} is the full set of accuracy-simplicity trade-off plots for all $16$ data sets, including the $4$ from the main text. 

\section{Results for Additional Classifiers} \label{app:additional_classifiers}

As discussed in the main text, we were unable to execute code from the authors of Interpretable Decision Sets (IDS) \citep{lakkaraju2016} with practical running time when the number of candidate rules mined by Apriori \citep{agrawal1994} exceeded $1000$.  While it is possible to limit this number by increasing the minimum support and decreasing the maximum length parameters of Apriori, we did not do so beyond a support of $5\%$ and length of $3$ (same values as with FPGrowth for BRS) as it would severely constrain the resulting candidate rules. Thus we opted to run IDS only on those data sets for which Apriori generated fewer than $900$ candidates given minimum support of $5\%$ and either maximum length of $3$ or unbounded length.  

In terms of the settings for IDS itself, we ran a deterministic version of the local search algorithm with $\epsilon = 0.05$ as recommended by the authors. We set $\lambda_6 = \lambda_7 = 1$ to have equal costs for false positive and negatives, consistent with the other algorithms. For simplicity, the overlap parameters $\lambda_3$ and $\lambda_4$ were set equal to each other and tuned separately for accuracy, yielding $\lambda_3 = \lambda_4 = 0.5$.  $\lambda_5$ was set to $0$ as it is not necessary for binary classification.  Lastly, $\lambda_1$ and $\lambda_2$ were set equal to each other to reflect the choice of complexity metric as the number of rules plus the sum of their lengths.  We then varied $\lambda_1 = \lambda_2$ over a range to trade accuracy against complexity.

    \begin{figure}[H]
    \centering
        \subfigure[banknote]{
          \includegraphics[width=0.4\textwidth]{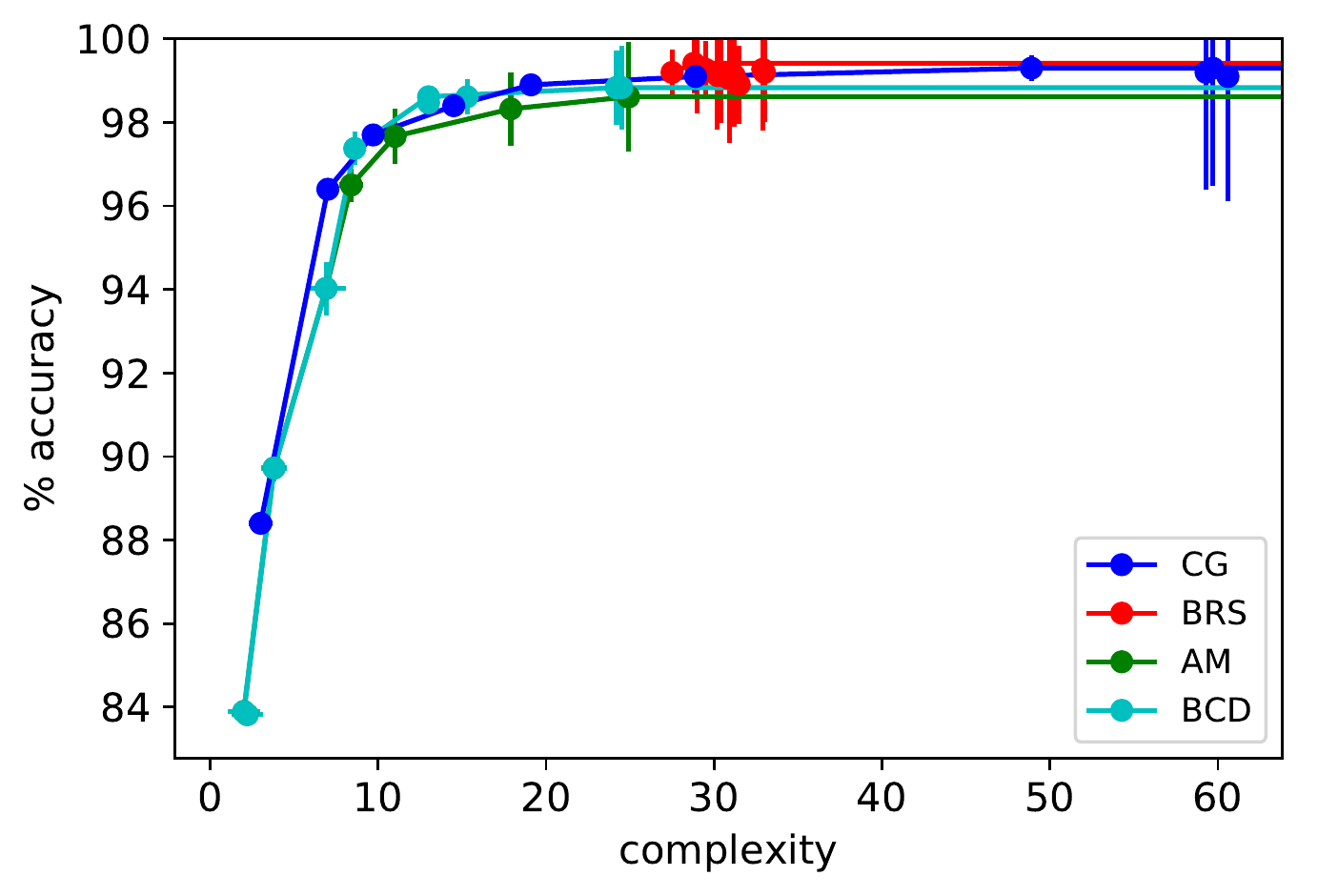}
        }
        \subfigure[heart]{
          \includegraphics[width=0.4\textwidth]{figures/pareto_heart.pdf}
        }
        \subfigure[ILPD]{
          \centering
          \includegraphics[width=0.4\textwidth]{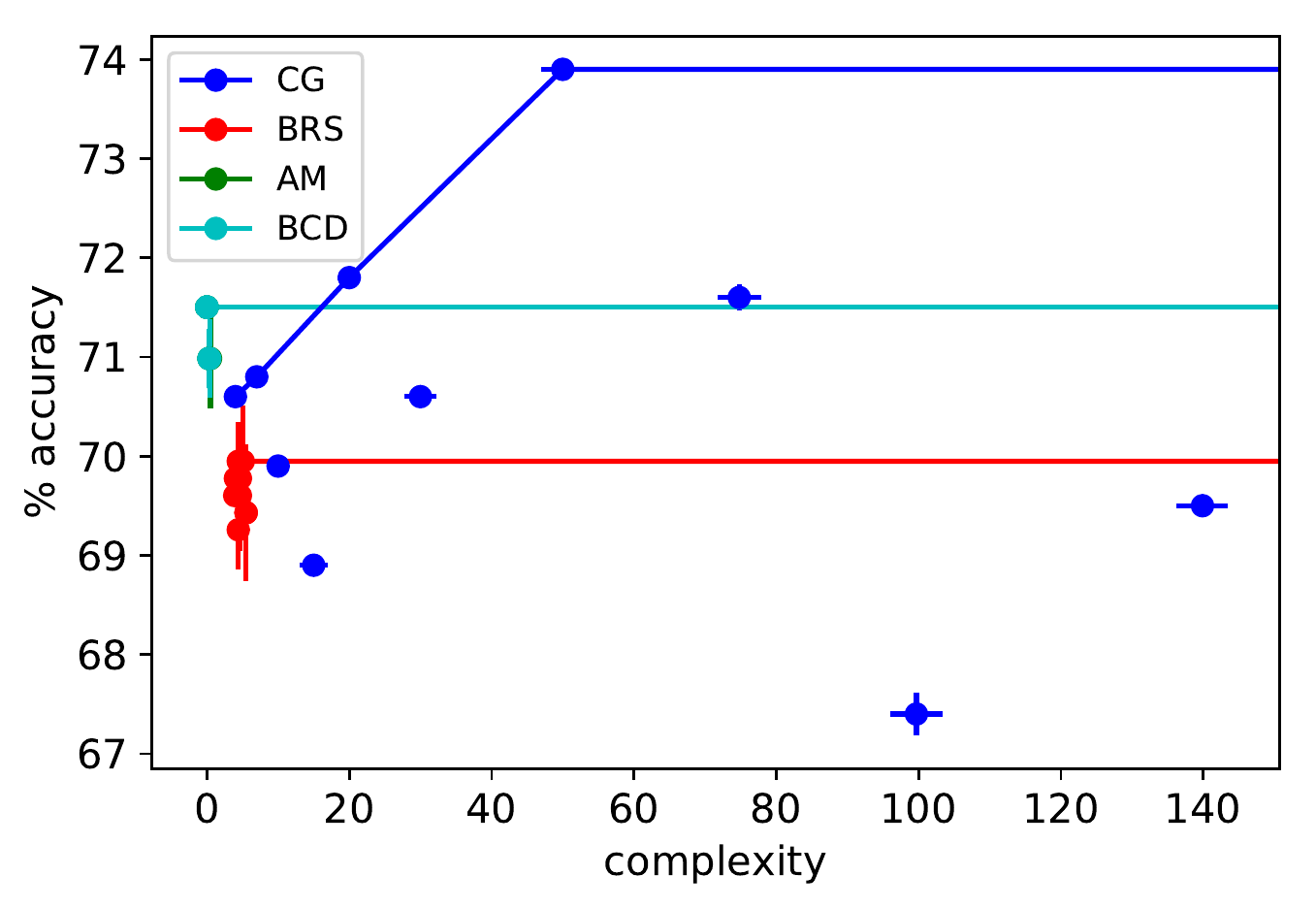}
        }
        \subfigure[ionosphere]{
          \centering
          \includegraphics[width=0.4\textwidth]{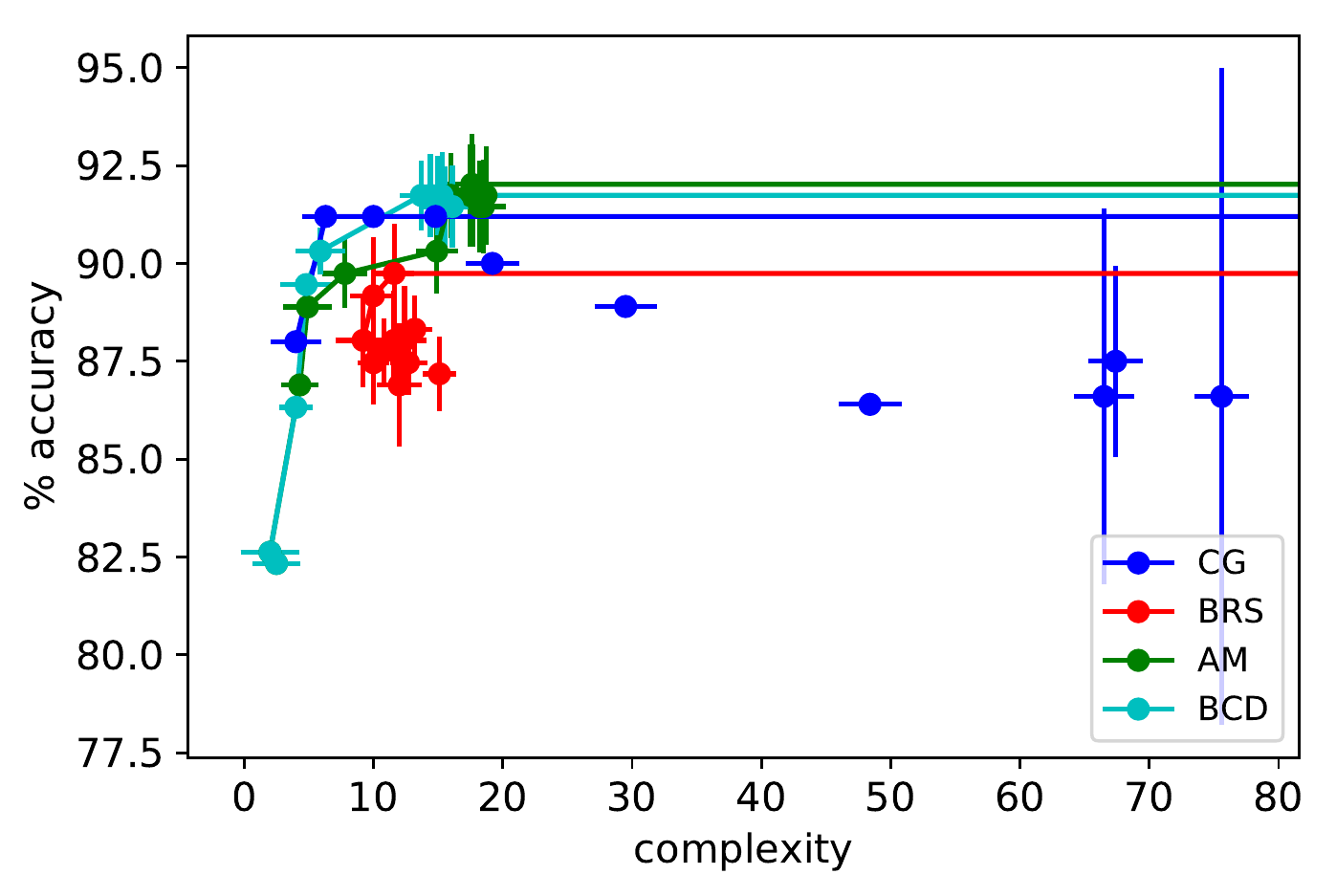}
       }
       \subfigure[liver]{
          \centering
          \includegraphics[width=0.4\textwidth]{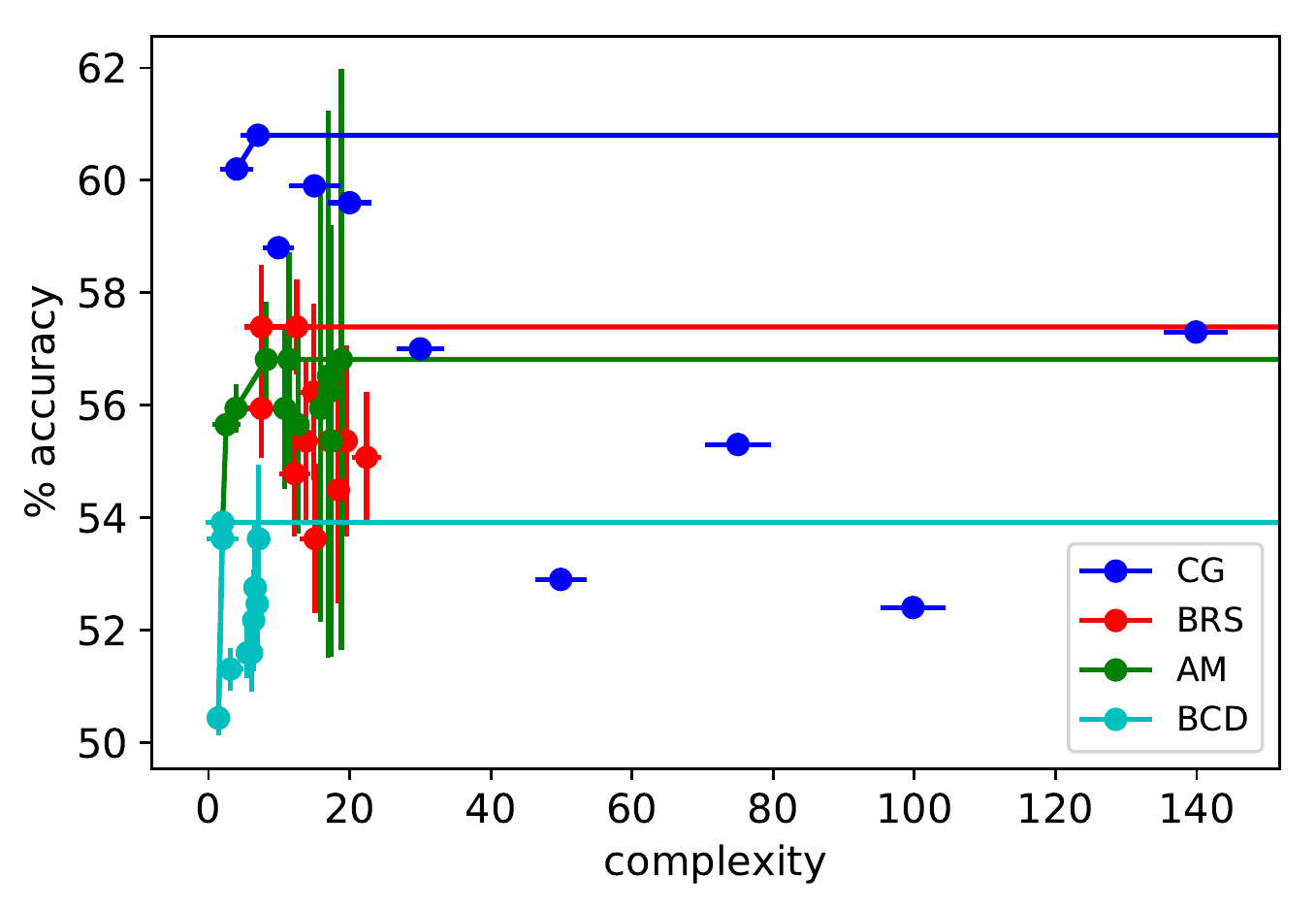}
       }
       \subfigure[pima]{
          \centering
          \includegraphics[width=0.4\textwidth]{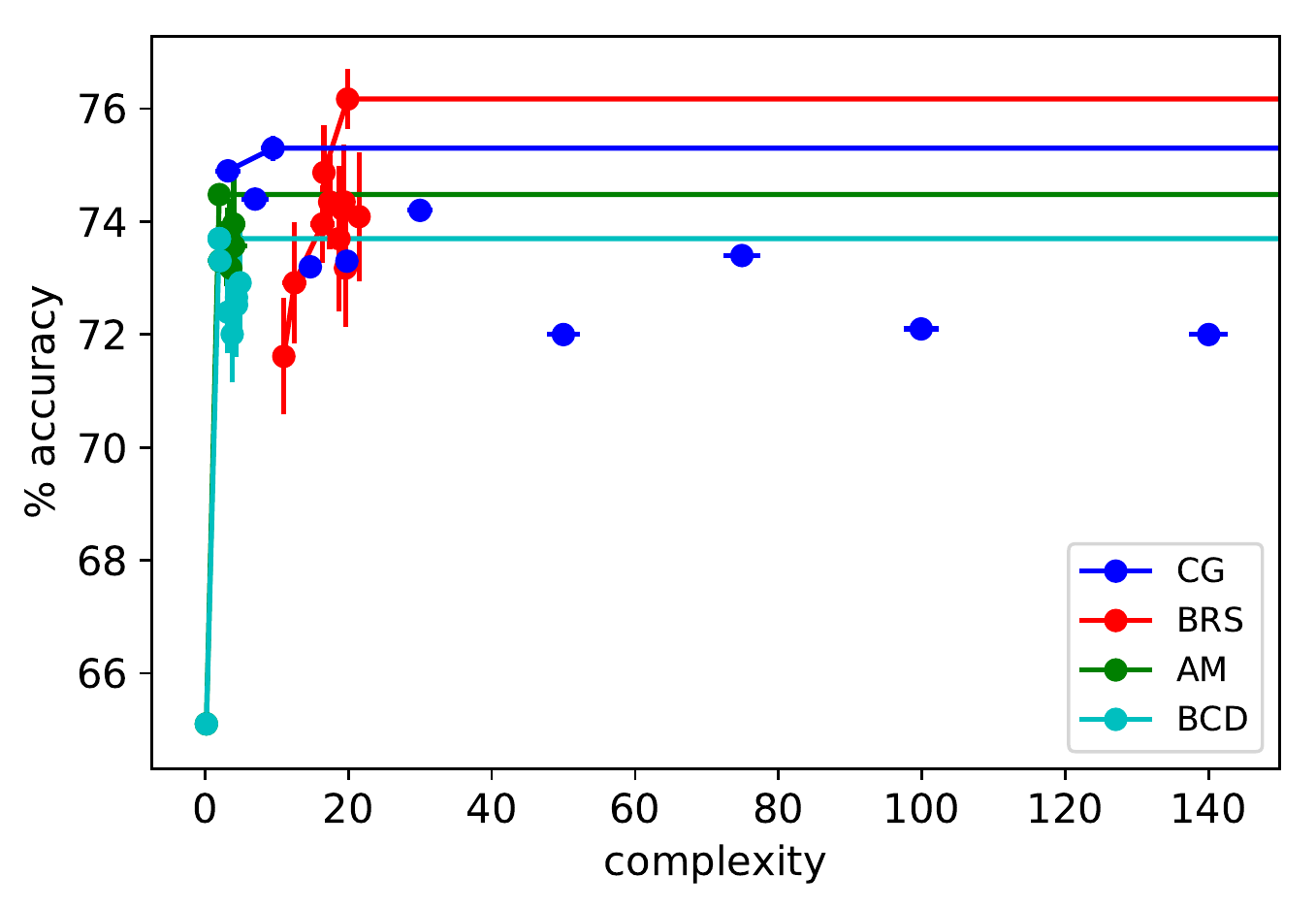}
       }
       \subfigure[tic-tac-toe]{
          \centering
          \includegraphics[width=0.4\textwidth]{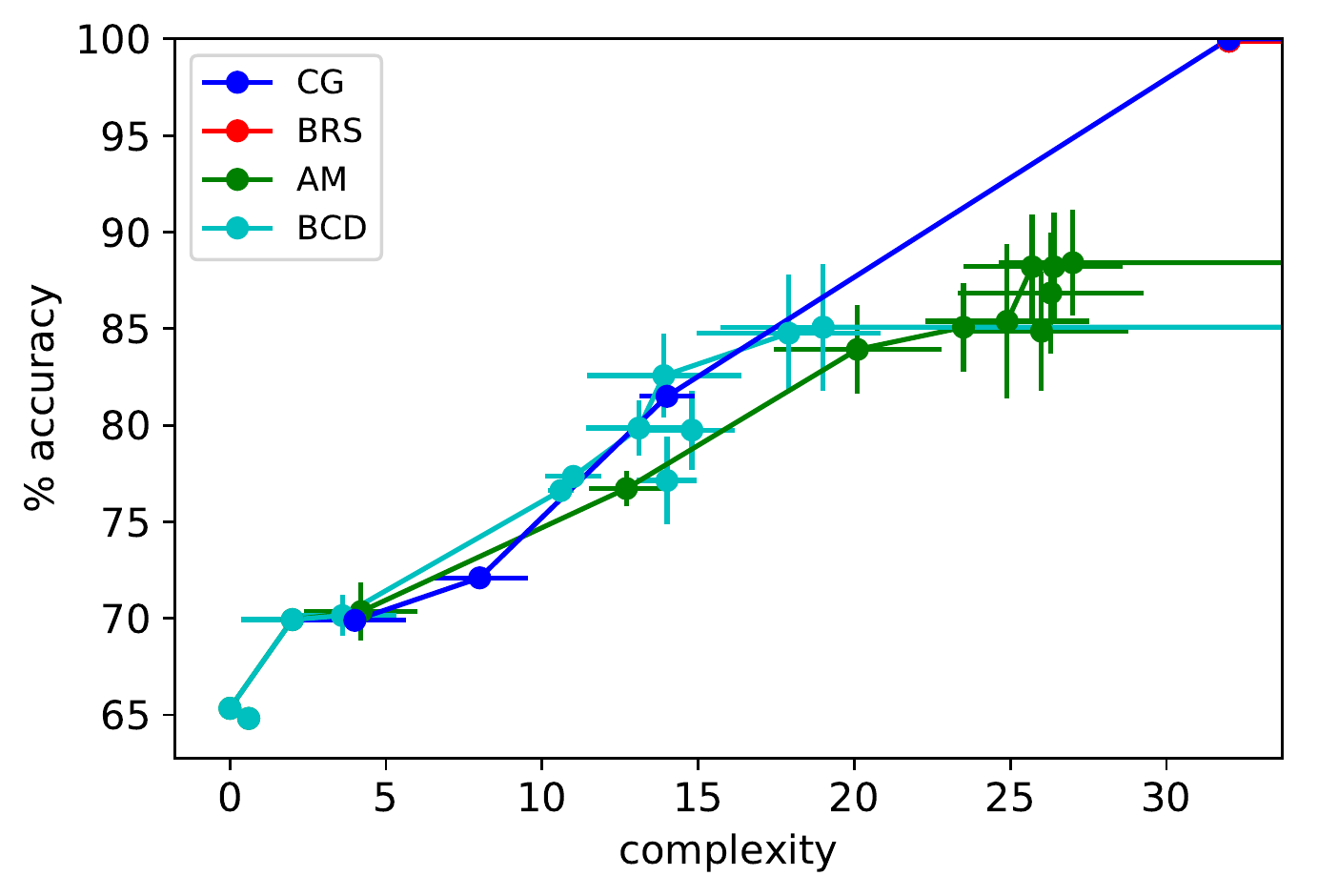}
       }
       \subfigure[transfusion]{
          \centering
          \includegraphics[width=0.4\textwidth]{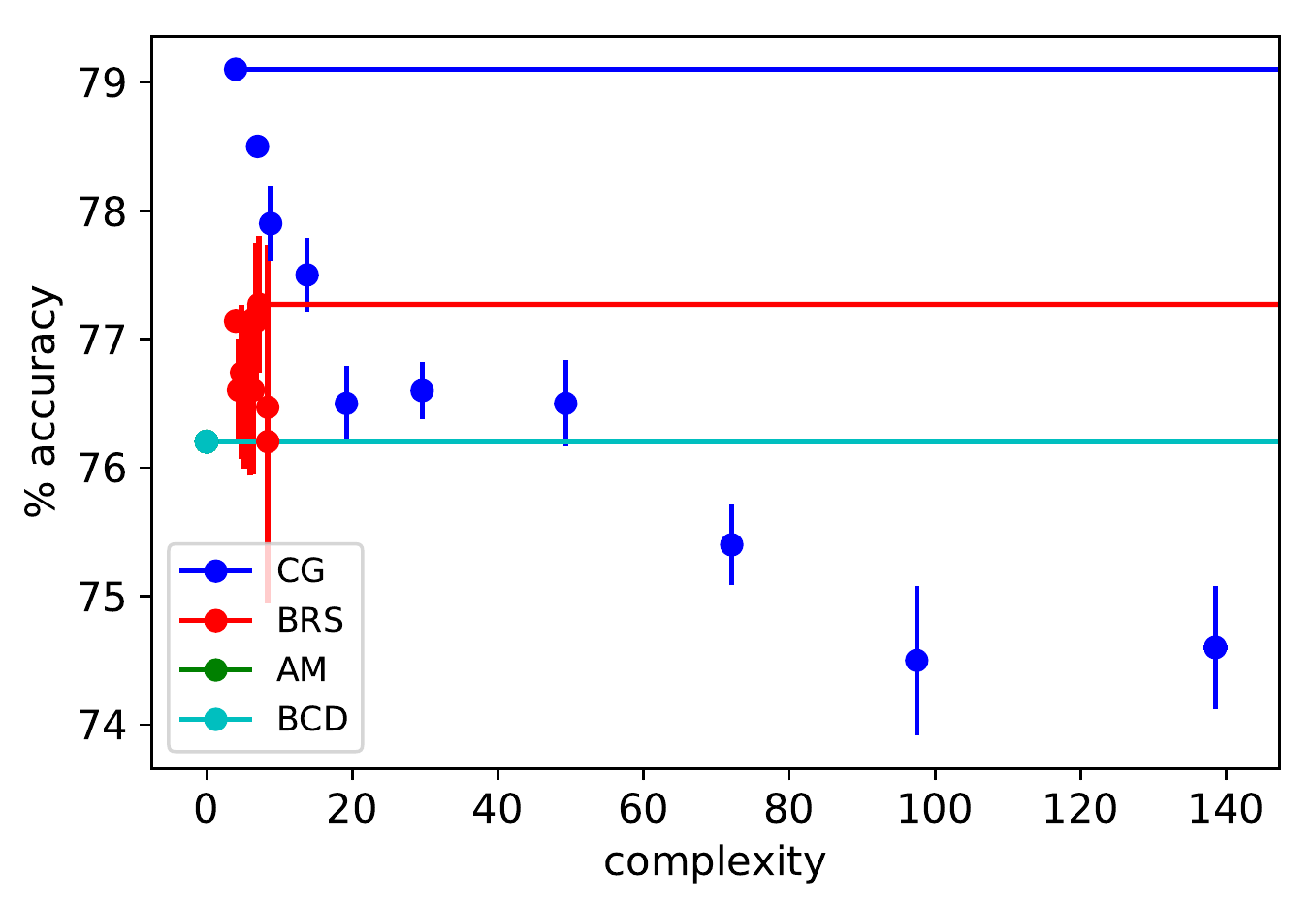}
       }
         \caption {Rule complexity-test accuracy trade-offs. Pareto efficient points are connected by line segments.)}
        \label{fig:paretoAll2}
    \end{figure}

    \begin{figure}[H] 
    \centering
        \subfigure[WDBC]{
          \includegraphics[width=0.45\textwidth]{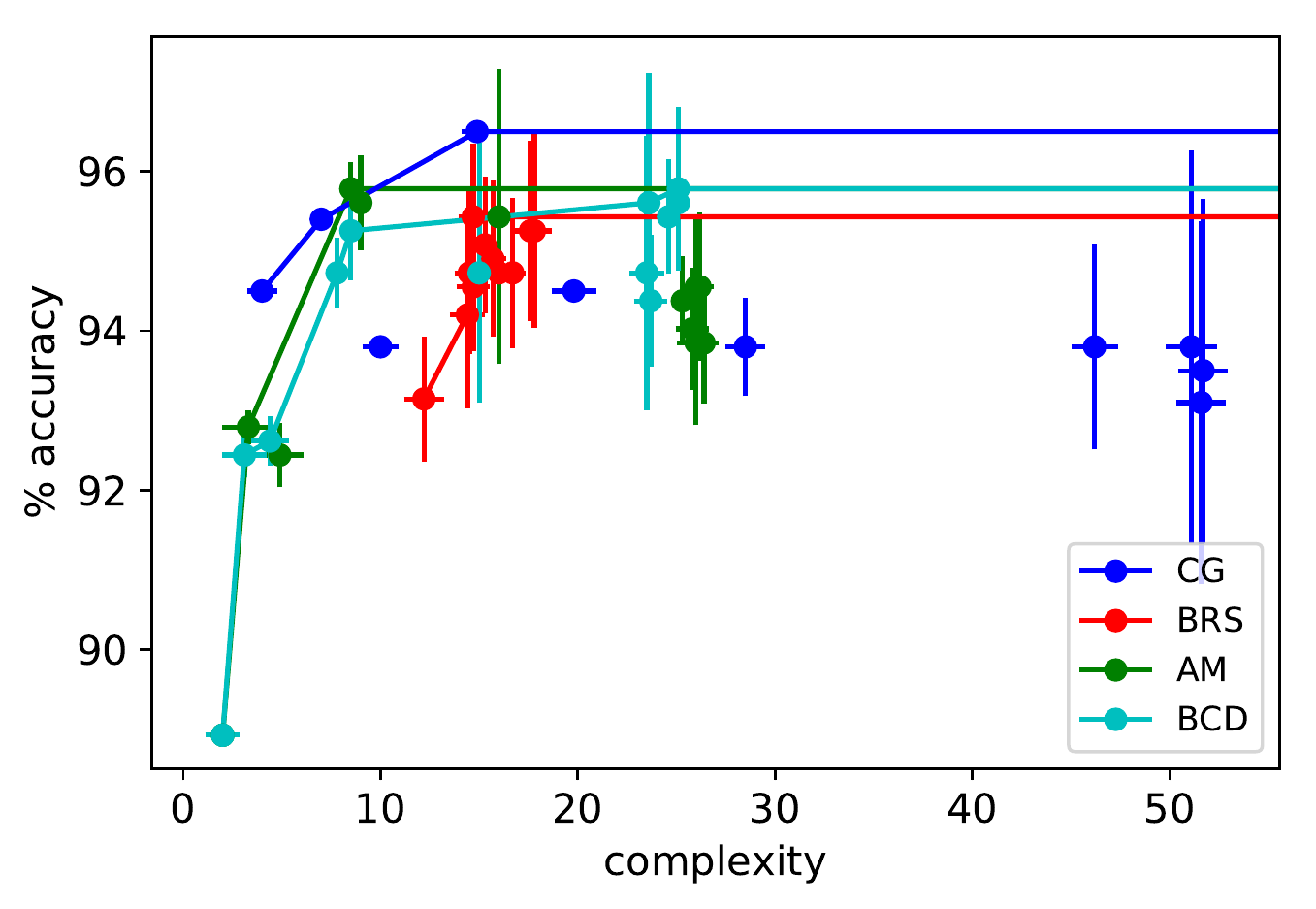}
        }
        \subfigure[adult]{
          \includegraphics[width=0.45\textwidth]{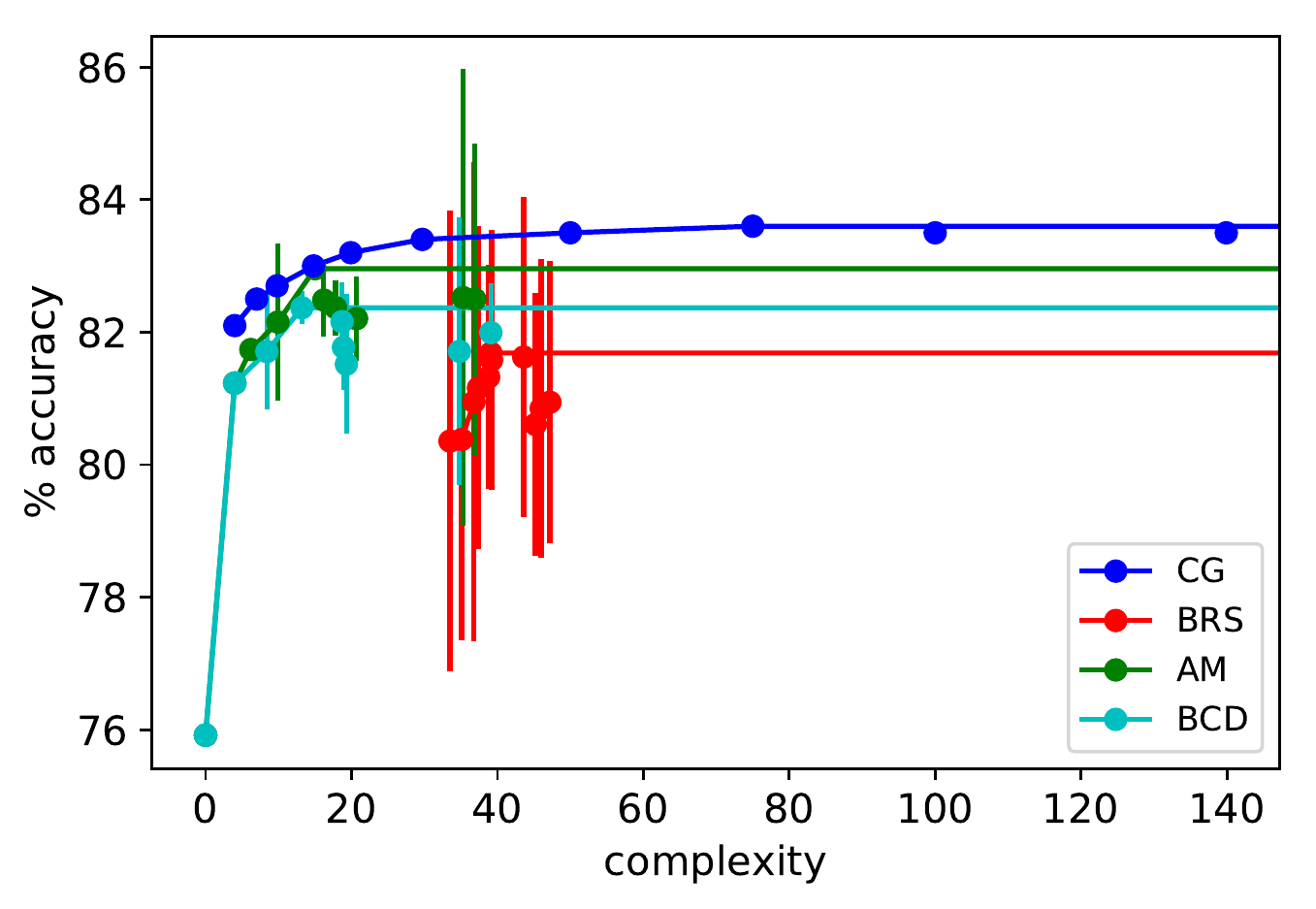}
        }
        \subfigure[bank-marketing]{
          \centering
          \includegraphics[width=0.45\textwidth]{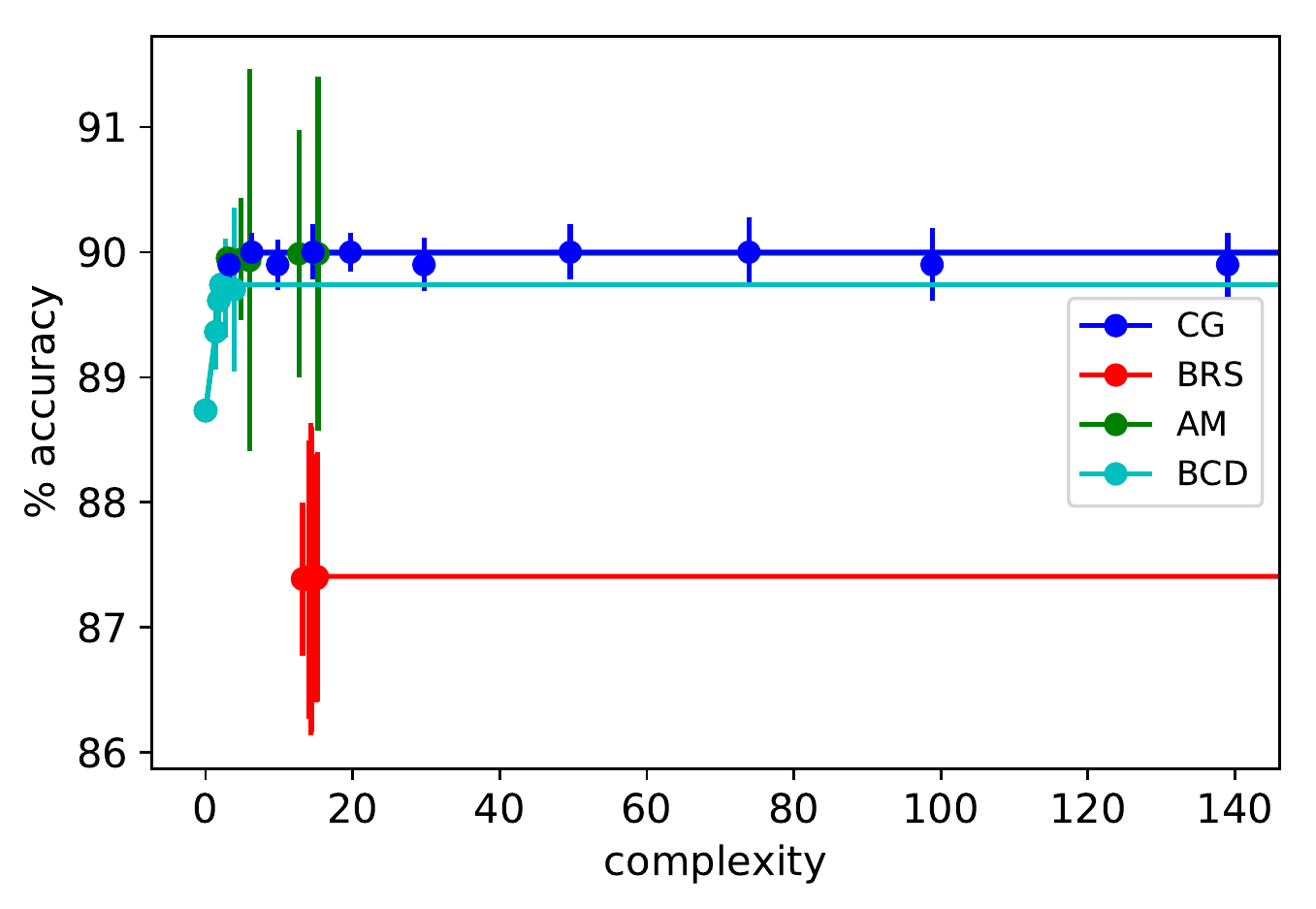}
        }
        \subfigure[gas]{
          \centering
          \includegraphics[width=0.45\textwidth]{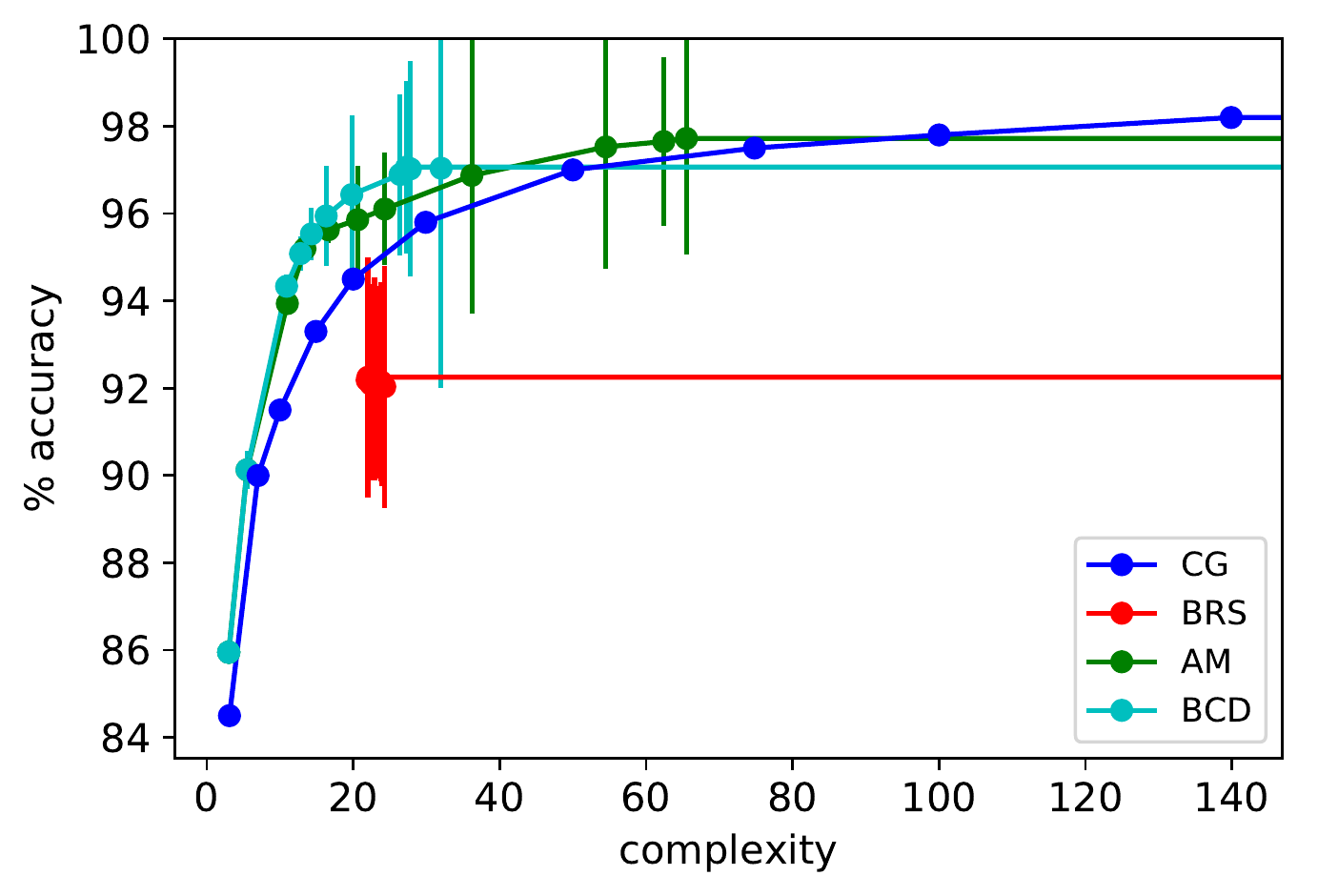}
       }
       \subfigure[magic]{
          \centering
          \includegraphics[width=0.45\textwidth]{figures/pareto_magic.pdf}
       }
       \subfigure[mushroom]{
          \centering
          \includegraphics[width=0.45\textwidth]{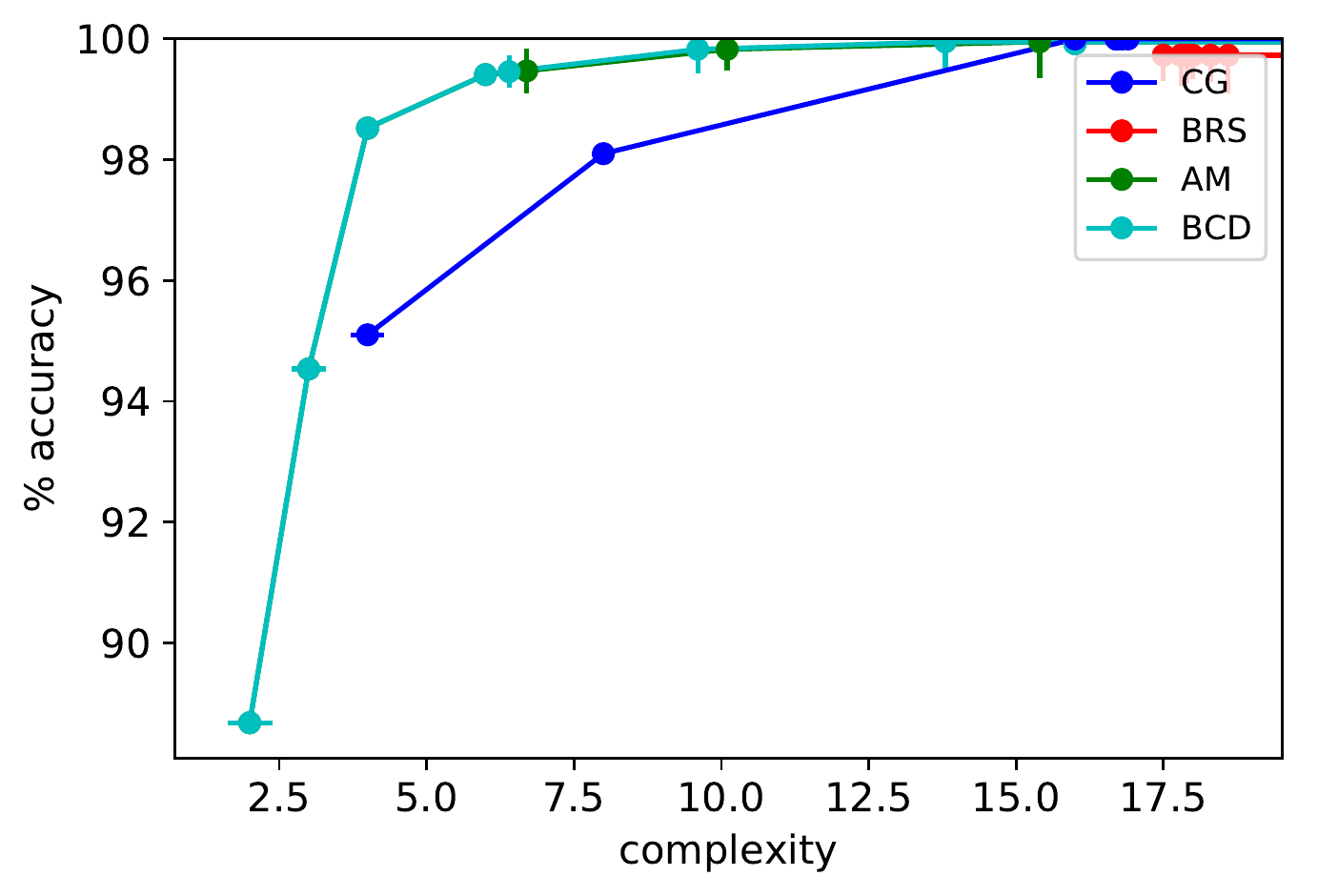}
       }
       \subfigure[musk]{
          \centering
          \includegraphics[width=0.45\textwidth]{figures/pareto_musk.pdf}
       }
       \subfigure[FICO]{
          \centering
          \includegraphics[width=0.45\textwidth]{figures/pareto_FICO.pdf}
       }
         \caption {Rule complexity-test accuracy trade-offs. Pareto efficient points are connected by line segments.)}
        \label{fig:paretoAll3}
    \end{figure}

Our partial results for IDS are shown in Tables~\ref{tbl:suppRuleAcc} and \ref{tbl:suppComplex}.  Despite ``cheating'' in the sense of choosing $\lambda_1 = \lambda_2$ to maximize accuracy after all the test results were known, the performance is not competitive with the other rule set algorithms on most data sets.  In addition to the constraints placed on Apriori, we suspect that another reason is that the IDS implementation available to us is designed primarily for the interval representation of numerical features and is not easily adapted to handle the alternative $(\leq, >)$ representation.

\begin{table}
\caption{Mean test accuracy for rule set classifiers (\%, standard error in parentheses)}
\label{tbl:suppRuleAcc}
\centering
\begin{tabular}{lllllll}
\toprule
data set&CG&BRS&AM&BCD&IDS&RIPPER\\
\midrule
banknote&$99.1$ ($0.3$)&$99.1$ ($0.2$)&$98.5$ ($0.4$)&$98.7$ ($0.2$)&$65.2$ ($2.1$)&$99.2$ ($0.2$)\\
heart&$78.9$ ($2.4$)&$78.9$ ($2.4$)&$72.9$ ($1.8$)&$74.2$ ($1.9$)&&$79.3$ ($2.2$)\\
ILPD&$69.6$ ($1.2$)&$69.8$ ($0.8$)&$71.5$ ($0.1$)&$71.5$ ($0.1$)&$71.5$ ($0.1$)&$69.8$ ($1.4$)\\
ionosphere&$90.0$ ($1.8$)&$86.9$ ($1.7$)&$90.9$ ($1.7$)&$91.5$ ($1.7$)&&$88.0$ ($1.9$)\\
liver&$59.7$ ($2.4$)&$53.6$ ($2.1$)&$55.7$ ($1.3$)&$51.9$ ($1.9$)&$51.0$ ($0.2$)&$57.1$ ($2.8$)\\
pima&$74.1$ ($1.9$)&$74.3$ ($1.2$)&$73.2$ ($1.7$)&$73.4$ ($1.7$)&$68.4$ ($0.9$)&$73.4$ ($2.0$)\\
tic-tac-toe&$100.0$ ($0.0$)&$99.9$ ($0.1$)&$84.3$ ($2.4$)&$81.5$ ($1.8$)&&$98.2$ ($0.4$)\\
transfusion&$77.9$ ($1.4$)&$76.6$ ($0.2$)&$76.2$ ($0.1$)&$76.2$ ($0.1$)&$76.2$ ($0.1$)&$78.9$ ($1.1$)\\
WDBC&$94.0$ ($1.2$)&$94.7$ ($0.6$)&$95.8$ ($0.5$)&$95.8$ ($0.5$)&$85.1$ ($2.2$)&$93.0$ ($0.9$)\\
\midrule
adult&$83.5$ ($0.3$)&$81.7$ ($0.5$)&$83.0$ ($0.2$)&$82.4$ ($0.2$)&&$83.6$ ($0.3$)\\
bank-mkt&$90.0$ ($0.1$)&$87.4$ ($0.2$)&$90.0$ ($0.1$)&$89.7$ ($0.1$)&&$89.9$ ($0.1$)\\
gas&$98.0$ ($0.1$)&$92.2$ ($0.3$)&$97.6$ ($0.2$)&$97.0$ ($0.3$)&&$99.0$ ($0.1$)\\
magic&$85.3$ ($0.3$)&$82.5$ ($0.4$)&$80.7$ ($0.2$)&$80.3$ ($0.3$)&$72.0$ ($0.1$)&$84.5$ ($0.3$)\\
mushroom&$100.0$ ($0.0$)&$99.7$ ($0.1$)&$99.9$ ($0.0$)&$99.9$ ($0.0$)&&$100.0$ ($0.0$)\\
musk&$95.6$ ($0.2$)&$93.3$ ($0.2$)&$96.9$ ($0.7$)&$92.1$ ($0.2$)&&$95.9$ ($0.2$)\\
FICO&$71.7$ ($0.5$)&$71.2$ ($0.3$)&$71.2$ ($0.4$)&$70.9$ ($0.4$)&&$71.8$ ($0.2$)\\
\bottomrule
\end{tabular}
\end{table}

In Table~\ref{tbl:suppOtherAcc}, accuracy results of logistic regression (LR) and support vector machine (SVM) classifiers are included along with those of non-rule set classifiers from the main text (CART and RF).  Although LR is a generalized linear model, it may not be regarded as interpretable in many application domains.  For SVM, we used a radial basis function (RBF) kernel and tuned both the kernel width as well as the complexity parameter $C$ using nested cross-validation.

\begin{table}
\caption{Mean test accuracy for other classifiers (\%, standard error in parentheses)}
\label{tbl:suppOtherAcc}
\centering
\begin{tabular}{lllll}
\toprule
data set&CART&RF&LR&SVM\\
\midrule
banknote&$96.8$ ($0.4$)&$99.5$ ($0.1$)&$98.8$ ($0.2$)&$99.9$ ($0.1$)\\
heart&$81.6$ ($2.4$)&$82.5$ ($0.7$)&$83.6$ ($2.5$)&$82.9$ ($1.9$)\\
ILPD&$67.4$ ($1.6$)&$69.8$ ($0.5$)&$72.9$ ($0.8$)&$70.8$ ($0.6$)\\
ionosphere&$87.2$ ($1.8$)&$93.6$ ($0.7$)&$86.9$ ($2.6$)&$94.9$ ($1.8$)\\
liver&$55.9$ ($1.4$)&$60.0$ ($0.8$)&$59.1$ ($2.0$)&$59.4$ ($1.7$)\\
pima&$72.1$ ($1.3$)&$76.1$ ($0.8$)&$77.9$ ($1.9$)&$76.8$ ($1.9$)\\
tic-tac-toe&$90.1$ ($0.9$)&$98.8$ ($0.1$)&$98.3$ ($0.4$)&$98.3$ ($0.4$)\\
transfusion&$78.7$ ($1.1$)&$77.3$ ($0.3$)&$77.0$ ($0.8$)&$77.0$ ($0.3$)\\
WDBC&$93.3$ ($0.9$)&$97.2$ ($0.2$)&$95.4$ ($0.9$)&$98.2$ ($0.4$)\\
\midrule
adult&$83.1$ ($0.3$)&$84.7$ ($0.1$)&$85.1$ ($0.2$)&$84.8$ ($0.2$)\\
bank-mkt&$89.1$ ($0.2$)&$88.7$ ($0.0$)&$89.8$ ($0.1$)&$88.7$ ($0.0$)\\
gas&$95.4$ ($0.1$)&$99.7$ ($0.0$)&$99.4$ ($0.1$)&$99.5$ ($0.1$)\\
magic&$82.8$ ($0.2$)&$86.6$ ($0.1$)&$79.0$ ($0.2$)&$87.7$ ($0.3$)\\
mushroom&$96.2$ ($0.3$)&$99.9$ ($0.0$)&$99.9$ ($0.1$)&$100.0$ ($0.0$)\\
musk&$90.1$ ($0.3$)&$86.2$ ($0.4$)&$93.1$ ($0.2$)&$97.8$ ($0.1$)\\
FICO&$70.9$ ($0.3$)&$73.1$ ($0.1$)&$71.6$ ($0.3$)&$72.3$ ($0.4$)\\
\bottomrule
\end{tabular}
\end{table}

\begin{table}
\footnotesize
\caption{Mean complexity (\# rules $+$ total \# conditions, standard error in parentheses)}
\label{tbl:suppComplex}
\centering
\begin{tabular}{llllllll}
\toprule
data set&CG&BRS&AM&BCD&IDS&RIPPER&CART\\
\midrule
banknote&$25.0$ ($1.9$)&$30.4$ ($1.1$)&$24.2$ ($1.5$)&$21.3$ ($1.9$)&$11.2$ ($0.5$)&$28.6$ ($1.1$)&$51.8$ ($1.4$)\\
heart&$11.3$ ($1.8$)&$24.0$ ($1.6$)&$11.5$ ($3.0$)&$15.4$ ($2.9$)&&$16.0$ ($1.5$)&$32.0$ ($8.1$)\\
ILPD&$10.9$ ($2.7$)&$4.4$ ($0.4$)&$0.0$ ($0.0$)&$0.0$ ($0.0$)&$2.0$ ($0.0$)&$9.5$ ($2.5$)&$56.5$ ($10.9$)\\
ionosphere&$12.3$ ($3.0$)&$12.0$ ($1.6$)&$16.0$ ($1.5$)&$14.6$ ($1.4$)&&$14.6$ ($1.2$)&$46.1$ ($4.2$)\\
liver&$5.2$ ($1.2$)&$15.1$ ($1.3$)&$8.7$ ($1.8$)&$4.0$ ($1.1$)&$0.0$ ($0.0$)&$5.4$ ($1.3$)&$60.2$ ($15.6$)\\
pima&$4.5$ ($1.3$)&$17.4$ ($0.8$)&$2.7$ ($0.6$)&$2.1$ ($0.1$)&$6.0$ ($0.3$)&$17.0$ ($2.9$)&$34.7$ ($5.8$)\\
tic-tac-toe&$32.0$ ($0.0$)&$32.0$ ($0.0$)&$24.9$ ($3.1$)&$12.6$ ($1.1$)&&$32.9$ ($0.7$)&$67.2$ ($5.0$)\\
transfusion&$5.6$ ($1.2$)&$6.0$ ($0.7$)&$0.0$ ($0.0$)&$0.0$ ($0.0$)&$2.0$ ($0.0$)&$6.8$ ($0.6$)&$14.3$ ($2.3$)\\
WDBC&$13.9$ ($2.4$)&$16.0$ ($0.7$)&$11.6$ ($2.2$)&$17.3$ ($2.5$)&$15.2$ ($0.7$)&$16.8$ ($1.5$)&$15.6$ ($2.2$)\\
\midrule
adult&$88.0$ ($11.4$)&$39.1$ ($1.3$)&$15.0$ ($0.0$)&$13.2$ ($0.2$)&&$133.3$ ($6.3$)&$95.9$ ($4.3$)\\
bank-mkt&$9.9$ ($0.1$)&$13.2$ ($0.6$)&$6.8$ ($0.7$)&$2.1$ ($0.1$)&&$56.4$ ($12.8$)&$3.0$ ($0.0$)\\
gas&$123.9$ ($6.5$)&$22.4$ ($2.0$)&$62.4$ ($1.9$)&$27.8$ ($2.5$)&&$145.3$ ($4.2$)&$104.7$ ($1.0$)\\
magic&$93.0$ ($10.7$)&$97.2$ ($5.3$)&$11.5$ ($0.2$)&$9.0$ ($0.0$)&$10.0$ ($0.0$)&$177.3$ ($8.9$)&$125.5$ ($3.2$)\\
mushroom&$17.8$ ($0.3$)&$17.5$ ($0.4$)&$15.4$ ($0.6$)&$14.6$ ($0.6$)&&$17.0$ ($0.4$)&$9.3$ ($0.2$)\\
musk&$123.9$ ($6.5$)&$33.9$ ($1.3$)&$101.3$ ($11.6$)&$24.4$ ($1.9$)&&$143.4$ ($5.5$)&$17.0$ ($0.7$)\\
FICO&$13.3$ ($4.1$)&$23.2$ ($1.4$)&$8.7$ ($0.4$)&$4.8$ ($0.3$)&&$88.1$ ($7.0$)&$155.0$ ($27.5$)\\
\bottomrule
\end{tabular}
\end{table}

\newpage

\cl{
\section{Stability Example} \label{app:instability}
 Consider the following rule sets generated for two different splits of the compas data (both trained with $C=30$ and no fairness constraint).
    
    \medskip\noindent\textbf{Rule Set 1:}\text{ Predict repeat offence if:}
    \begin{center}
    \big[\text{(Score Factor$=$True) and (Age $\leq$ 45)  }\big] \\ 
    \text{~OR~}\\
    \big[\text{(Score Factor$=$True) and (Age $>$ 45) and (Misdemeanor$=$False) and } \\ 
    \text{(Priors$\geq$ 4) and (Gender$=$Male)}\big] 
    \end{center}
    
    \medskip\noindent\textbf{Rule Set 2:}\text{ Predict repeat offence if:}
    \begin{center}
    \big[\text{(Score Factor$=$True) and (Gender$=$Female)  }\big] \\ 
    \text{~OR~}\\
    \big[\text{(Score Factor$=$True) and (Age $\leq$ 45) and (Age$\geq$25) and } \\ 
    \text{(Priors$\geq$ 10) and (Gender$=$Male)}\big] 
    \end{center}
    
    \noindent
    Despite both rule sets having near identical performance on their respective train and test sets, the two rule sets are qualitatively quite different. For instance, the rule sets share no common rules or even rules with the exact same features. The second rule in both rule sets involves the number of priors but use different thresholds (4 vs. 10). This lack of stability is a well documented problem in interpretable models \citep{guidotti2018assessing}. It remains an open research problem on how to improve the stability of rule sets and other interpretable classification models.}

\begin{table}[b]
\centering
\caption{\label{eps_hp} Overview of $\epsilon$ hyperparameters Tested}
\setlength{\tabcolsep}{5pt} 
\begin{tabular}{c c c c c}		\toprule
data set & $\epsilon$ Phase 1 & $\epsilon$ Phase 2 \\ \midrule 
Adult &\{0.01, 0.1, 1\}  & \{0, 0.01, 0.05, 0.1, 0.15, 0.2, 1\} \\
Compas &\{0.01, 0.1, 1\}  & \{0, 0.01, 0.05, 0.1, 0.15, 0.2, 1\} \\
Default &\{0.01, 0.1, 1\}  & \{0, 0.01, 0.03, 0.05, 0.1, 0.2, 1\} \\
\bottomrule
\end{tabular}%
\end{table}%

\section{Parameters for Fairness-Accuracy Trade-offs} \label{app:params_fairness}

For our experiments we took a two-phase approach. During the first rule generation phase, we ran our column generation algorithm with a set of different hyperparameters to generate a set of potential rules. To warm start this procedure, we also start the column generation process with a set of rules mined from a random forest classifier. We then solve the master IP with the set of candidate rules and a larger set of hyperparameters to generate the curves included in the body of the report. Table \ref{eps_hp} summarizes the hyperparameters used for both Phase I and II. Note that for the equalized odds formulation, we set $\epsilon_1 = \epsilon_2$ and use the values in Table \ref{eps_hp}.

\begin{table}[hbt]
\centering
\caption{\label{c_hp} Overview of $C$ Hyperparameters Tested}
\setlength{\tabcolsep}{5pt} 
\begin{tabular}{c c c c c}		\toprule
data set & Phase 1 & Phase 2  \\ \midrule 
Adult &\{5, 20, 40, 80, 100\}  & \{5, 15, 20, 30, 40, 50, 60, 80, 100\} \\
Compas &\{5, 15, 30\}  & \{5, 10, 15, 20, 30\} \\
Default &\{5, 15, 30\}  & \{5, 10, 15, 20, 30\} \\
\bottomrule
\end{tabular}%
\end{table}%

\begin{table}[hbt]
\centering

\caption{\label{zafar_hp} Overview of Zafar Optimizer Hyperparameters}
\footnotesize
\setlength{\tabcolsep}{5pt} 
\begin{tabular}{c c c}		\toprule
data set & $\tau$ & $\mu$  \\ \midrule 
Adult & 5 & 1.2\\
Compas &20  & 1.2\\
Default & 0.5 & 1.2 \\
\bottomrule
\end{tabular}%
\end{table}%

We tested our algorithm against three other popular interpretable fair classifiers: Zafar 2017 \citep{Zafar_2017}, Hardt 2016 \citep{hardt2016equality}, and Fair Decision trees trained using the exponentiated gradient algorithm in fairlearn \citep{agarwal2018reductions}. For the Zafar algorithm we used the optimizer parameters specified in Table \ref{zafar_hp}, and tested a range of 30 different $\epsilon$ values (linearly spaced between 0 and 0.5) for the covariance threshold. For Hardt 2016 we used the logistic regression implementation from scikit-learn \citep{scikit-learn} and tested 100 different decision thresholds for each sub-group (1\% increments). For the exponentiated gradient algorthm from fairlearn we used scikit-learn's decision tree as the base estimator and tested both a range of maximum depth hyperparameters (20 values linearly spaced between 1 and 30), and 30 different $\epsilon$ values (linearly spaced between 0 and 0.5) for the fairness constraints. For all the algorithms we used 10-fold nested cross-validation to select the best hyperparameters for every level of fairness and removed dominated points from the figures present in the results section of the main paper.

\cl{
\section{Results for Additional Fair Classifiers} \label{app:faircorels}

The following section reports additional results for the Fair CORELS classifier. We set a maximum cardinality, i.e., the number of features allowed in a rule, of 2 for Adult and Default, and 5 for compas. We found that increasing the cardinality above 2 for the two large data sets led to memory issues on our computing infrastructure with 16 GB RAM. We tested possible values for the complexity regularization parameter $\lambda$ on a logarithmic scale from $0.0001$ to $10$, and use the default parameters for other settings. Table \ref{accTableFairCorelsNoConst} shows the performance of the classifier in the absence of fairness constraints. For every data set, the Fair CORELS classifier has lower accuracy than the other benchmark approaches. This is in contrast to the reported numbers in \cite{aivodji2021faircorels} which show competitive accuracy on both Adult and compas. It is unclear if the degradation in performance is a function of our binarization process and data set pre-processing, or the limitations of our computing infrastructure, or a discrepancy between the publicly available code for Fair CORELS and the code used to obtain the results in \cite{aivodji2021faircorels}.

\begin{table*}[t]
\footnotesize
\centering
\caption{\label{accTableFairCorelsNoConst} Mean test accuracy and fairness results with no fairness constraints (standard deviation in parenthesis). Equality of opportunity and equalized odds refer to the amount of unfairness between the two groups under each fairness metric.}
\setlength{\tabcolsep}{5pt} 
\begin{tabular}{l l  c c c c c}		\toprule
& & Fair CG & Zafar & Hardt & Fair Learn & Fair CORELS\\\midrule
\multirow{3}{*}{Adult} & Accuracy & 82.5 (0.5) & 85.2 (0.5) & 83.0 (0.4) & 82.4 (0.4) &  77.7 (2.7) \\
& Equality of Opportunity & 7.6 (0.5) & 11.9 (3.7) & 18.2 (4.8) & 11.5 (4.6) & 3.0 (5.3)\\
& Equalized Odds & 7.6 (0.5) & 11.9 (3.7) &  18.2 (4.8) & 11.5 (4.6) & 3.0 (5.3)\\ \midrule
\multirow{3}{*}{Compas} & Accuracy & 67.6 (1.1) & 64.6 (1.9) & 65.9 (2.7) & 65.8 (2.9) & 60.8 (6.1)\\
& Equality of Opportunity & 23.8 (5.3)& 42.8 (5.4) & 23.7 (6.4) & 21.7 (7.1) &  14.1 (11.5) \\
& Equalized Odds &24.1 (5.1) & 47.6 (5.8)& 27.0 (5.2) & 24.9 (4.5) &  14.8 (13.1)\\ \midrule
\multirow{3}{*}{Default} & Accuracy & 82.0 (0.7) & 81.2 (0.8) & 77.9 (1.7) & 77.9 (1.7) & 77.9 (1.7) \\
& Equality of Opportunity & 1.3 (0.6) & 2.7 (1.9) & 0 (0) & 0 (0) & 0 (0) \\
& Equalized Odds & 1.9 (0.5) & 4.2 (2.5) & 0 (0) & 0 (0) &  0 (0)\\ 
\bottomrule
\end{tabular}%
\end{table*}%

Figure \ref{benchmark_combined_faircorels} shows the full Pareto curve of accuracy versus fairness for the benchmark fair classifiers including Fair CORELS. In every data set Fair CORELS is dominated by the Fair CG approach, though it outperforms the other benchmark algorithms in areas with low unfairness.

\begin{figure}[H]
    \centering
        \begin{subfigure}
          \centering
          \includegraphics[width=0.4\linewidth]{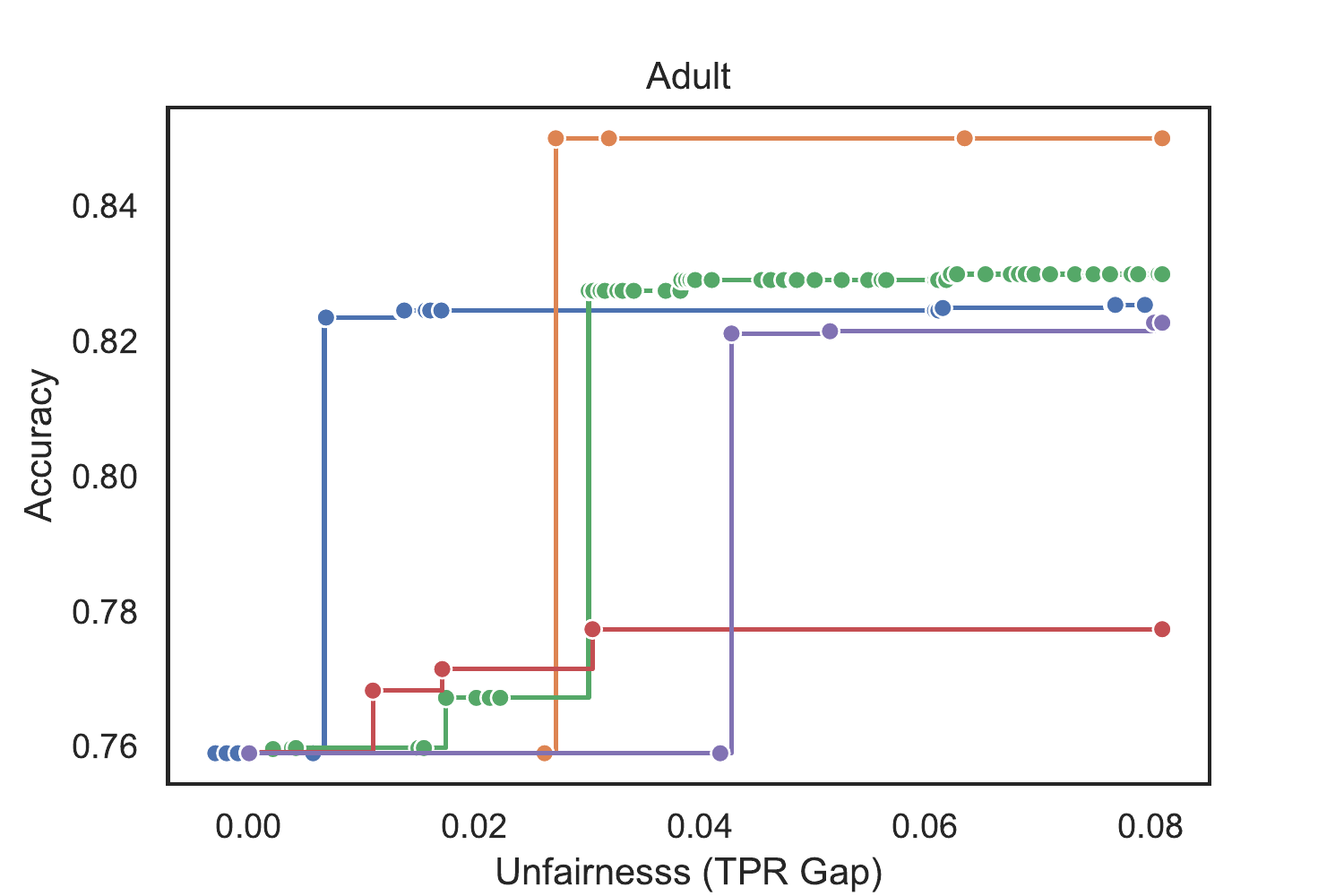}
        \end{subfigure}
        \begin{subfigure}
          \centering
          \includegraphics[width=0.4\linewidth]{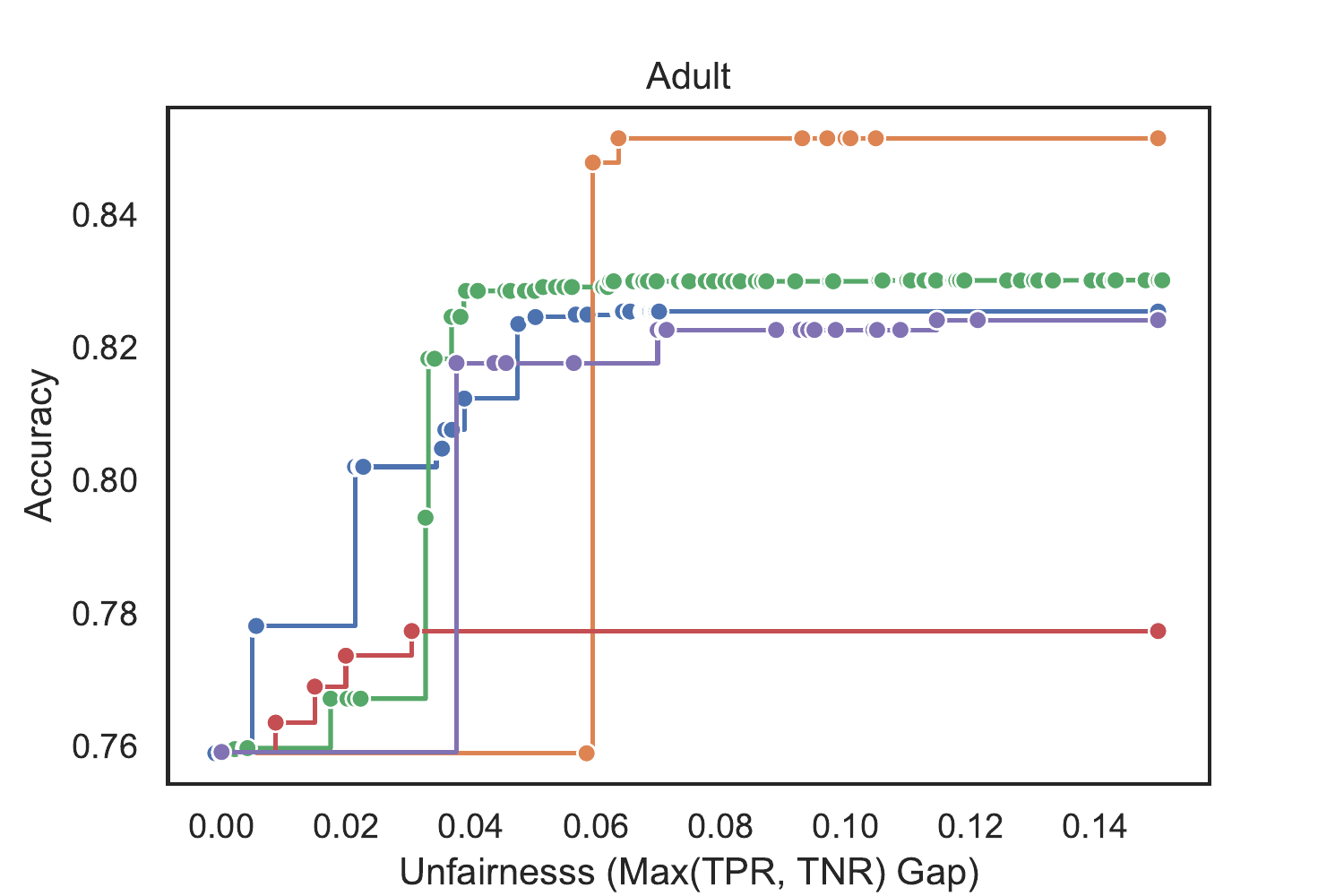}
        \end{subfigure}
        \begin{subfigure}
          \centering
          \includegraphics[width=0.4\linewidth]{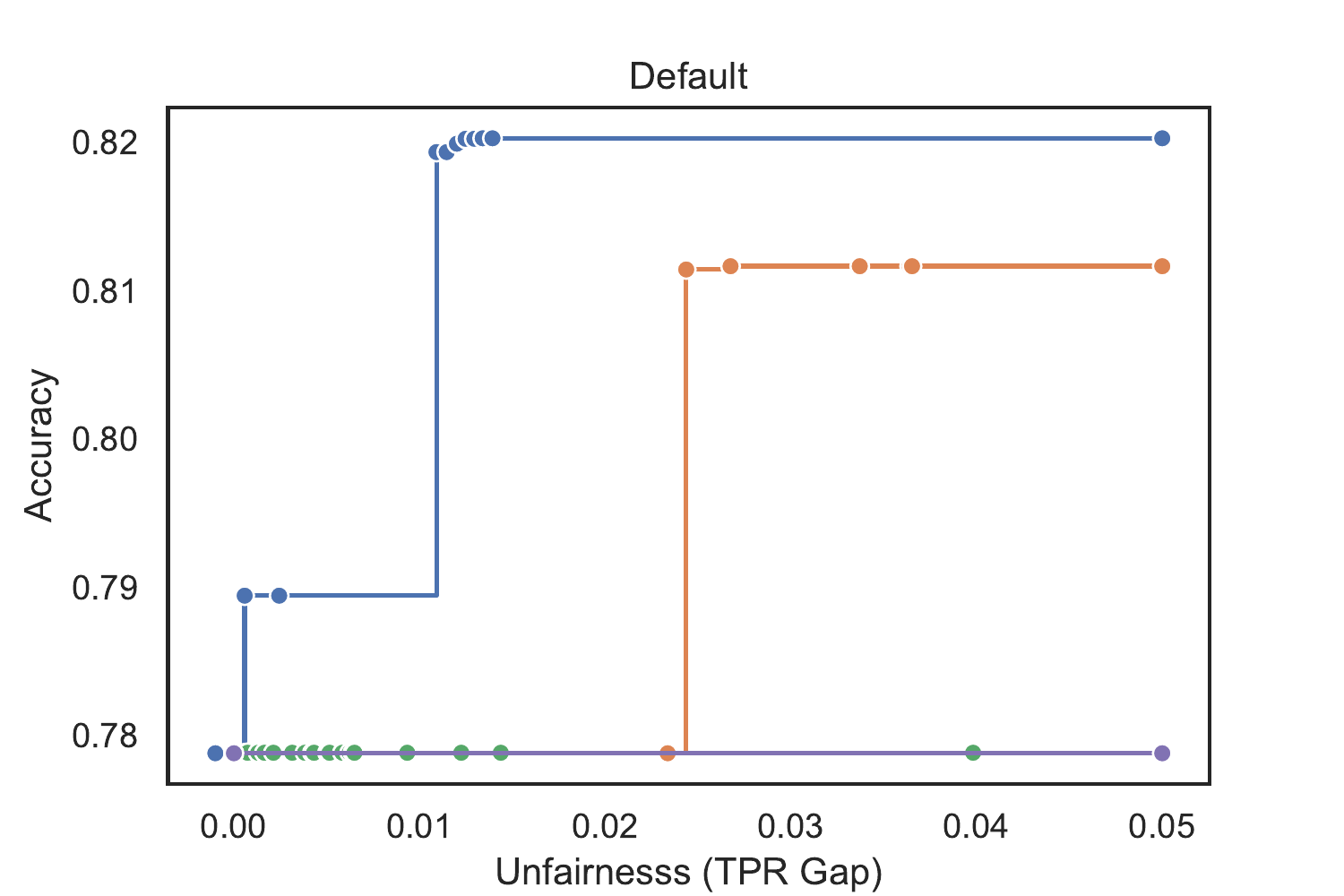}
        \end{subfigure}
        \begin{subfigure}
          \centering
          \includegraphics[width=0.4\linewidth]{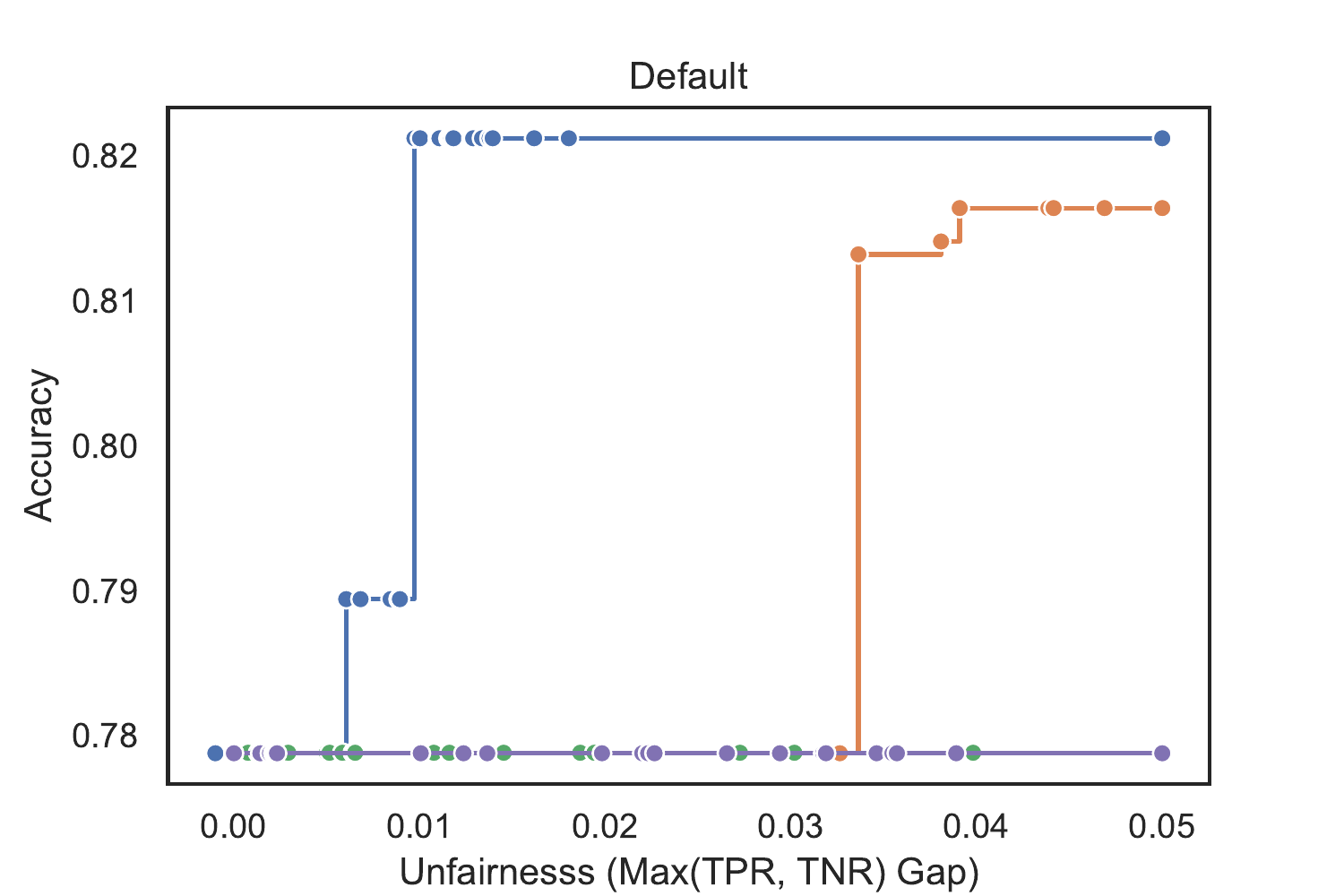}
        \end{subfigure}
    \begin{subfigure}
          \centering
          \includegraphics[width=0.4\linewidth]{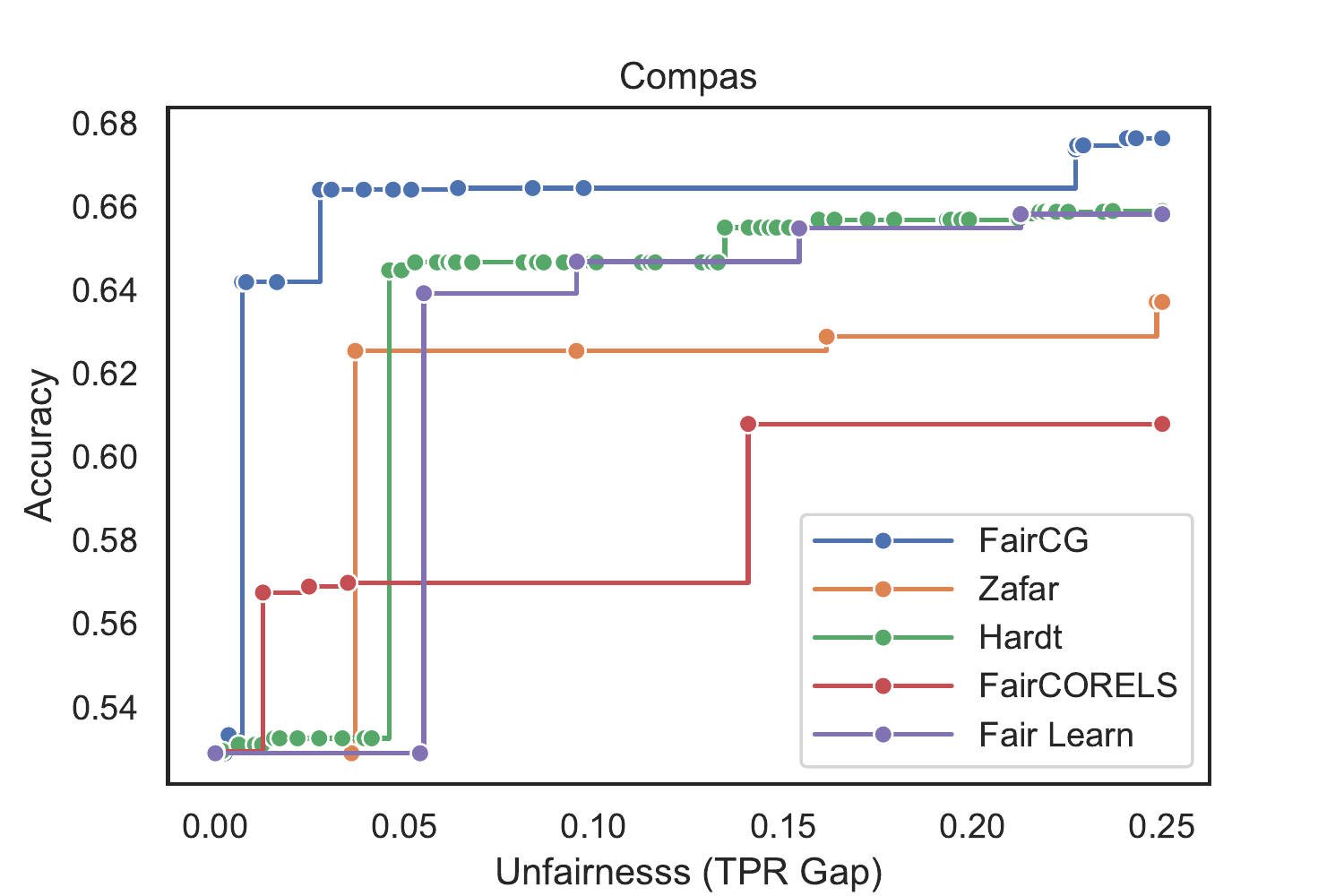}
        \end{subfigure}
    \begin{subfigure}
          \centering
          \includegraphics[width=0.4\linewidth]{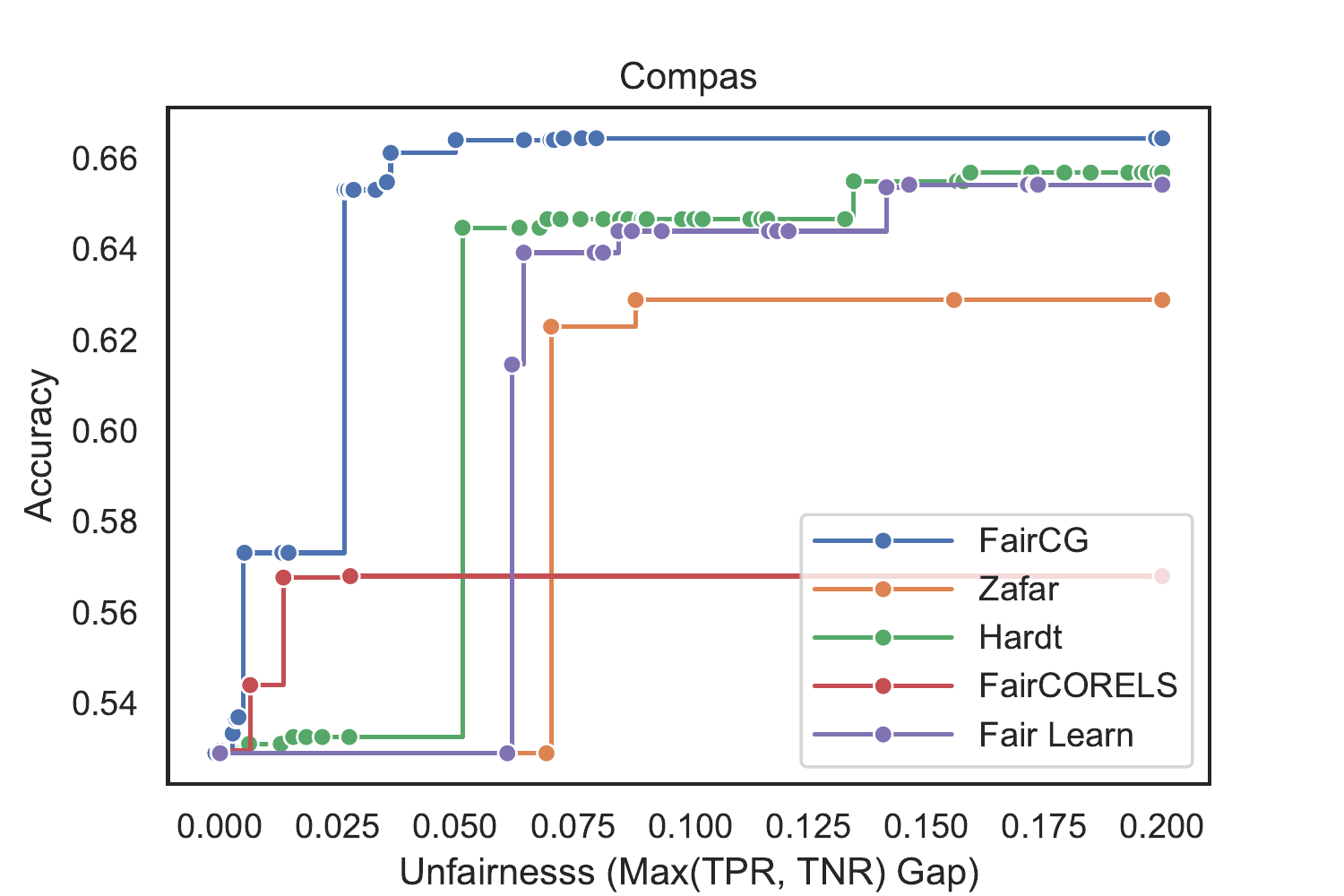}
    \end{subfigure}
    \caption{Test Accuracy Fairness Frontier for Fair CG and other interpretable fair classifiers with respect to \emph{equality of opportunity} (left column) and \emph{equalized odds} (right column).}
    \label{benchmark_combined_faircorels} 
    \end{figure}
}

\newpage

\bibliography{bibliography}

\end{document}

%% file: preamble.tex
\usepackage{amsfonts,cancel}
\usepackage{thmtools,enumitem}
\usepackage{algorithm}
\usepackage{mathtools}
\usepackage{dsfont}
\usepackage{bbm}
\usepackage{tensor}
\usepackage{bm}
\usepackage{graphicx}
\usepackage{multirow}
\usepackage{subfigure}
\usepackage{comment}
\usepackage{afterpage}
\usepackage{float}

\newcommand{\bx}{\mathbf{x}}

\newcommand{\cA}{\mathcal{A}}
\newcommand{\cJ}{\mathcal{J}}

\renewcommand{\P}{\mathcal{P}}

\renewcommand{\Pr}{\mathbb{P}}

\newcommand{\K}{\mathcal{K}}
\newcommand{\G}{\mathcal{G}}

\newcommand{\cl}[1]{#1}

\newcommand{\ignore}[1]{}

\newcommand{\mysecx}{\vspace{.02cm}}
\newcommand{\mysec}[1]{\mysecx {\bf #1: }}

\mathchardef\mhyphen="2D 
\DeclareMathOperator*{\argmin}{\arg\!\min}

\DeclarePairedDelimiter{\abs}{\lvert}{\rvert}

\let\originalleft\left
\let\originalright\right
\renewcommand{\left}{\mathopen{}\mathclose\bgroup\originalleft}
\renewcommand{\right}{\aftergroup\egroup\originalright}

